\documentclass[oneside,11pt]{article}

\usepackage{times,url,mathrsfs,algorithm}
\usepackage{amsmath}
\usepackage{amsfonts}
\usepackage{graphicx}
\usepackage{subfigure}

\begin{document}

\title{An Empirical Evaluation of Four Algorithms for Multi-Class Classification: Mart, ABC-Mart, Robust LogitBoost,  and ABC-LogitBoost}

\author{ Ping Li \\
       Department of Statistical Science\\
       Faculty of Computing and Information Science\\
       Cornell University\\
       Ithaca, NY 14853\\
       pingli@cornell.edu}
\date{}

\maketitle

\begin{abstract}

This empirical study is mainly devoted to comparing \textbf{four} tree-based boosting algorithms:
 \textbf{\em mart}, \textbf{\em abc-mart}, \textbf{\em robust logitboost}, and \textbf{\em abc-logitboost}, for multi-class classification on a variety of publicly available datasets. Some of those datasets have been thoroughly tested in prior studies using a broad range of classification algorithms including SVM, neural nets, and deep learning.

In terms of the empirical classification errors, our experiment results  demonstrate:
\begin{enumerate}
\item {\em Abc-mart} considerably improves {\em mart}.
\item {\em Abc-logitboost} considerably improves {\em (robust) logitboost}.
\item {\em (Robust) logitboost} considerably improves {\em mart} on most datasets.
\item {\em Abc-logitboost}  considerably improves {\em abc-mart} on most datasets.
\item These four boosting algorithms (especially {\em abc-logitboost})  outperform SVM on many datasets.
\item Compared to the best deep learning methods, these four boosting algorithms (especially {\em abc-logitboost}) are competitive.
\end{enumerate}

\end{abstract}

\section{Introduction}
Boosting algorithms \cite{Article:Schapire_ML90,Article:Freund_95,Article:Freund_JCSS97,Article:Bartlett_AS98,Article:Schapire_ML99,Article:FHT_AS00,Proc:Mason_NIPS00,Article:Friedman_AS01} have become very successful in machine learning.  In this paper, we provide an empirical evaluation of \textbf{four} tree-based  boosting algorithms for multi-class classification: \textbf{\em mart}\cite{Article:Friedman_AS01}, \textbf{\em abc-mart}\cite{Proc:ABC_ICML09}, \textbf{\em robust logitboost}\cite{Report:Li_Robust-LogitBoost}, and \textbf{\em abc-logitboost}\cite{Report:Li_ABC-LogitBoost}, on a wide range of datasets.

{\em Abc-boost}\cite{Proc:ABC_ICML09}, where ``{\em abc}'' stands for {\em adaptive base class}, is a recent new idea for improving multi-class classification.  Both {\em abc-mart}\cite{Proc:ABC_ICML09} and {\em abc-logitboost}\cite{Report:Li_ABC-LogitBoost} are specific implementations of {\em abc-boost}.  Although the  experiments in \cite{Proc:ABC_ICML09,Report:Li_ABC-LogitBoost} were reasonable, we consider a more thorough study is necessary.  Most datasets used in \cite{Proc:ABC_ICML09,Report:Li_ABC-LogitBoost} are (very) small.  While those datasets (e.g., {\em pendigits}, {\em zipcode}) are still popular in machine learning research papers, they may be too small to be practically very meaningful. Nowadays, applications with millions of training samples are not uncommon, for example, in search engines\cite{Proc:McRank_NIPS07}.

It would be also interesting to compare these four tree-based boosting algorithms with other  popular learning
methods such as {\em support vector machines (SVM)} and {\em deep learning}. A recent study\cite{Proc:Larochelle_ICML07}\footnote{
\url{http://www.iro.umontreal.ca/~lisa/twiki/bin/view.cgi/Public/DeepVsShallowComparisonICML2007}} conducted a thorough empirical comparison of many learning algorithms including SVM, neural nets, and deep learning. The authors of \cite{Proc:Larochelle_ICML07} maintain a nice Web site from which one can download the datasets and compares the test mis-classification errors.

In this paper, we provide extensive experiment results using {\em mart}, {\em abc-mart}, {\em robust logitboost}, and {\em abc-logitboost} on the  datasets  used in \cite{Proc:Larochelle_ICML07}, plus other  publicly available datasets. One interesting dataset is the UCI {\em Poker}. By private communications with C.J. Lin (the author of LibSVM), we learn that SVM achieved a classification accuracy of $\leq 60\%$ on this dataset. Interestingly,  all four  boosting algorithms can easily achieve $>90\%$ accuracies.\\

We try to make this paper self-contained by providing a detailed introduction to {\em abc-mart}, {\em robust logitboost}, and {\em abc-logitboost} in the next section.

\section{LogitBoost, Mart, Abc-mart, Robust LogitBoost, and Abc-LogitBoost}

We denote a training dataset by $\{y_i,\mathbf{x}_i\}_{i=1}^N$, where $N$ is the number of feature vectors (samples), $\mathbf{x}_i$ is the $i$th feature vector, and  $y_i \in \{0, 1, 2, ..., K-1\}$ is the $i$th class label, where $K\geq 3$ in multi-class classification.

Both {\em logitboost}\cite{Article:FHT_AS00} and {\em mart} (multiple additive regression trees)\cite{Article:Friedman_AS01} algorithms can be viewed as generalizations to logistic regression, which assumes class probabilities $p_{i,k}$  as
\begin{align}\label{eqn_logit}
p_{i,k} = \mathbf{Pr}\left(y_i = k|\mathbf{x}_i\right) = \frac{e^{F_{i,k}(\mathbf{x_i})}}{\sum_{s=0}^{K-1} e^{F_{i,s}(\mathbf{x_i})}}.
\end{align}
While traditional logistic regression assumes $F_{i,k}(\mathbf{x}_i) = \beta^\text{T}\mathbf{x}_i$, {\em logitboost} and {\em mart} adopt the flexible ``additive model,''  which is a function of $M$ terms:
\begin{align}\label{eqn_F_M}
F^{(M)}(\mathbf{x}) = \sum_{m=1}^M \rho_m h(\mathbf{x};\mathbf{a}_m),
\end{align}
where  $h(\mathbf{x};\mathbf{a}_m)$, the base learner, is typically a regression tree. The parameters, $\rho_m$ and $\mathbf{a}_m$, are learned from the data, by maximum likelihood, which is equivalent to minimizing the {\em negative log-likelihood loss}
\begin{align}\label{eqn_loss}
L = \sum_{i=1}^N L_i, \hspace{0.4in} L_i = - \sum_{k=0}^{K-1}r_{i,k}  \log p_{i,k}
\end{align}
where $r_{i,k} = 1$ if $y_i = k$ and $r_{i,k} = 0$ otherwise.

For identifiability, $\sum_{k=0}^{K-1}F_{i,k} = 0$, i.e., the \textbf{sum-to-zero} constraint, is routinely adopted
\cite{Article:FHT_AS00,Article:Friedman_AS01,Article:Zhang_JMLR04,Article:Lee_JASA04,Article:Tewari_JMLR07,Article:Zou_AOAS08,Article:Zhu_Adaboost09}.
\subsection{Logitboost}

As described in Alg. \ref{alg_LogitBoost}, \cite{Article:FHT_AS00} builds the additive model (\ref{eqn_F_M}) by a greedy stage-wise  procedure, using a second-order (diagonal) approximation, which requires knowing the first two derivatives of the loss function (\ref{eqn_loss}) with respective to the function values $F_{i,k}$. \cite{Article:FHT_AS00} obtained:
\begin{align}\label{eqn_mart_d1d2}
&\frac{\partial L_i}{\partial F_{i,k}} = - \left(r_{i,k} - p_{i,k}\right),
\hspace{0.5in}
\frac{\partial^2 L_i}{\partial F_{i,k}^2} = p_{i,k}\left(1-p_{i,k}\right).
\end{align}
Those derivatives can be derived by assuming no relations among $F_{i,k}$, $k = 0$ to $K-1$. However, \cite{Article:FHT_AS00} used the ``sum-to-zero'' constraint $\sum_{k=0}^{K-1}F_{i,k} = 0$ throughout the paper and they provided an alternative explanation.   \cite{Article:FHT_AS00}  showed (\ref{eqn_mart_d1d2}) by conditioning on a ``base class'' and noticed the resultant derivatives are independent of the choice of the base.

\begin{algorithm}[h]{\small
0: $r_{i,k} = 1$, if $y_{i} = k$, $r_{i,k} =0$ otherwise.\\
1: $F_{i,k} = 0$,\ \  $p_{i,k} = \frac{1}{K}$, \ \ \ $k = 0$ to  $K-1$, \ $i = 1$ to $N$ \\
2: For $m=1$ to $M$ Do\\
3: \hspace{0.2in}    For $k=0$ to $K-1$, Do\\
4: \hspace{0.4in}    Compute $w_{i,k} = p_{i,k}\left(1-p_{i,k}\right)$.\\
5: \hspace{0.4in}    Compute $z_{i,k} = \frac{r_{i,k} - p_{i,k}}{p_{i,k}\left(1-p_{i,k}\right) }$.\\
6: \hspace{0.4in}   Fit the function $f_{i,k}$  by a weighted  least-square of $z_{i,k}$ \\
: \hspace{0.8in} to $\mathbf{x}_i$ with weights $w_{i,k}$.\\
7:  \hspace{0.4in}  $F_{i,k} = F_{i,k} + \nu \frac{K-1}{K}\left( f_{i,k} - \frac{1}{K}\sum_{k=0}^{K-1}f_{i,k}\right)$\\
8:   \hspace{0.2in} End\\
9: \hspace{0.2in}  $p_{i,k} = \exp(F_{i,k})/\sum_{s=0}^{K-1}\exp(F_{i,s})$\\
10: End
\caption{\small LogitBoost\cite[Alg. 6]{Article:FHT_AS00}. $\nu$ is the shrinkage. }\label{alg_LogitBoost}}
\end{algorithm}

At each stage, {\em logitboost} fits an individual regression function separately for each class. This is analogous to the popular {\em individualized regression} approach in multinomial logistic regression, which is known \cite{Article:Begg_84,Book:Agresti} to result in  loss of statistical efficiency, compared to the full (conditional) maximum likelihood approach.

On the other hand, in order to use trees as base learner, the diagonal approximation appears to be a must, at least from the practical perspective.

\subsection{Adaptive Base Class Boost (ABC-Boost)}

\cite{Proc:ABC_ICML09} derived the derivatives of the loss function (\ref{eqn_loss}) under the sum-to-zero constraint. Without loss of generality, we can assume that class 0 is the base class. For any $k\neq 0$,
\begin{align}\label{eqn_abc_derivatives}
&\frac{\partial L_i}{\partial F_{i,k}}  = \left(r_{i,0} - p_{i,0}\right) - \left(r_{i,k} - p_{i,k}\right),\hspace{0.3in}
&\frac{\partial^2 L_i}{\partial F_{i,k}^2} = p_{i,0}(1-p_{i,0}) + p_{i,k}(1-p_{i,k}) + 2p_{i,0}p_{i,k}.
\end{align}
The base class must be identified at each boosting iteration during training. \cite{Proc:ABC_ICML09} suggested an exhaustive procedure to adaptively find the best base class to minimize the training loss (\ref{eqn_loss}) at each iteration.

\cite{Proc:ABC_ICML09} combined the idea of {\em abc-boost} with {\em mart}. The algorithm, named {\em abc-mart}, achieved good performance in multi-class classification on the  datasets used in \cite{Proc:ABC_ICML09}.

\subsection{Robust LogitBoost}

The {\em mart} paper\cite{Article:Friedman_AS01} and a recent (2008) discussion paper
\cite{Article:FHT_JMLR08} commented that {\em logitboost} (Alg. \ref{alg_LogitBoost}) can be numerically unstable. In fact, the  {\em logitboost} paper\cite{Article:FHT_AS00} suggested some ``crucial implementation protections'' on page 17 of \cite{Article:FHT_AS00}:
\begin{itemize}
\item In Line 5 of Alg. \ref{alg_LogitBoost}, compute the response $z_{i,k}$ by $\frac{1}{p_{i,k}}$ (if $r_{i,k}=1$) or $\frac{-1}{1-p_{i,k}}$ (if $r_{i,k}=0$).
\item Bound the response $|z_{i,k}|$ by $z_{max}\in[2,4]$. The value of $z_{max}$ is not sensitive as long as in $[2,4]$
\end{itemize}
Note that the above operations were applied to each individual sample. The goal was to ensure that the response $|z_{i,k}|$ should  not be too large. On the other hand, we should hope to use larger $|z_{i,k}|$ to better capture the data variation. Therefore, this thresholding operation occurs very frequently and it is expected that part of the useful information is lost.

The next subsection explains that, if implemented carefully, {\em logitboost} is almost identical to {\em mart}. The only difference is the tree-splitting criterion.

\subsection{Tree-Splitting Criterion Using Second-Order Information}\label{sec_split}

Consider $N$ weights $w_i$, and $N$ response values $z_i$, $i=1$ to $N$, which are assumed to be ordered according to the sorted order of the corresponding feature values. The tree-splitting procedure is to find the index $s$, $1\leq s<N$, such that the weighted mean square error (MSE) is reduced the most if split at $s$.  That is, we seek the $s$ to maximize

\begin{align}\notag
Gain(s) = &MSE_{T} - (MSE_{L} + MSE_{R})\\\notag
=&\sum_{i=1}^N (z_i - \bar{z})^2w_i - \left[
\sum_{i=1}^s (z_i - \bar{z}_L)^2w_i + \sum_{i=s+1}^N (z_i - \bar{z}_R)^2w_i\right]
\end{align}
where $\bar{z} = \frac{\sum_{i=1}^N z_iw_i}{\sum_{i=1}^N w_i}$,
$\bar{z}_L = \frac{\sum_{i=1}^s z_iw_i}{\sum_{i=1}^s w_i}$,
$\bar{z}_R = \frac{\sum_{i=s+1}^N z_iw_i}{\sum_{i=s+1}^{N} w_i}$.  After simplification, one can obtain
\begin{align}\notag
Gain(s) =& \frac{\left[\sum_{i=1}^s z_iw_i\right]^2}{\sum_{i=1}^s w_i}+\frac{\left[\sum_{i=s+1}^N z_iw_i\right]^2}{\sum_{i=s+1}^{N} w_i}- \frac{\left[\sum_{i=1}^N z_iw_i\right]^2}{\sum_{i=1}^N w_i}
\end{align}

Plugging in  $w_i = p_{i,k}(1-p_{i,k})$, $z_i = \frac{r_{i,k}-p_{i,k}}{p_{i,k}(1-p_{i,k})}$  yields,
\begin{align}\notag
Gain(s) =&  \frac{\left[\sum_{i=1}^s \left(r_{i,k} - p_{i,k}\right) \right]^2}{\sum_{i=1}^s p_{i,k}(1-p_{i,k})}+\frac{\left[\sum_{i=s+1}^N \left(r_{i,k}- p_{i,k}\right) \right]^2}{\sum_{i=s+1}^{N} p_{i,k}(1-p_{i,k})}- \frac{\left[\sum_{i=1}^N \left(r_{i,k} - p_{i,k}\right) \right]^2}{\sum_{i=1}^N p_{i,k}(1-p_{i,k})}.
\end{align}
Because the computations involve $\sum p_{i,k}(1-p_{i,k})$ as a group, this procedure is actually numerically stable.\\

In comparison, {\em mart}\cite{Article:Friedman_AS01} only used the first order information to construct the trees, i.e.,
\begin{align}\notag
MartGain(s) =&  \left[\sum_{i=1}^s \left(r_{i,k} - p_{i,k}\right) \right]^2+
\left[\sum_{i=s+1}^N \left(r_{i,k} - p_{i,k}\right) \right]^2-
\left[\sum_{i=1}^N \left(r_{i,k} - p_{i,k}\right) \right]^2.
\end{align}

{\scriptsize\begin{algorithm}{\small
1: $F_{i,k} = 0$, $p_{i,k} = \frac{1}{K}$, $k = 0$ to  $K-1$, $i = 1$ to $N$ \\
2: For $m=1$ to $M$ Do\\
3: \hspace{0.1in}    For $k=0$ to $K-1$ Do\\
4:  \hspace{0.2in}  $\left\{R_{j,k,m}\right\}_{j=1}^J = J$-terminal node regression tree from
 $\{r_{i,k} - p_{i,k}, \ \ \mathbf{x}_{i}\}_{i=1}^N$, \\
 :\hspace{1.5in} with weights $p_{i,k}(1-p_{i,k})$ as in Sec. \ref{sec_split}. \\
5:   \hspace{0.2in}  $\beta_{j,k,m} = \frac{K-1}{K}\frac{ \sum_{\mathbf{x}_i \in
  R_{j,k,m}} r_{i,k} - p_{i,k}}{ \sum_{\mathbf{x}_i\in
  R_{j,k,m}}\left(1-p_{i,k}\right)p_{i,k} }$ \\
6:  \hspace{0.2in}  $F_{i,k} = F_{i,k} +
\nu\sum_{j=1}^J\beta_{j,k,m}1_{\mathbf{x}_i\in R_{j,k,m}}$ \\
7:   \hspace{0.1in} End\\
8:\hspace{0.12in} $p_{i,k} = \exp(F_{i,k})/\sum_{s=0}^{K-1}\exp(F_{i,s})$\\
9: End
\caption{\small {\em Robust logitboost}, which is very similar to {\em mart}, except for Line 4.  }
\label{alg_robust_logitboost}}
\end{algorithm}}

Alg. \ref{alg_robust_logitboost} describes {\em robust logitboost} using the tree-splitting criterion  in Sec. \ref{sec_split}. Note that after trees are constructed, the values of the terminal nodes are computed by
\begin{align}\notag
\frac{\sum_{node} z_{i,k} w_{i,k}}{\sum_{node} w_{i,k}} =
\frac{\sum_{node} \left(r_{i,k} - p_{i,k}\right)}{\sum_{node} p_{i,k}(1-p_{i,k})},
\end{align}
which explains Line 5 of Alg. \ref{alg_robust_logitboost}.

\subsection{Adaptive Base Class Logitboost (ABC-LogitBoost)}

The {\em abc-boost} \cite{Proc:ABC_ICML09} algorithm consists of two key components:
\begin{enumerate}
\item Using the {\em sum-to-zero} constraint\cite{Article:FHT_AS00,Article:Friedman_AS01,Article:Zhang_JMLR04,Article:Lee_JASA04,Article:Tewari_JMLR07,Article:Zou_AOAS08,Article:Zhu_Adaboost09} on the loss function,  one can formulate boosting algorithms only for $K-1$ classes, by treating one class as the \textbf{base class}.
\item At each boosting iteration, \textbf{adaptively} select the base class according to the training loss. \cite{Proc:ABC_ICML09} suggested an exhaustive search strategy.
\end{enumerate}

\cite{Proc:ABC_ICML09} combined {\em abc-boost} with {\em mart} to develop {\em abc-mart}. More recently,
\cite{Report:Li_ABC-LogitBoost} developed \textbf{\em abc-logitboost}, the combination of {\em abc-boost} with {\em (robust) logitboost}.

\begin{algorithm}[h]{\small
1: $F_{i,k} = 0$,\ \  $p_{i,k} = \frac{1}{K}$, \ \ \ $k = 0$ to  $K-1$, \ $i = 1$ to $N$ \\
2: For $m=1$ to $M$ Do\\
3: \hspace{0.1in}    For $b=0$ to $K-1$, Do\\
4: \hspace{0.2in}    For $k=0$ to $K-1$, $k\neq b$, Do\\
5:  \hspace{0.3in}  $\left\{R_{j,k,m}\right\}_{j=1}^J = J$-terminal
node regression tree from  $\{-(r_{i,b} - p_{i,b}) +  (r_{i,k} - p_{i,k}), \ \ \mathbf{x}_{i}\}_{i=1}^N$ \\ :\hspace{1.5in} with weights $p_{i,b}(1-p_{i,b})+p_{i,k}(1-p_{i,k})+2p_{i,b}p_{i,k}$, as in Sec. \ref{sec_split}.
 \\
6:   \hspace{0.3in}  $\beta_{j,k,m} = \frac{ \sum_{\mathbf{x}_i \in
  R_{j,k,m}} -(r_{i,b} - p_{i,b}) + (r_{i,k} - p_{i,k})  }{ \sum_{\mathbf{x}_i\in
  R_{j,k,m}} p_{i,b}(1-p_{i,b})+ p_{i,k}\left(1-p_{i,k}\right) + 2p_{i,b}p_{i,k} }$ \\\\
7:  \hspace{0.3in}  $G_{i,k,b} = F_{i,k} +
\nu\sum_{j=1}^J\beta_{j,k,m}1_{\mathbf{x}_i\in R_{j,k,m}}$ \\
8:   \hspace{0.2in} End\\
9: \hspace{0.2in} $G_{i,b,b} = - \sum_{k\neq b} G_{i,k,b}$ \\
10: \hspace{0.2in}  $q_{i,k} = \exp(G_{i,k,b})/\sum_{s=0}^{K-1}\exp(G_{i,s,b})$ \\
11: \hspace{0.2in} $L^{(b)} = -\sum_{i=1}^N \sum_{k=0}^{K-1} r_{i,k}\log\left(q_{i,k}\right)$\\
12: \hspace{0.1in} End\\
13: \hspace{0.1in} $B(m) = \underset{b}{\text{argmin}} \  \ L^{(b)}$\\
14: \hspace{0.1in} $F_{i,k} = G_{i,k,B(m)}$\\
15:\hspace{0.1in}  $p_{i,k} = \exp(F_{i,k})/\sum_{s=0}^{K-1}\exp(F_{i,s})$ \\
16: End}
\caption{{\em Abc-logitboost} using the exhaustive search strategy for the base class, as suggested in \cite{Proc:ABC_ICML09}.  The vector $B$ stores the base class numbers. }
\label{alg_abc-logitboost}
\end{algorithm}%

Alg. \ref{alg_abc-logitboost} presents  {\em abc-logitboost}, using the derivatives in (\ref{eqn_abc_derivatives}) and the same exhaustive search strategy as in {\em abc-mart}. Again, {\em abc-logitboost} differs from {\em abc-mart} only in the tree-splitting procedure (Line 5).

\subsection{Main Parameters}

Alg. \ref{alg_robust_logitboost} and Alg. \ref{alg_abc-logitboost}  have three parameters ($J$, $\nu$ and $M$), to which the performance is in general not very sensitive, as long as they fall in some reasonable range. This is a significant advantage in practice.

The number of terminal nodes, $J$,  determines the capacity of the base learner.  \cite{Article:Friedman_AS01} suggested $J=6$. \cite{Article:FHT_AS00,Article:Zou_AOAS08} commented that $J> 10$ is unlikely. In our experience, for large datasets (or moderate datasets in high-dimensions), $J=20$ is often a reasonable choice; also see \cite{Proc:McRank_NIPS07} for more examples.

The shrinkage, $\nu$, should be large enough to make sufficient progress at each step and  small enough to avoid over-fitting.  \cite{Article:Friedman_AS01} suggested $\nu\leq 0.1$. Normally, $\nu=0.1$ is  used.

The number of boosting iterations, $M$, is largely determined  by the affordable computing time. A commonly-regarded merit of boosting is that, on many datasets, over-fitting can be largely avoided for reasonable $J$, and $\nu$.

\section{Datasets}

Table \ref{tab_data} lists the datasets used in our study. \cite{Proc:ABC_ICML09,Report:Li_ABC-LogitBoost} provided  experiments on several other (small) datasets.

\begin{table}[h]
\caption{Datasets}
\begin{center}{\small
\begin{tabular}{l r r r r}
\hline \hline
dataset &$K$ & \# training & \# test &\# features\\
\hline
Covertype290k &7 & 290506 & 290506 & 54\\
Covertype145k &7 & 145253 & 290506 & 54\\
Poker525k &10 & 525010 &500000 &25\\
Poker275k &10 & 275010 &500000 &25\\
Poker150k &10 & 150010 &500000 &25\\
Poker100k &10 & 100010 &500000 &25\\
Poker25kT1 &10 & 25010 &500000 &25\\
Poker25kT2 &10 & 25010 &500000 &25\\
Mnist10k &10 &10000 &60000&784\\
M-Basic &10 &12000 &50000&784\\
M-Rotate &10 &12000 &50000&784\\
M-Image &10 &12000 &50000&784\\
M-Rand &10 &12000 &50000&784\\
M-RotImg &10 &12000 &50000&784\\
M-Noise1 &10 &10000 &2000&784\\
M-Noise2 &10 &10000 &2000&784\\
M-Noise3 &10 &10000 &2000&784\\
M-Noise4 &10 &10000 &2000&784\\
M-Noise5 &10 &10000 &2000&784\\
M-Noise6 &10 &10000 &2000&784\\
Letter15k &26 &15000 &5000 &16\\
Letter4k &26   & 4000   &16000 &16\\
Letter2k &26   & 2000   &18000 &16\\
\hline\hline
\end{tabular}
}
\end{center}
\label{tab_data}
\end{table}

\subsection{Covertype}

The original UCI {\em Covertype} dataset is fairly large, with $581012$ samples. To generate {\em Covertype290k}, we randomly split the original data into halves, one half for training and another half for testing. For {\em Covertype145k}, we randomly select one half from the training set of {\em Covertype290k} and still keep the  test set.

\subsection{Poker}

The UCI {\em Poker} dataset originally used only $25010$ samples for training and $1000000$ samples for testing. Since the test set is very large, we randomly divide it equally into two parts (I and II). {\em Poker25kT1} uses the original training set for training and Part I of the original test set for testing. {\em Poker25kT2} uses the original training set for training and Part II of the original test set for testing. This way, {\em Poker25kT1} can use the test set of {\em Poker25kT2} for validation, and {\em Poker25kT2} can use the test set of {\em Poker25kT1} for validation. As the two test sets are still very large, this treatment will provide reliable results.

Since the original training set (about $25k$) is too small compared to the size of the test set, we enlarge the training set to form {\em Poker525k}, {\em Poker275k},  {\em Poker150k}, and {\em Poker100k}. All four enlarged training datasets use the same test set as {\em Pokere25kT2} (i.e., Part II of the original test set). The training set of {\em Poker525k} contains the original ($25010$) training set plus Part I of the original test set. Similarly, the training set of {\em Poker275k} / {\em Poker150k}  / {\em Poker100k} contains the original training set plus 250k/125k/75k samples from Part I of the original test set.

The original {\em Poker} dataset provides 10 features, 5 ``suit'' features and 5 ''rank'' features. While the ``ranks'' are naturally ordinal, it appears reasonable to treat ``suits'' as nominal features. By private communications, R. Cattral, the  donor of the {\em Poker} data, suggested us to treat the ``suits'' as nominal. C.J. Lin also kindly told us that the performance of SVM was not affected whether ``suits'' are treated nominal or ordinal. In our experiments, we choose to use ``suits'' as nominal feature; and hence the total number of features becomes 25 after expanding each ``suite'' feature with 4 binary features.

\subsection{Mnist}

While the original {\em Mnist} dataset is extremely popular, this dataset is known to be too easy\cite{Proc:Larochelle_ICML07}. Originally, {\em Mnist} used 60000 samples for training and 10000 samples for testing.


{\em Mnist10k} uses the original (10000) test set for training and the original (60000) training set for testing. This creates a more challenging task.

\subsection{Mnist with Many Variations}

\cite{Proc:Larochelle_ICML07}
({\scriptsize\url{www.iro.umontreal.ca/~lisa/twiki/bin/view.cgi/Public/DeepVsShallowComparisonICML2007}}) created a variety of much more difficult datasets by adding various background (correlated) noise, background images,  rotations, etc, to the original {\em Mnist} dataset. We shortened the notations of the generated datasets to be {\em M-Basic}, {\em M-Rotate}, {\em M-Image}, {\em M-Rand}, {\em M-RotImg}, and {\em M-Noise1}, {\em M-Noise2} to {\em M-Noise6}.

By private communications with D. Erhan, one of the authors of \cite{Proc:Larochelle_ICML07}, we learn that the sizes of the training sets actually vary depending on the learning algorithms. For some methods such as SVM, they retrained the algorithms using all 120000 training samples after choosing the best parameters; and for other methods, they used 10000 samples for training. In our experiments, we use 12000 training samples for {\em M-Basic}, {\em M-Rotate}, {\em M-Image}, {\em M-Rand} and {\em M-RotImg}; and we use 10000 training samples for {\em M-Noise1} to {\em M-Noise6}.

Note that the datasets {\em M-Noise1} to {\em M-Noise6}  have merely 2000 test samples each. By private communications with D. Erhan, we understand this was because \cite{Proc:Larochelle_ICML07} did not mean to compare the statistical significance of the test errors for those six datasets.

\subsection{Letter}

The UCI {\em Letter} dataset has in total 20000 samples. In our experiments, {\em Letter4k} ({\em Letter2k}) use the last 4000 (2000) samples for training and the rest for testing. The purpose is to demonstrate the performance of the algorithms using only small training sets.

We also include {\em Letter15k}, which is one of the standard partitions of the {\em Letter} dataset, by using 15000 samples for training and 5000 samples for testing.

\section{Summary of Experiment Results}

We simply use {\em logitboost} (or even {\em logit} in the plots) to denote {{\em robust logitboost}.

Table \ref{tab_summary} summarizes the test mis-classification errors. For all datasets except {\em Poker25kT1} and {\em Poker25kT2}, we report the test errors with the tree size $J$=20 and shrinkage $\nu=0.1$. For {\em Poker25kT1} and {\em Poker25kT2}, we use $J=6$ and $\nu=0.1$. We report more detailed experiment results in Sec. \ref{sec_detailed_exp}.

For {\em Covertype290k}, {\em Poker525k}, {\em Poker275k}, {\em Poker150k}, and {\em Poker100k}, as they are fairly large, we only train $M=5000$ boosting iterations. For all other datasets, we always train $M=10000$ iterations or terminate when the training loss (\ref{eqn_loss}) is close to the machine accuracy. Since we do not notice obvious over-fitting on those datasets, we simply report the test errors at the last iterations.

\begin{table}[h]
\caption{Summary of test mis-classification errors.   }
\begin{center}{\small
\begin{tabular}{l r r r r r}
\hline \hline
Dataset &\textbf{mart} & \textbf{abc-mart} &\textbf{logitboost} & \textbf{abc-logitboost} &\# test\\
\hline
Covertype290k &11350 &10454  &10765  &9727 &290506 \\
Covertype145k &15767       &14665       &14928       &13986 &290506 \\
Poker525k &7061  &2424  &2704 &1736 &500000\\
Poker275k &15404        &3679        &6533        &2727 &500000\\
Poker150k &22289 &12340 &16163  &5104 &500000\\
Poker100k &27871       &21293       &25715       &13707 &500000\\
Poker25kT1 &43575  &34879  &46789 &37345 &500000\\
Poker25kT2 &42935  &34326  &46600 &36731 &500000\\
Mnist10k  &2815  &2440   &2381   &2102 &60000\\
M-Basic   &2058  &1843   &1723   &1602 &50000\\
M-Rotate  &7674 &6634   &6813   &5959 &50000\\
M-Image   &5821  &4727   &4703   &4268&50000\\
M-Rand   &6577  &5300   &5020   &4725  &50000\\
M-RotImg   &24912  &23072   &22962   &22343 &50000\\
M-Noise1 &305   &245   &267   &234 &2000\\
M-Noise2 &325   &262   &270   &237 &2000\\
M-Noise3 &310   &264   &277   &238 &2000\\
M-Noise4 &308   &243   &256   &238 &2000\\
M-Noise5 &294   &244   &242   &227 &2000\\
M-Noise6 &279 &224   &226   &201 &2000\\
Letter15k &155   &125   &139   &109 &5000\\
Letter4k  &1370 & 1149 & 1252 & 1055 &16000\\
Letter2k  &2482 & 2220 & 2309 & 2034 &18000\\
\hline\hline
\end{tabular}
}
\end{center}
\label{tab_summary}
\end{table}

\clearpage

\subsection{$P$-Values}\label{sec_P-Value}

Table \ref{tab_summary_P} summarizes the following four types of $P$-values:
\begin{itemize}
\item $P1$: for testing if {\em abc-mart} has significantly lower \textbf{\em error rates} than  {\em mart}.
\item $P2$: for testing if {\em (robust) logitboost}  has significantly lower  error rates than  {\em mart}.
\item $P3$: for testing if {\em abc-logitboost}  has significantly lower error rates than  {\em abc-mart}.
\item $P4$: for testing if {\em abc-logitboost}  has significantly lower error rates than  {\em (robust) logitboost}.
\end{itemize}

The $P$-values are computed using  binomial distributions  and normal approximations. Recall, if a random variable $z\sim Binomial(n,p)$, then the probability parameter $p$ can be estimated by $\hat{p} = \frac{z}{n}$, and the variance of $\hat{p}$ can be estimated by $\hat{p}(1-\hat{p})/n$. The $P$-values  can then be computed using  normal approximation of binomial distributions.

Note that the test sets for {\em M-Noise1} to {\em M-Noise6} are very small because \cite{Proc:Larochelle_ICML07} originally did not intend to compare the statistical significance on those six datasets. We compute their $P$-values anyway.

\begin{table}[h]
\caption{Summary of test $P$-Values.   }
\begin{center}{\small
\begin{tabular}{l r r r r}
\hline \hline
Dataset &$P1$ & $P2$ &$P3$ &$P4$\\
\hline
Covertype290k &$3\times10^{-10}$& $3\times 10^{-5}$  &$9\times 10^{-8}$ &$8\times10^{-14}$ \\
Covertype145k &$4\times10^{-11}$& $4\times 10^{-7}$  &$2\times 10^{-5}$ &$7\times10^{-9}$ \\
Poker525k & 0 & 0 & 0 &0\\
Poker275k & 0 & 0 & 0 &0\\
Poker150k & 0 & 0 & 0 &0\\
Poker100k & 0 & 0 & 0 &0\\
Poker25kT1 & 0 & ---- & ---- &0\\
Poker25kT2 & 0 & ---- & ---- &0\\
Mnist10k  &$5\times10^{-8}$& $3\times 10^{-10}$ &$1\times10^{-7}$ & $1\times10^{-5}$\\
M-Basic  &$2\times10^{-4}$ & $1\times 10^{-8}$ &$1\times 10^{-5}$ &0.0164\\
M-Rotate  &$0$ & $5\times10^{-15}$ &$6\times 10^{-11}$ &$3\times10^{-16}$\\
M-Image  &$0$ & $0$ &$2\times10^{-7}$ &$7\times10^{-7}$\\
M-Rand  &$0$ & $0$ &$7\times10^{-10}$ &$8\times10^{-4}$\\
M-RotImg  &$0$ & $0$ &$2\times10^{-6}$ &$4\times10^{-5}$\\
M-Noise1 &$0.0029$ &$0.0430$ &$0.2961$ &0.0574\\
M-Noise2 &0.0024    &0.0072    &0.1158    &0.0583\\
M-Noise3 &0.0190    &0.0701    &0.1073    &0.0327\\
M-Noise4 &0.0014    &0.0090    &0.4040    &0.1935\\
M-Noise5 &0.0102    &0.0079    &0.2021    &0.2305\\
M-Noise6 &0.0043  &0.0058 &0.1189 &0.1002\\
Letter15k &0.0345    &0.1718    &0.1449    &0.0268\\
Letter4k  &$2\times10^{-6}$ & $0.008$ &0.019 &$1\times10^{-5}$\\
Letter2k  &$2\times10^{-5}$ & $0.003$ &0.001 &$4\times10^{-6}$\\

\hline\hline
\end{tabular}
}
\end{center}
\label{tab_summary_P}
\end{table}

The results demonstrate that {\em abc-logitboost} and {\em abc-mart} considerably outperform {\em logitboost} and {\em mart}, respectively. In addition, except for {\em Poker25kT1} and {\em Poker25kT2}, we observe that {\em abc-logitboost} outperforms {\em abc-mart}, and {\em logitboost} outperforms {\em mart}.

\subsection{Comparisons with SVM and Deep Learning}

For UCI {\em Poker}, we know that SVM could only achieve an error rate of about $40\%$ (by private communications with C.J. Lin). In comparison, all four algorithms, {\em mart}, {\em abc-mart}, {\em (robust) logitboost}, and {\em abc-logitboost}, could achieve much smaller error rates (i.e., $<10\%$) on {\em Poker25kT1} and {\em Poker25kT2}. \\

Figure \ref{fig_corr} provides the comparisons on the six (correlated) noise datasets: {\em M-Noise1} to {\em M-Noise6}. Table \ref{tab_Deep} compares the error rates on {\em M-Basic}, {\em M-Rotate}, {\em M-Image}, {\em M-Rand}, and {\em M-RotImg}.

\begin{figure}[h]
\begin{center}\mbox{
\includegraphics[width = 2.2in]{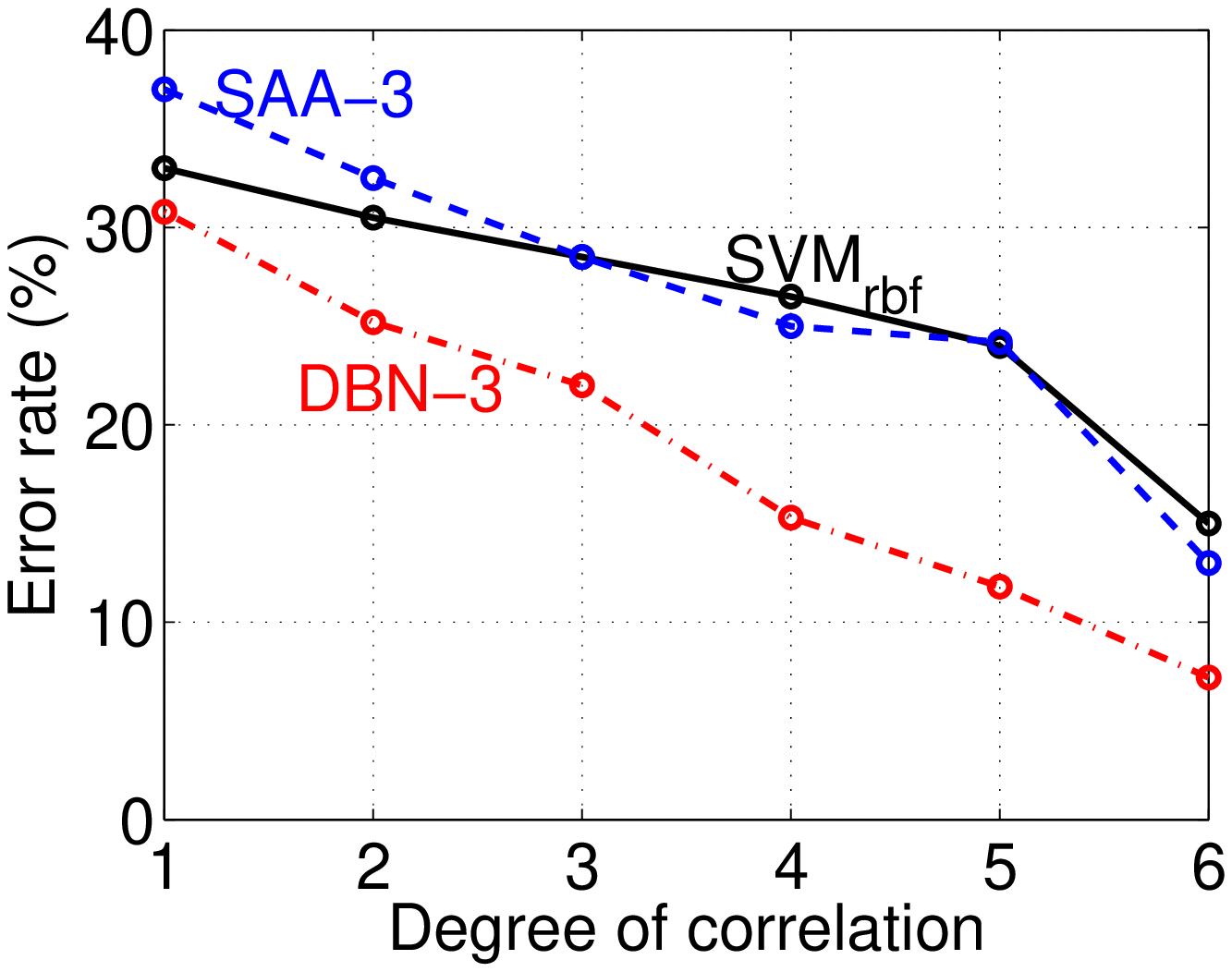}
\includegraphics[width = 2.2in]{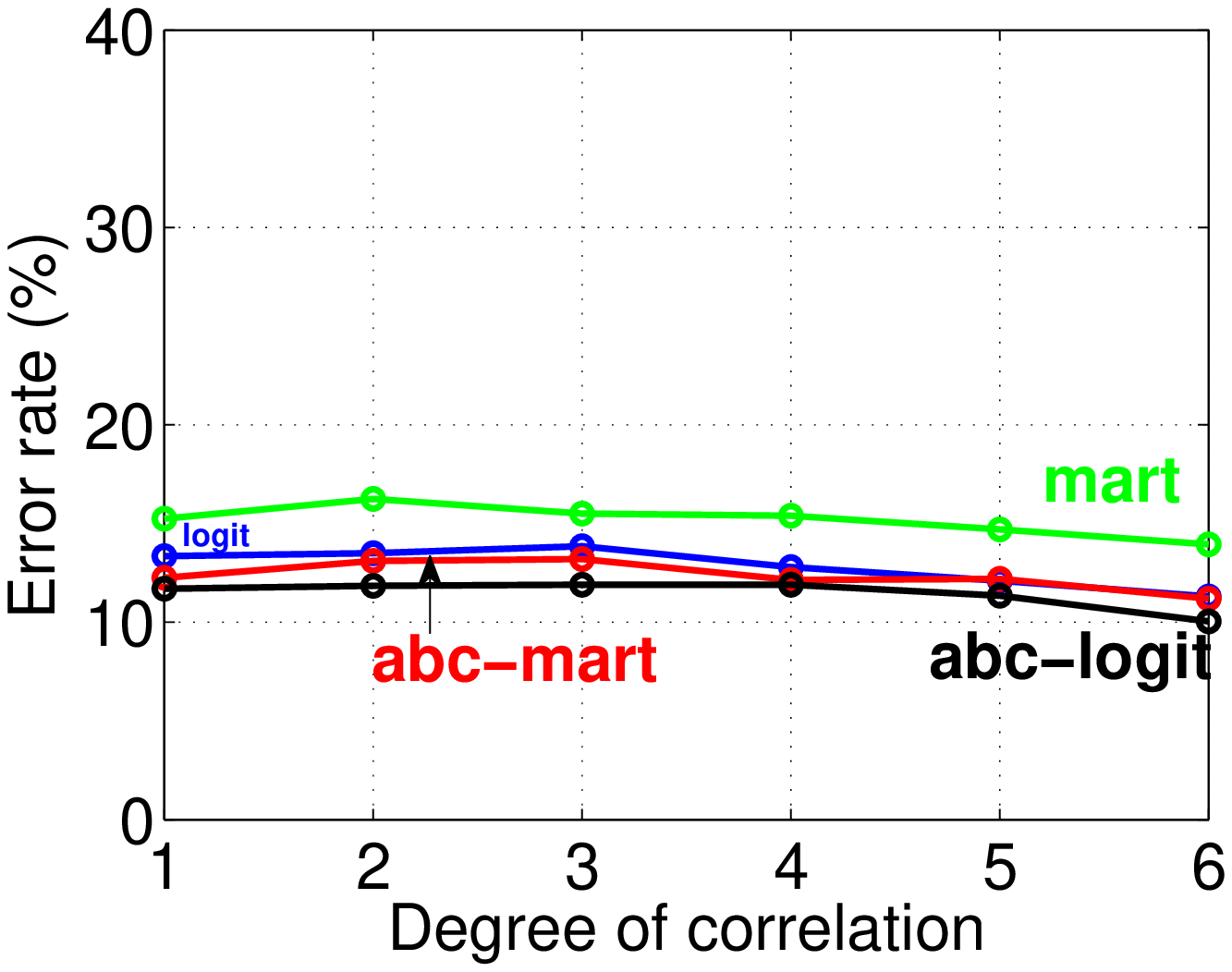}
\includegraphics[width = 2.2in]{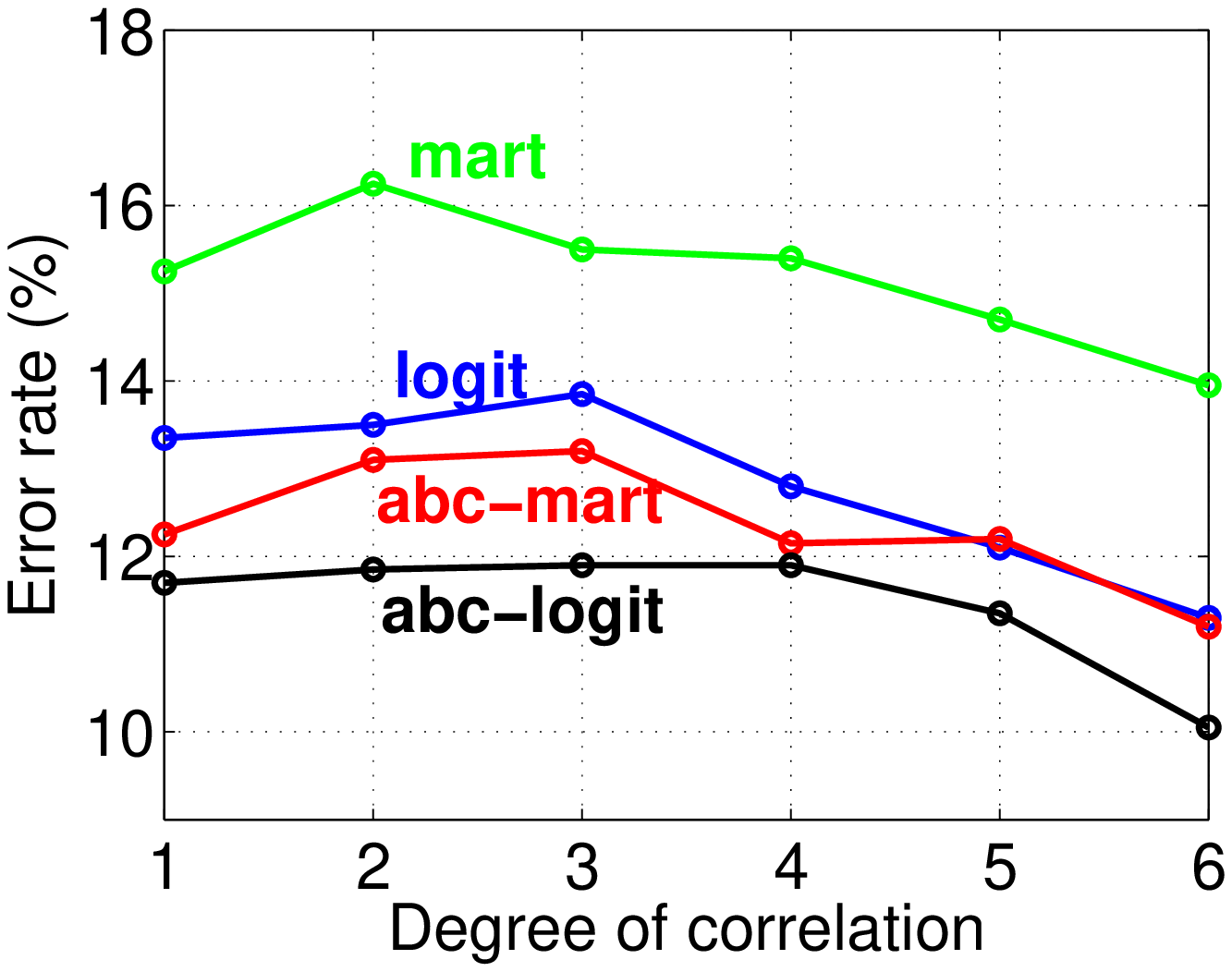}
}
\end{center}\vspace{-0.15in}
\caption{Six datasets: \textbf{\em M-Noise1}  to \textbf{\em M-Noise6}. Left panel: Error rates of SVM and deep learning \cite{Proc:Larochelle_ICML07}.  Middle and right panels: Errors rates of four boosting algorithms. X-axis: degree of correlation from high to low; the values 1 to 6 correspond to the datasets {\em M-Noise1} to {\em M-Noise6}. }\label{fig_corr}
\end{figure}

\begin{table}[h]
\caption{Summary of error rates of various algorithms on the modified {\em Mnist} dataset\cite{Proc:Larochelle_ICML07}. }
\begin{center}{\small
\begin{tabular}{l r r r r r}
\hline \hline
 &M-Basic & M-Rotate &M-Image &M-Rand & M-RotImg\\\hline
SVM-RBF &$\mathbf{3.05\%}$  &$11.11\%$ &$22.61\%$ &$14.58\%$ &$55.18\%$\\
SVM-POLY &$3.69\%$ &$15.42\%$ &$24.01\%$ &$16.62\%$ &$56.41\%$\\
NNET &$4.69\%$ &$18.11\%$ &$27.41\%$ &$20.04\%$ &$62.16\%$\\
DBN-3 &$3.11\%$ &$\mathbf{10.30\%}$ &$16.31\%$ &$\mathbf{6.73\%}$ &$47.39\%$\\
SAA-3 &$3.46\%$ &$\mathbf{10.30\%}$ &$23.00\%$ &$11.28\%$ &$51.93\%$\\
DBN-1 &$3.94\%$ &$14.69\%$ &$16.15\%$ &$9.80\%$ &$52.21\%$\\\\
\textbf{mart} & $4.12\%$ &$15.35\%$ &$11.64\%$ & $13.15\%$ &$49.82\%$\\
\textbf{abc-mart} &$3.69\%$ &$13.27\%$ &$9.45\%$ & $10.60\%$ &$46.14\%$\\
\textbf{logitboost} &$3.45\%$ &$13.63\%$ & $9.41\%$ &$10.04\%$&$45.92\%$\\
\textbf{abc-logitboost} &$3.20\%$ &$11.92\%$ & $\textbf{8.54\%}$ &$9.45\%$&$\mathbf{44.69\%}$\\
\hline\hline
\end{tabular}
}
\end{center}
\label{tab_Deep}
\end{table}

\clearpage

\subsection{Performance vs. Boosting Iterations}

Figure \ref{fig_Cover_Poker_train} presents the training loss, i.e., Eq. (\ref{eqn_loss}), on {\em Covertype290k} and {\em Poker525k}, for all boosting iterations. Figures \ref{fig_Mnist_Letter} and \ref{fig_Cover_Poker} provide the test mis-classification errors on {\em Covertype}, {\em Poker}, {\em Mnist10k}, and {\em Letter}.

\begin{figure}[h]
\begin{center}
\mbox{
\includegraphics[width=2.2in]{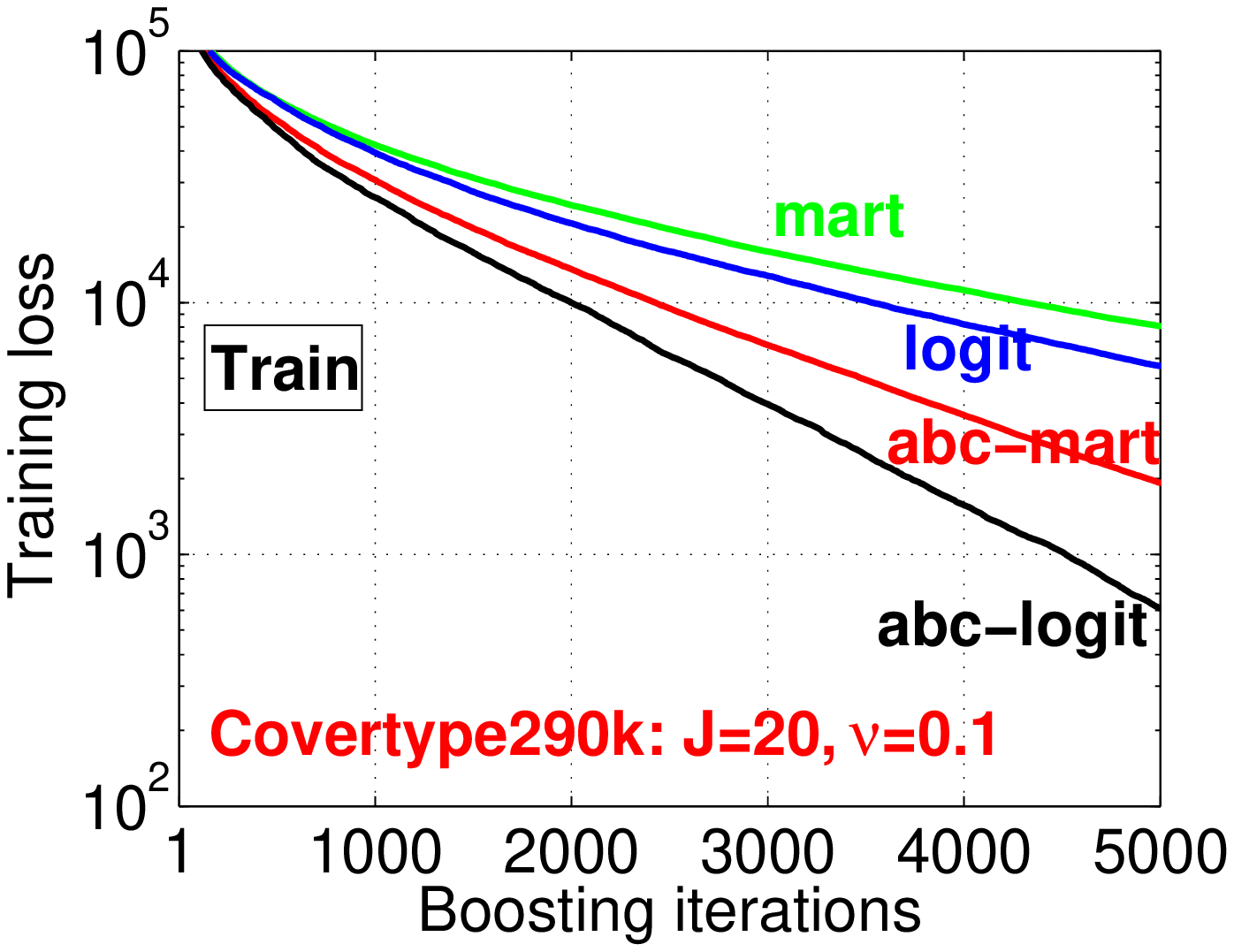}\hspace{0.2in}
\includegraphics[width=2.2in]{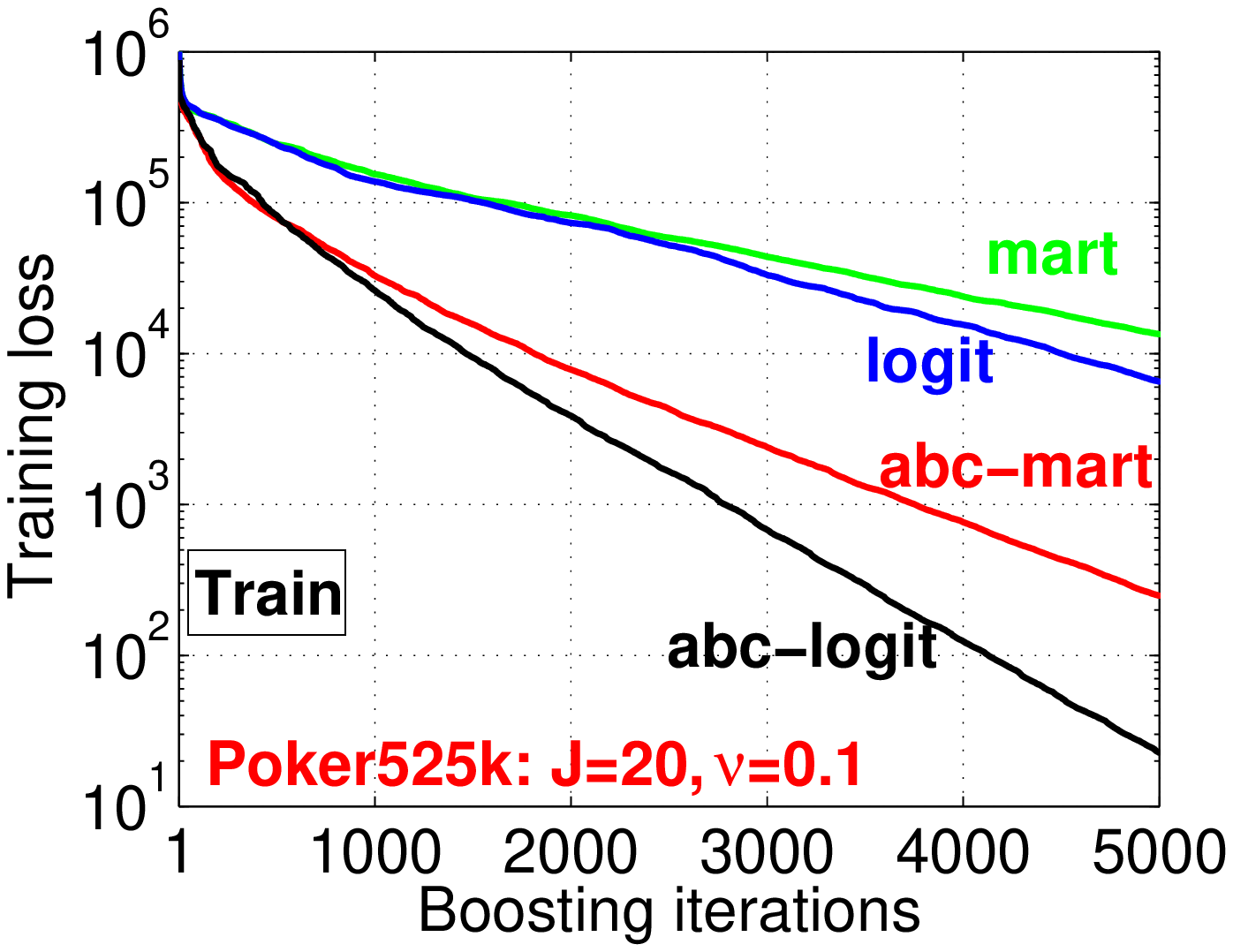}
}
\end{center}
\vspace{-0.1in}
\caption{Training loss, Eq. (\ref{eqn_loss}), on \textbf{\em Covertype290k} and \textbf{\em Poker525k}. }\label{fig_Cover_Poker_train}
\end{figure}

\begin{figure}[h]
\begin{center}
\mbox{
\includegraphics[width=2.2in]{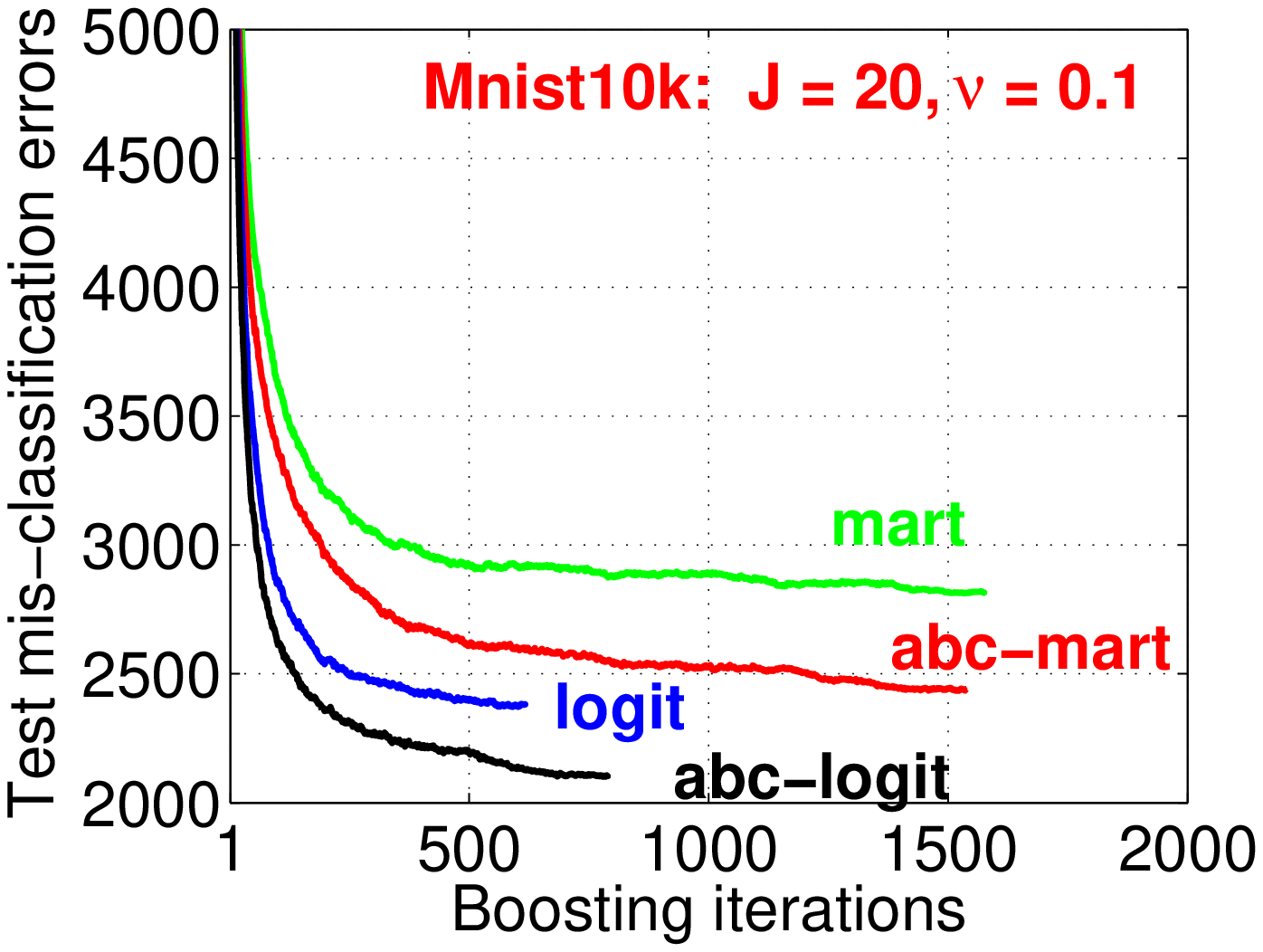}\hspace{0.2in}
\includegraphics[width=2.2in]{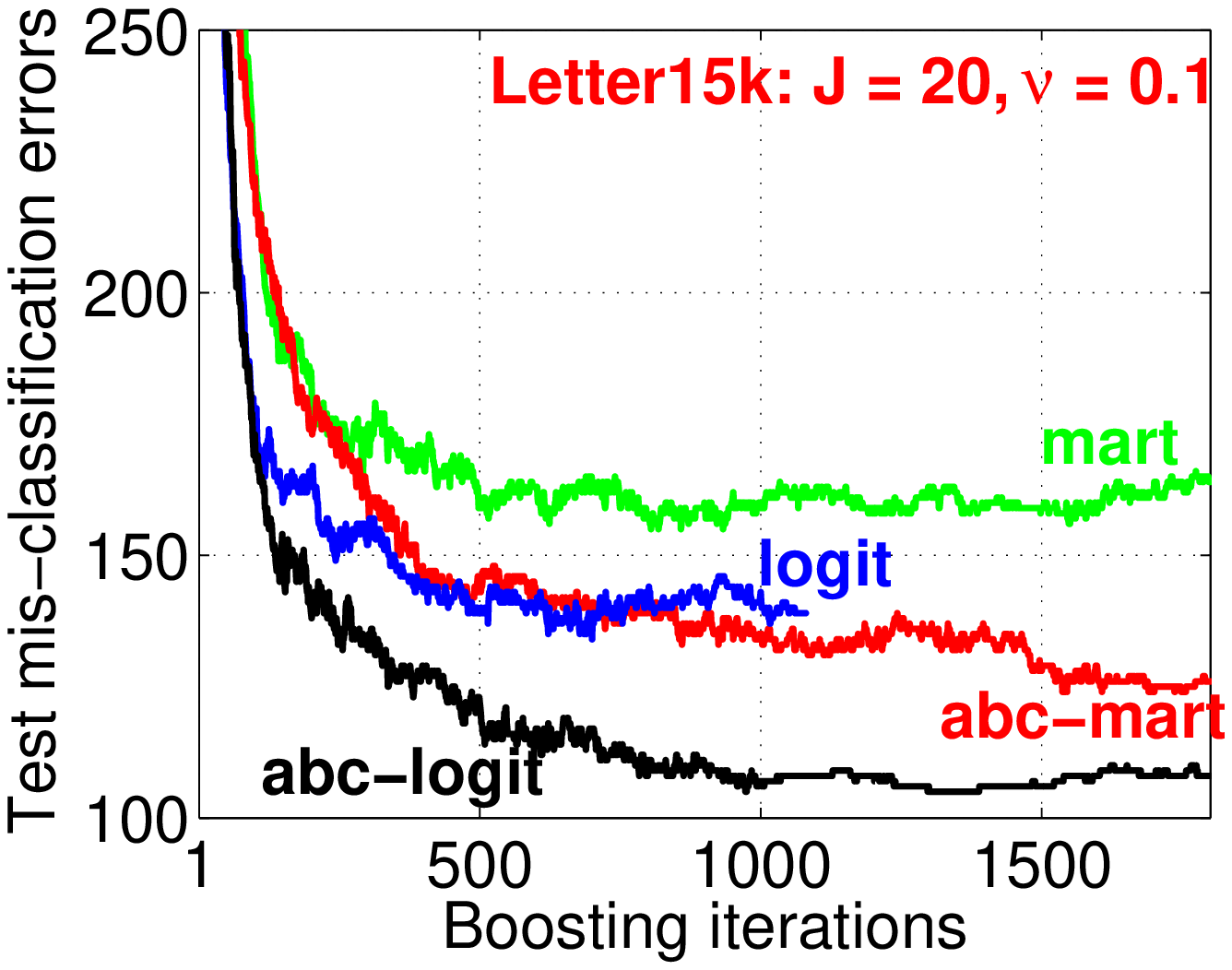}
}
\mbox{
\includegraphics[width=2.2in]{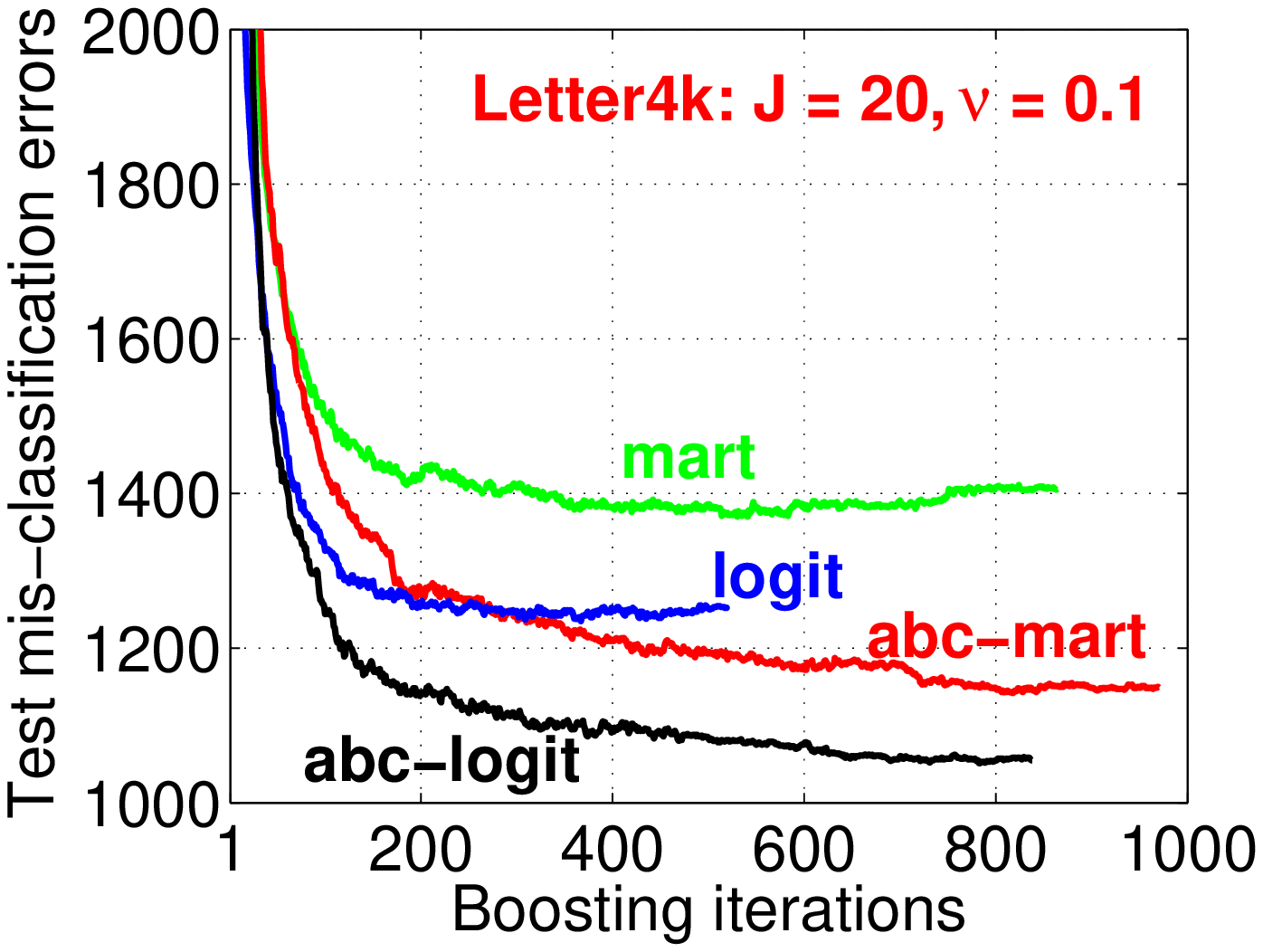}\hspace{0.2in}
\includegraphics[width=2.2in]{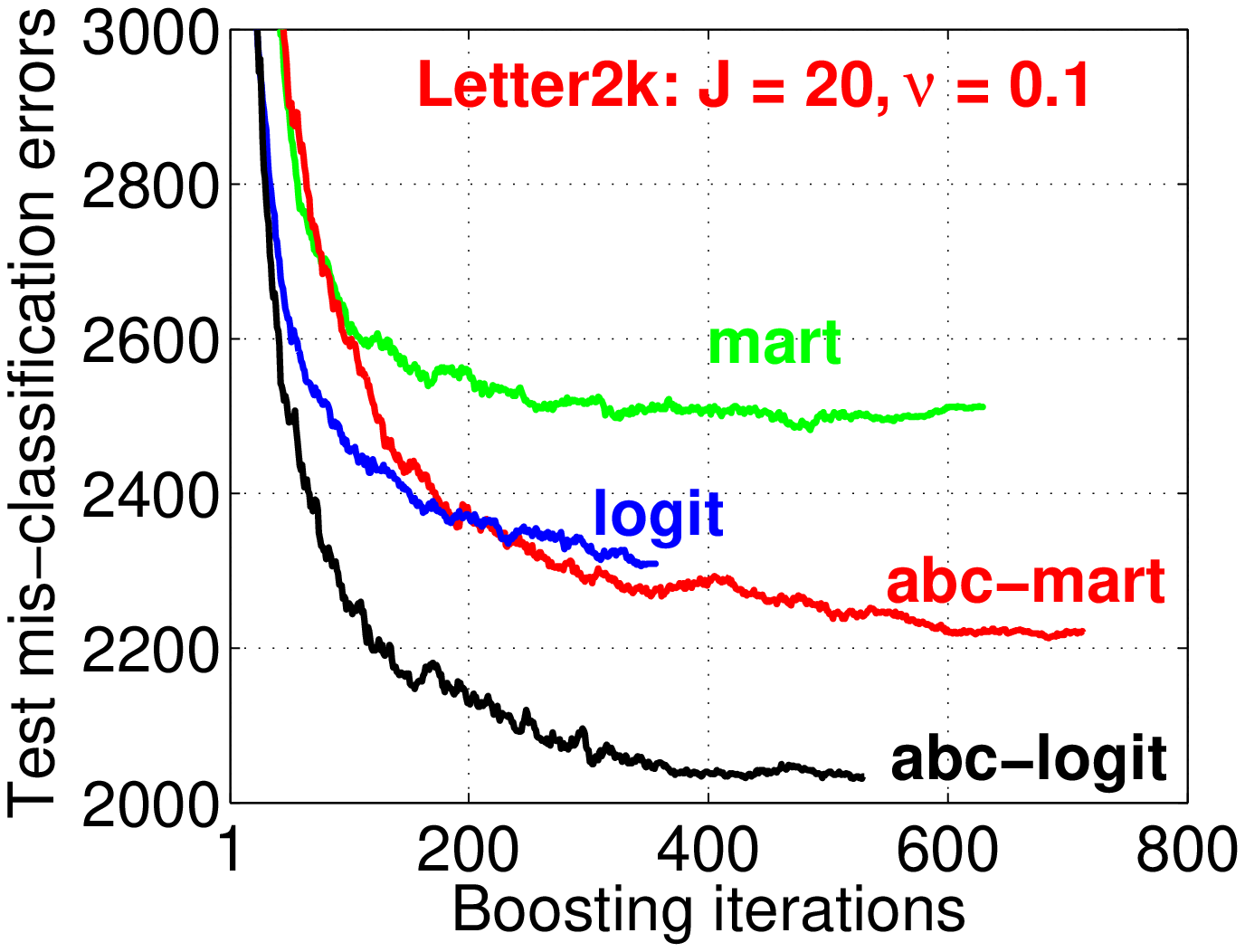}
}
\end{center}
\vspace{-0.1in}
\caption{Test mis-classification errors on \textbf{\em Mnist10k}, \textbf{\em Letter15k}, \textbf{\em Letter4k}, and \textbf{\em Letter2k}.  }\label{fig_Mnist_Letter}
\end{figure}

\begin{figure}[h!]
\begin{center}

\mbox{
\includegraphics[width=2.2in]{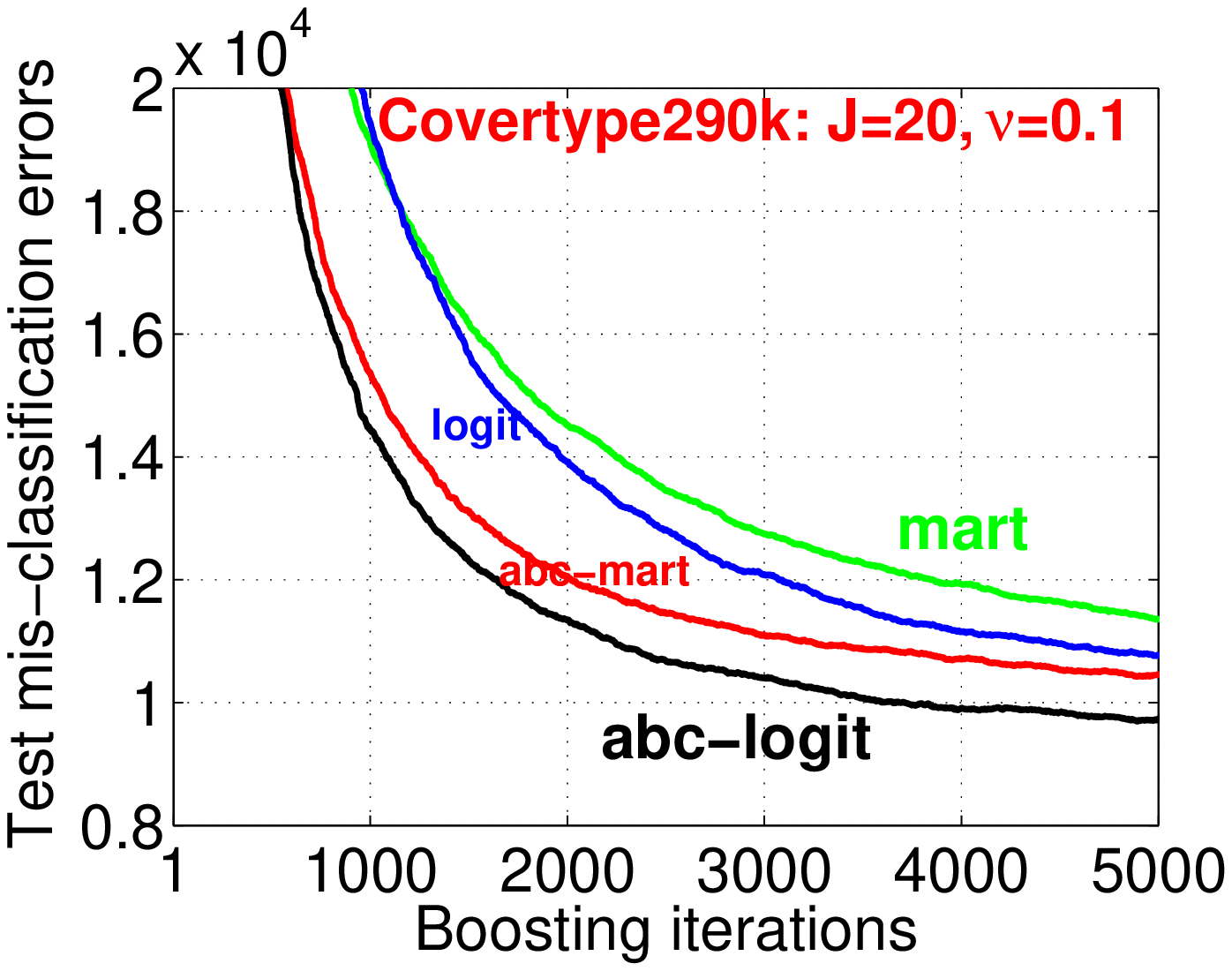}\hspace{0.2in}
\includegraphics[width=2.2in]{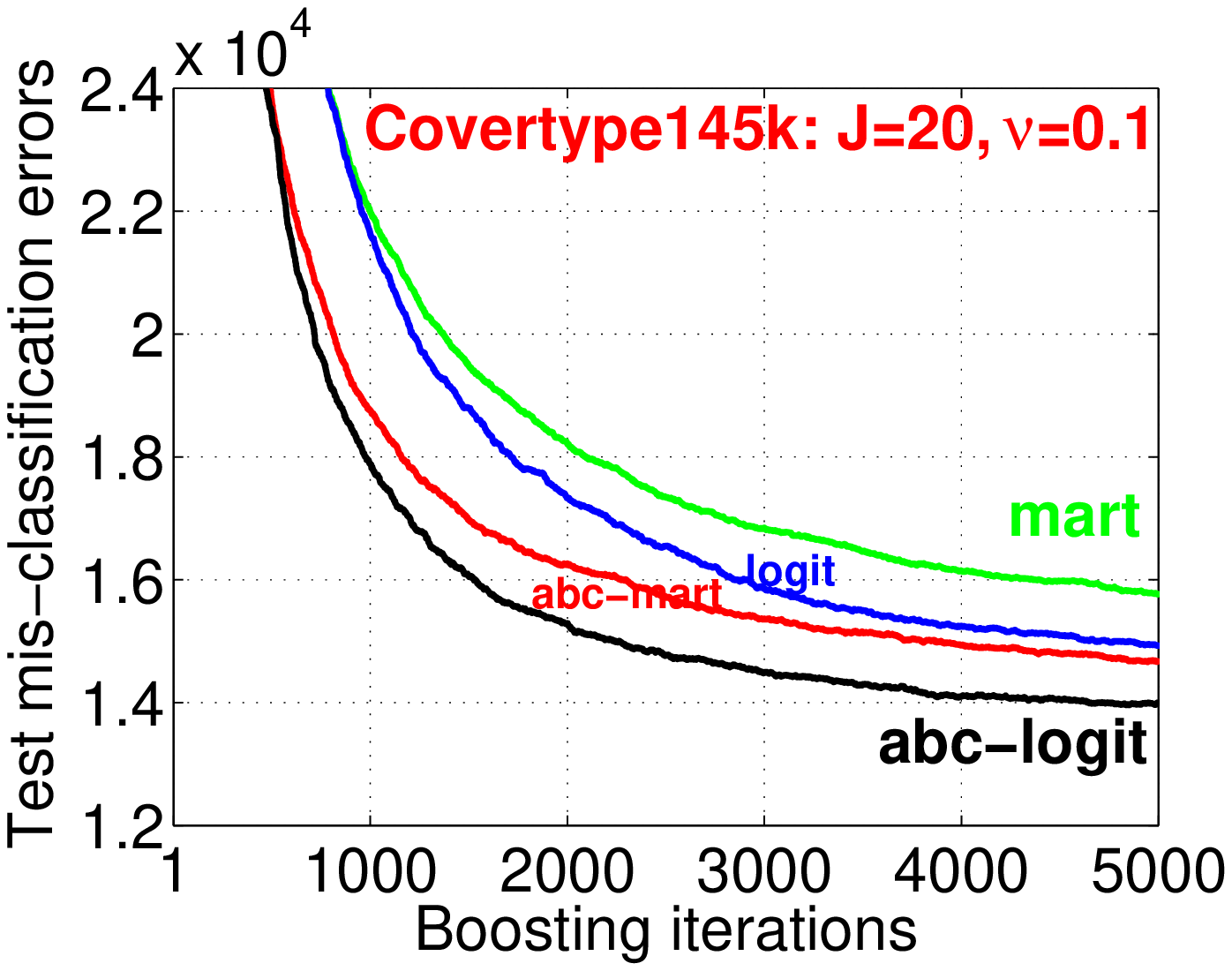}
}

\mbox{
\includegraphics[width=2.2in]{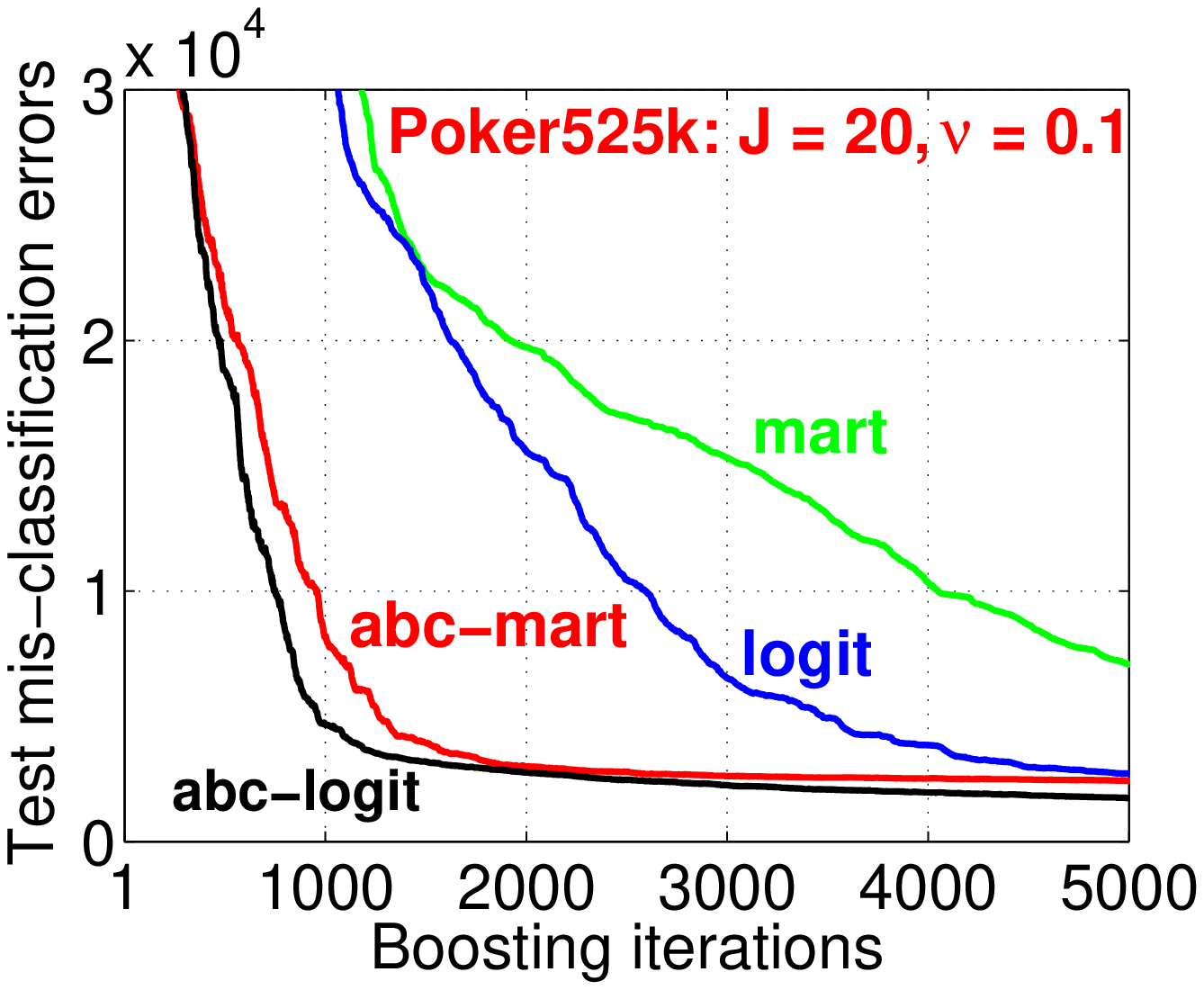}\hspace{0.2in}
\includegraphics[width=2.2in]{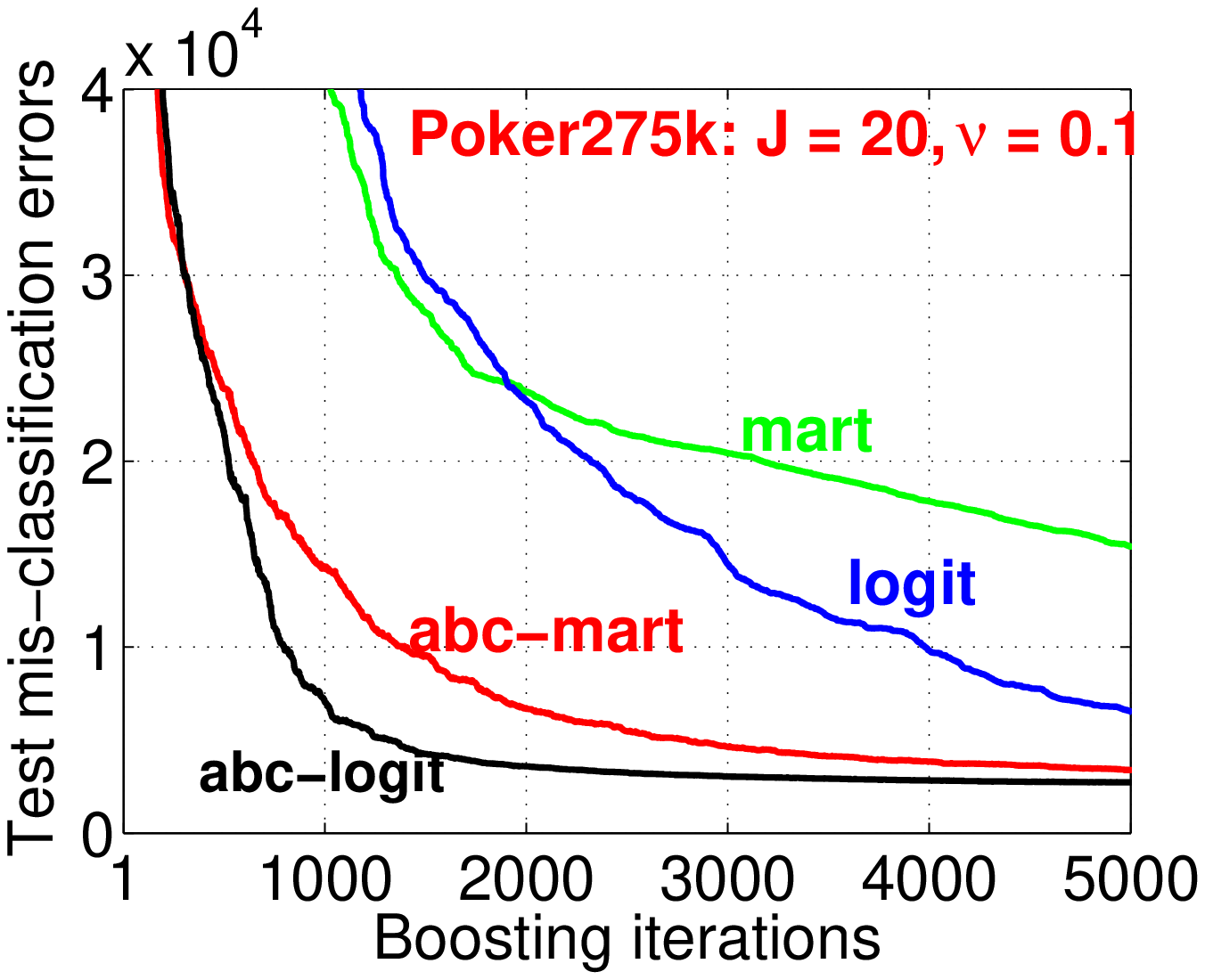}
}
\mbox{
\includegraphics[width=2.2in]{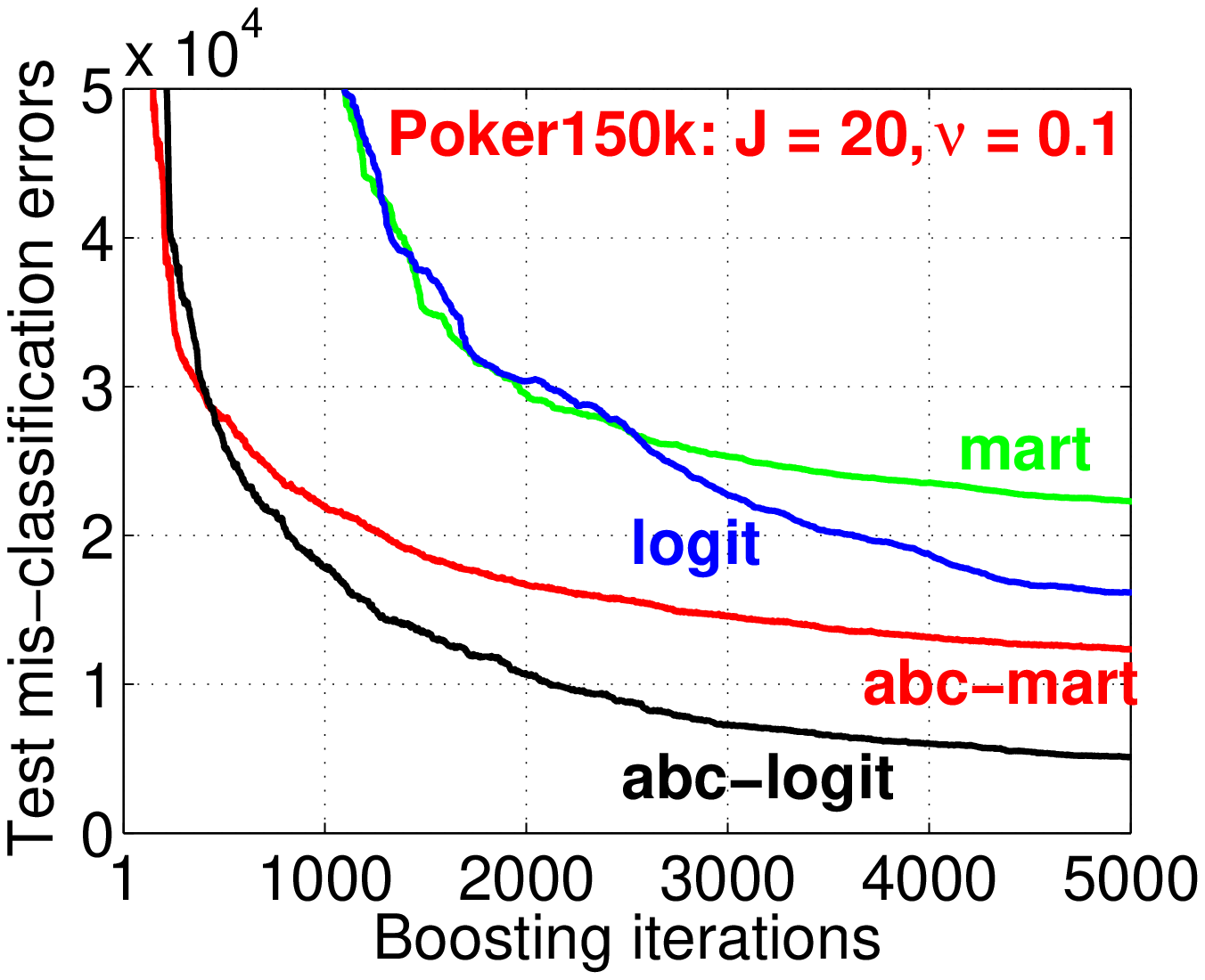}\hspace{0.2in}
\includegraphics[width=2.2in]{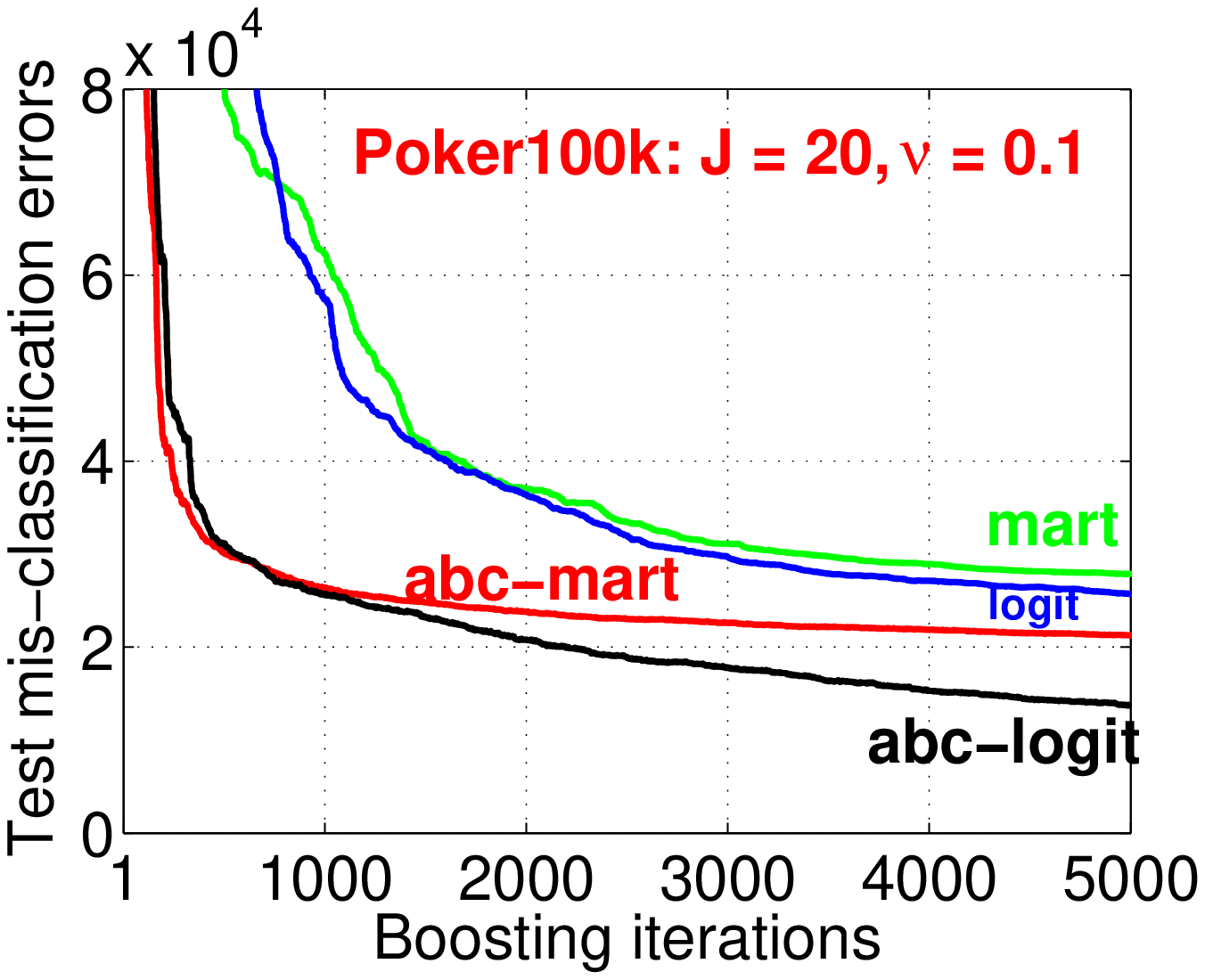}
}

\mbox{
\includegraphics[width=2.2in]{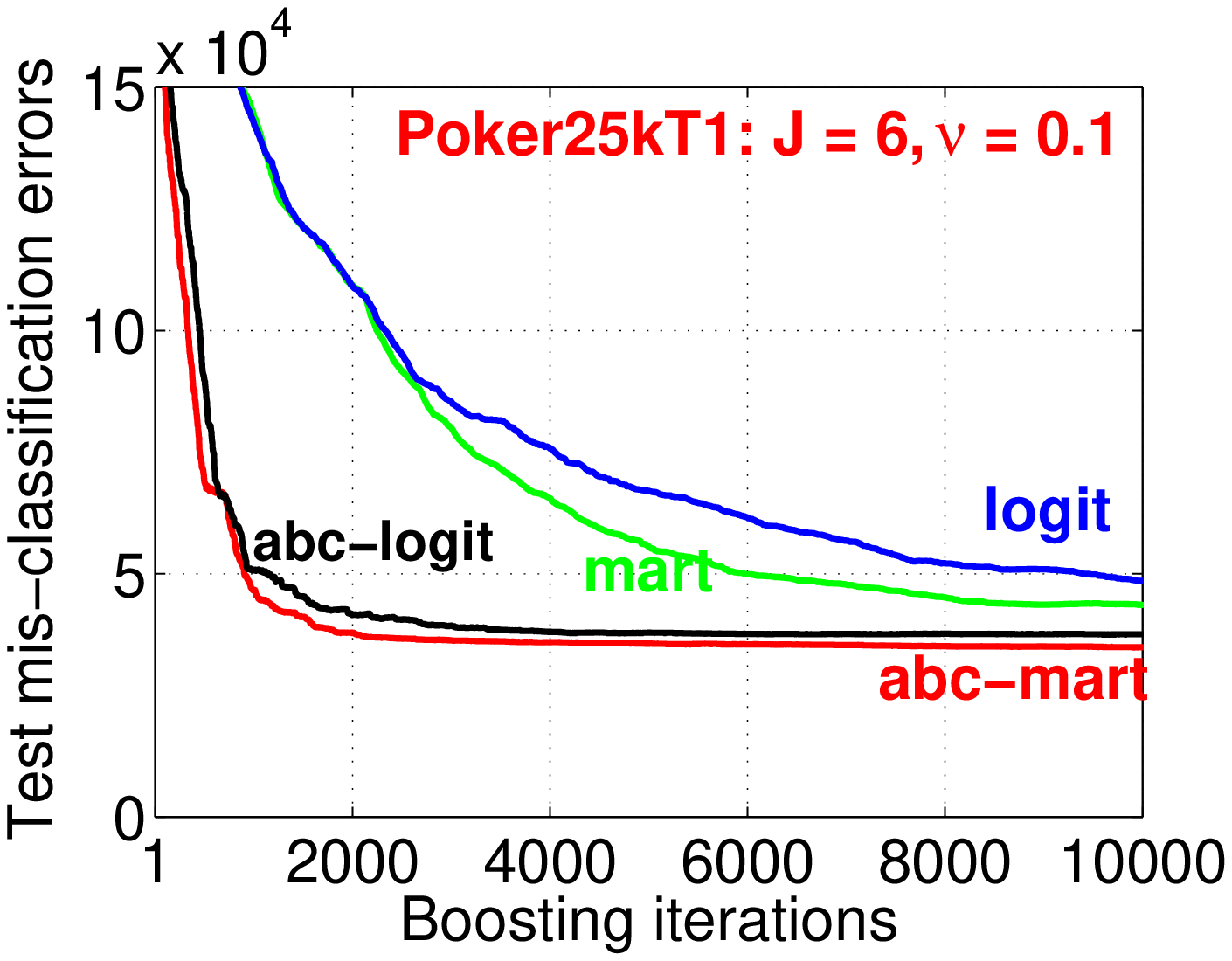}\hspace{0.2in}
\includegraphics[width=2.2in]{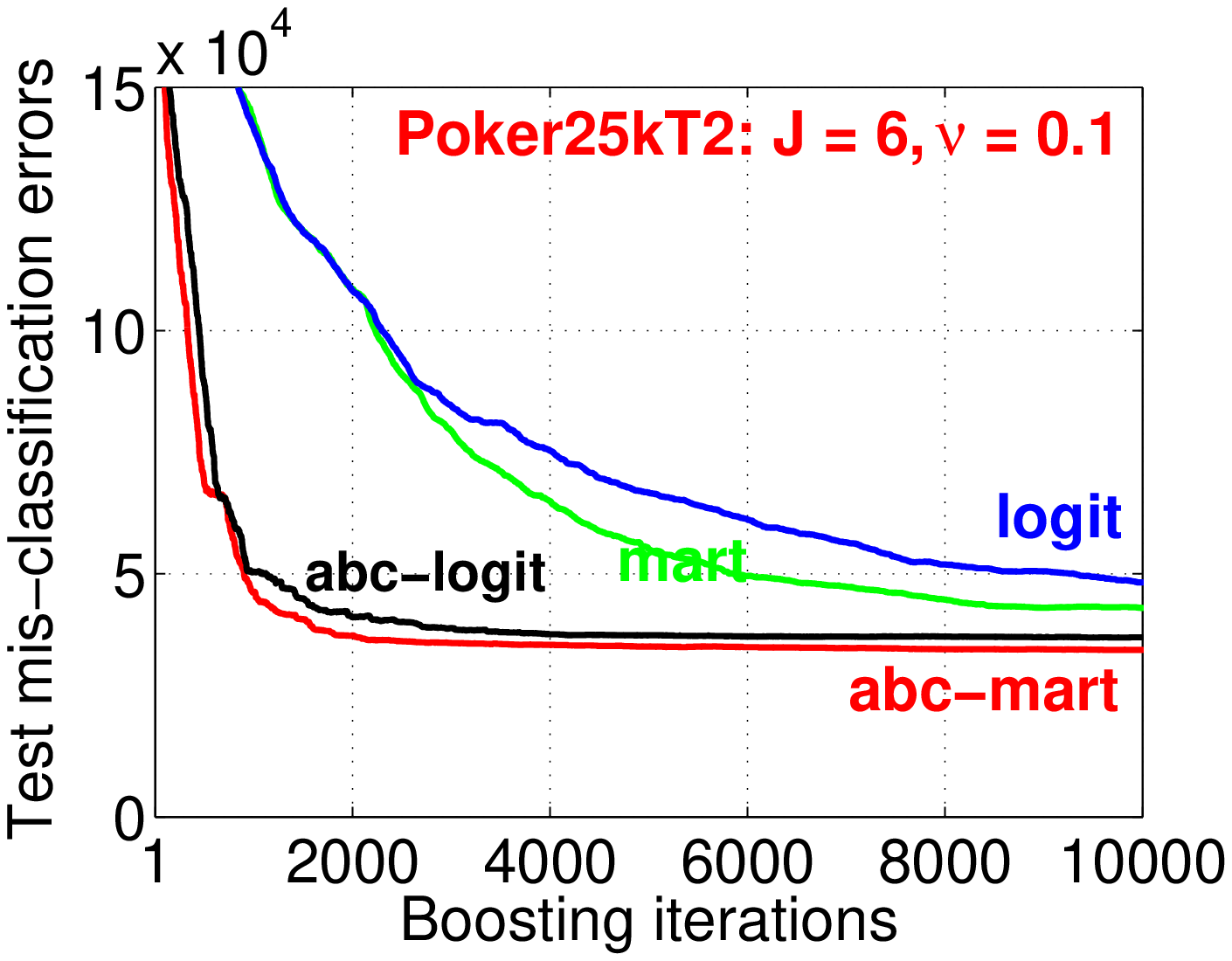}
}

\end{center}
\vspace{-0.1in}
\caption{Test mis-classification errors on \textbf{\em Covertype} and \textbf{\em Poker}. }\label{fig_Cover_Poker}
\end{figure}

\clearpage

\section{More Detailed Experiment Results}\label{sec_detailed_exp}

Ideally, we would like to demonstrate that, with any reasonable choice of parameters $J$ and $\nu$, {\em abc-mart} and {\em abc-logitboost} will always improve {\em mart} and {\em logitboost}, respectively. This is actually indeed the case on the datasets we have experimented.  In this section, we provide the detailed experiment results on {\em Mnist10k}, {\em Poker25kT1}, {\em Poker25kT2}, {\em Letter4k}, and {\em Letter2k}.

\subsection{Detailed Experiment Results on {\em Mnist10k}}

For this  dataset, we experiment with every combination of $J\in\{4, 6, 8, 10, 12, 14, 16,18, 20, 24, 30, 40, 50\}$ and $\nu \in\{0.04, 0.06, 0.08, 0.1\}$. We train the four boosting algorithms till the training loss (\ref{eqn_loss}) is close to the machine accuracy, to exhaust the capacity of the learner so that we could provide a reliable comparison, up to $M=10000$ iterations.

Table \ref{tab_Mnist10k} presents the test mis-classification errors and Table \ref{tab_Mnist10k_P-Value} presents the $P$-values. Figures \ref{fig_Mnist10k_4-10}, \ref{fig_Mnist10k_12-18}, and \ref{fig_Mnist10k_20-50} provide the test mis-classification errors for all boosting iterations.

\begin{table}[h]
\caption{\textbf{\em Mnist10k}. Upper table: The test mis-classification errors of  {\em mart} and \textbf{\em abc-mart} (bold numbers). Bottom table: The test mis-classification errors of  {\em logitboost} and \textbf{\em abc-logitboost} (bold numbers)}
\begin{center}
{\small
{\begin{tabular}{l l l l l }
\hline \hline

&{\em mart} & \textbf{\em abc-mart}\\\hline
  &$\nu = 0.04$ &$\nu=0.06$ &$\nu=0.08$ &$\nu=0.1$ \\
\hline

$J=4$        &3356        \textbf{3060}        &3329        \textbf{3019}        &3318        \textbf{2855}        &3326        \textbf{2794}\\
$J=6$         &3185        \textbf{2760}        &3093        \textbf{2626}        &3129        \textbf{2656}        &3217        \textbf{2590}\\
$J=8$         &3049        \textbf{2558}        &3054        \textbf{2555}        &3054        \textbf{2534}        &3035        \textbf{2577}\\
$J=10$         &3020        \textbf{2547}        &2973        \textbf{2521}        &2990        \textbf{2520}        &2978        \textbf{2506}\\
$J=12$         &2927        \textbf{2498}        &2917        \textbf{2457}        &2945        \textbf{2488}       &2907        \textbf{2490}\\
$J=14$         &2925        \textbf{2487}        &2901        \textbf{2471}        &2877        \textbf{2470}       &2884        \textbf{2454}\\
$J=16$         &2899        \textbf{2478}        &2893        \textbf{2452}        &2873        \textbf{2465}       &2860        \textbf{2451}\\
$J=18$         &2857        \textbf{2469}        &2880        \textbf{2460}        &2870        \textbf{2437}        &2855        \textbf{2454}\\
$J=20$         &2833        \textbf{2441}        &2834        \textbf{2448}        &2834        \textbf{2444} &2815        \textbf{2440}\\
$J=24$         &2840        \textbf{2447}        &2827        \textbf{2431}        &2801        \textbf{2427}        &2784        \textbf{2455}\\
$J=30$         &2826        \textbf{2457}        &2822        \textbf{2443}        &2828        \textbf{2470}        &2807        \textbf{2450}\\
$J=40$         &2837        \textbf{2482}        &2809        \textbf{2440}        &2836        \textbf{2447}        &2782        \textbf{2506}\\
$J=50$         &2813        \textbf{2502}        &2826        \textbf{2459}        &2824        \textbf{2469}        &2786        \textbf{2499}\\
\hline \hline
 &{\em logitboost} & \textbf{\em abc-logit}\\\hline
  &$\nu = 0.04$ &$\nu=0.06$ &$\nu=0.08$ &$\nu=0.1$ \\
\hline

$J=4$        &2936        \textbf{2630}        &2970        \textbf{2600}        &2980        \textbf{2535}        &3017        \textbf{2522}\\
$J=6$        &2710        \textbf{2263}        &2693        \textbf{2252}        &2710        \textbf{2226}       &2711        \textbf{2223}\\
$J=8$        &2599        \textbf{2159}        &2619        \textbf{2138}        &2589        \textbf{2120}       &2597        \textbf{2143}\\
$J=10$        &2553        \textbf{2122}        &2527        \textbf{2118}       &2516        \textbf{2091}       &2500        \textbf{2097}\\
$J=12$        &2472        \textbf{2084}        &2468        \textbf{2090}       &2468        \textbf{2090}       &2464        \textbf{2095}\\
$J=14$        &2451        \textbf{2083}        &2420        \textbf{2094}       &2432        \textbf{2063}       &2419        \textbf{2050}\\
$J=16$        &2424        \textbf{2111}        &2437        \textbf{2114}       &2393        \textbf{2097}       &2395        \textbf{2082}\\
$J=18$        &2399        \textbf{2088}        &2402        \textbf{2087}       &2389        \textbf{2088}       &2380        \textbf{2097}\\
$J=20$        &2388        \textbf{2128}        &2414        \textbf{2112}       &2411        \textbf{2095}       &2381        \textbf{2102}\\
$J=24$        &2442        \textbf{2174}        &2415        \textbf{2147}       &2417       \textbf{2129}      &2419        \textbf{2138}\\
$J=30$       & 2468        \textbf{2235}        &2434        \textbf{2237}       &2423        \textbf{2221}        &2449        \textbf{2177}\\
$J=40$       & 2551        \textbf{2310}        &2509        \textbf{2284}       &2518        \textbf{2257}        &2531        \textbf{2260}\\
$J=50$       & 2612        \textbf{2353}        &2622        \textbf{2359}       &2579        \textbf{2332}        &2570        \textbf{2341}\\
\hline\hline
\end{tabular}}}
\end{center}
\label{tab_Mnist10k}
\end{table}

\begin{table}[h]
\caption{\textbf{\em Mnist10k}: $P$-values. See Sec. \ref{sec_P-Value} for the definitions of P1, P2, P3, and P4.}
\begin{center}{\scriptsize
\begin{tabular}{l r r r r}
\hline\hline
& & \textbf{P1} \\
\hline\hline
  &$\nu = 0.04$ &$\nu=0.06$ &$\nu=0.08$ &$\nu=0.1$ \\
\hline
$J=4$    &$7\times 10^{-5}$&$3\times 10^{-5}$&$7\times 10^{-10}$&$1\times 10^{-12}$\\
$J=6$   &$8\times 10^{-9}$ &$1\times 10^{-10}$&$9\times 10^{-11}$&0 \\
$J=8$   &$9\times 10^{-12}$&$4\times 10^{-12}$&$5\times 10^{-13}$&$2\times 10^{-10}$\\
$J=10$   &$4\times 10^{-11}$&$2\times 10^{-10}$&$4\times 10^{-11}$&$3\times 10^{-11}$\\
$J=12$   &$1\times 10^{-9}$&$7\times 10^{-11}$&$1\times 10^{-10}$&$3\times 10^{-9}$\\
$J=14$   &$6\times 10^{-10}$&$1\times 10^{-9}$&$6\times 10^{-9}$&$9\times 10^{-10}$\\
$J=16$   &$2\times 10^{-9}$&$3\times 10^{-10}$&$6\times 10^{-9}$&$5\times 10^{-9}$\\
$J=18$   &$3\times 10^{-8}$&$2\times 10^{-9}$&$6\times 10^{-10}$&$9\times 10^{-9}$\\
$J=20$   &$2\times 10^{-8}$&$3\times 10^{-8}$&$2\times 10^{-8}$&$6\times 10^{-8}$\\
$J=24$   &$2\times 10^{-8}$&$1\times 10^{-8}$&$6\times 10^{-8}$&$2\times 10^{-6}$\\
$J=30$   &$1\times 10^{-7}$&$5\times 10^{-8}$&$2\times 10^{-7}$&$2\times 10^{-7}$\\
$J=40$   &$3\times 10^{-7}$&$1\times 10^{-7}$&$2\times 10^{-8}$&$5\times 10^{-5}$\ \\
$J=50$   &$6\times 10^{-6}$&$1\times 10^{-7}$&$3\times 10^{-7}$&$3\times 10^{-5}$\\
\hline\hline
& & \textbf{P2}\\
\hline\hline
  &$\nu = 0.04$ &$\nu=0.06$ &$\nu=0.08$ &$\nu=0.1$ \\
\hline
$J=4$  &$2\times 10^{-8}$\     &$2\times 10^{-6}$\    &$6\times 10^{-6}$&$3\times 10^{-6}$\ \\
$J=6$   &$1\times 10^{-10}$&$4\times 10^{-8}$&$9\times 10^{-9}$&$8\times10^{-12}$\\
$J=8$   &$4\times 10^{-10}$&$2\times 10^{-9}$&$1\times 10^{-10}$&$1\times 10^{-9}$\\
$J=10$   &$7\times 10^{-11}$&$4\times 10^{-10}$&$3\times 10^{-11}$&$2\times 10^{-11}$\\
$J=12$   &$1\times 10^{-10}$&$2\times 10^{-10}$&$2\times 10^{-11}$&$3\times 10^{-10}$\\
$J=14$   &$2\times 10^{-11}$&$8\times 10^{-12}$&$2\times 10^{-10}$&$3\times 10^{-11}$\\
$J=16$   &$1\times 10^{-11}$&$8\times 10^{-11}$&$7\times 10^{-12}$&$3\times 10^{-11}$\\
$J=18$   &$5\times 10^{-11}$&$9\times 10^{-12}$&$6\times 10^{-12}$&$9\times 10^{-12}$\\
$J=20$   &$2\times 10^{-10}$&$2\times 10^{-9}$&$1\times 10^{-9}$&$4\times 10^{-10}$\\
$J=24$   &$1\times 10^{-8}$&$3\times 10^{-9}$&$3\times 10^{-8}$&$1\times 10^{-7}$\\
$J=30$   &$2\times 10^{-7}$&$2\times 10^{-8}$&$5\times 10^{-9}$&$2\times 10^{-7}$\\
$J=40$   &$3\times 10^{-5}$&$1\times 10^{-5}$&$4\times 10^{-6}$&$2\times 10^{-4}$\ \\
$J=50$   &$0.0026$\    &$0.0023$\    &$3\times 10^{-4}$\   &$0.0013$\ \\
\hline\hline
& & \textbf{P3} \\
\hline\hline
  &$\nu = 0.04$ &$\nu=0.06$ &$\nu=0.08$ &$\nu=0.1$ \\
\hline
$J=4$  &$3\times 10^{-9}$\     &$5\times 10^{-9}$\    &$4\times 10^{-6}$&$7\times 10^{-6}$\ \\
$J=6$   &$4\times 10^{-13}$&$2\times 10^{-8}$&$2\times 10^{-10}$&$3\times10^{-8}$\\
$J=8$   &$2\times 10^{-9}$&$3\times 10^{-10}$&$3\times 10^{-10}$&$6\times 10^{-11}$\\
$J=10$   &$1\times 10^{-10}$&$8\times 10^{-10}$&$6\times 10^{-11}$&$4\times 10^{-10}$\\
$J=12$   &$2\times 10^{-10}$&$2\times 10^{-8}$&$1\times 10^{-9}$&$1\times 10^{-9}$\\
$J=14$   &$5\times 10^{-10}$&$6\times 10^{-9}$&$4\times 10^{-10}$&$4\times 10^{-10}$\\
$J=16$   &$2\times 10^{-8}$&$2\times 10^{-7}$&$1\times 10^{-8}$&$1\times 10^{-8}$\\
$J=18$   &$4\times 10^{-9}$&$8\times 10^{-9}$&$6\times 10^{-8}$&$3\times 10^{-8}$\\
$J=20$   &$1\times 10^{-6}$&$2\times 10^{-7}$&$6\times 10^{-8}$&$2\times 10^{-7}$\\
$J=24$   &$2\times 10^{-5}$&$9\times 10^{-6}$&$3\times 10^{-6}$&$9\times 10^{-7}$\\
$J=30$   &$5\times 10^{-4}$&$0.0011$&$1\times 10^{-4}$&$2\times 10^{-5}$\\
$J=40$   &$0.0056$\    &$0.0103$\    &$0.0024$\   &$1\times 10^{-4}$\ \\
$J=50$   &$0.0145$\    &$0.0707$\    &$0.0218$\   &$0.0102$\ \\
\hline\hline
& & \textbf{P4} \\
\hline\hline
  &$\nu = 0.04$ &$\nu=0.06$ &$\nu=0.08$ &$\nu=0.1$ \\
\hline
$J=4$  &$1\times 10^{-5}$\     &$2\times 10^{-7}$\    &$4\times 10^{-10}$&$5\times 10^{-12}$\ \\
$J=6$   &$5\times 10^{-11}$&$7\times 10^{-11}$&$1\times 10^{-12}$&$6\times10^{-13}$\\
$J=8$   &$4\times 10^{-11}$&$5\times 10^{-13}$&$2\times 10^{-12}$&$8\times 10^{-12}$\\
$J=10$   &$6\times 10^{-11}$&$5\times 10^{-10}$&$8\times 10^{-11}$&$7\times 10^{-10}$\\
$J=12$   &$2\times 10^{-9}$&$6\times 10^{-9}$&$6\times 10^{-9}$&$1\times 10^{-8}$\\
$J=14$   &$1\times 10^{-8}$&$4\times 10^{-7}$&$1\times 10^{-8}$&$9\times 10^{-9}$\\
$J=16$   &$1\times 10^{-6}$&$5\times 10^{-7}$&$3\times 10^{-6}$&$9\times 10^{-7}$\\
$J=18$   &$1\times 10^{-6}$&$8\times 10^{-7}$&$2\times 10^{-6}$&$8\times 10^{-6}$\\
$J=20$   &$4\times 10^{-5}$&$2\times 10^{-6}$&$8\times 10^{-7}$&$1\times 10^{-5}$\\
$J=24$   &$3\times 10^{-5}$&$3\times 10^{-5}$&$7\times 10^{-6}$&$1\times 10^{-5}$\\
$J=30$   &$3\times 10^{-4}$\    &$0.0016$\    &$0.0012$\   &$2\times 10^{-5}$\\
$J=40$   &$2\times 10^{-4}$\    &$5\times 10^{-4}$&$6\times 10^{-5}$&$3\times 10^{-5}$\\
$J=50$   &$9\times10^{-5}$\    &$7\times10^{-5}$\    &$2\times 10^{-4}$\   &$4\times10^{-4}$\\
\hline\hline\end{tabular}}\end{center}
\label{tab_Mnist10k_P-Value}
\end{table}

\begin{figure}[h]
\begin{center}
\mbox{
\includegraphics[width=2.2in]{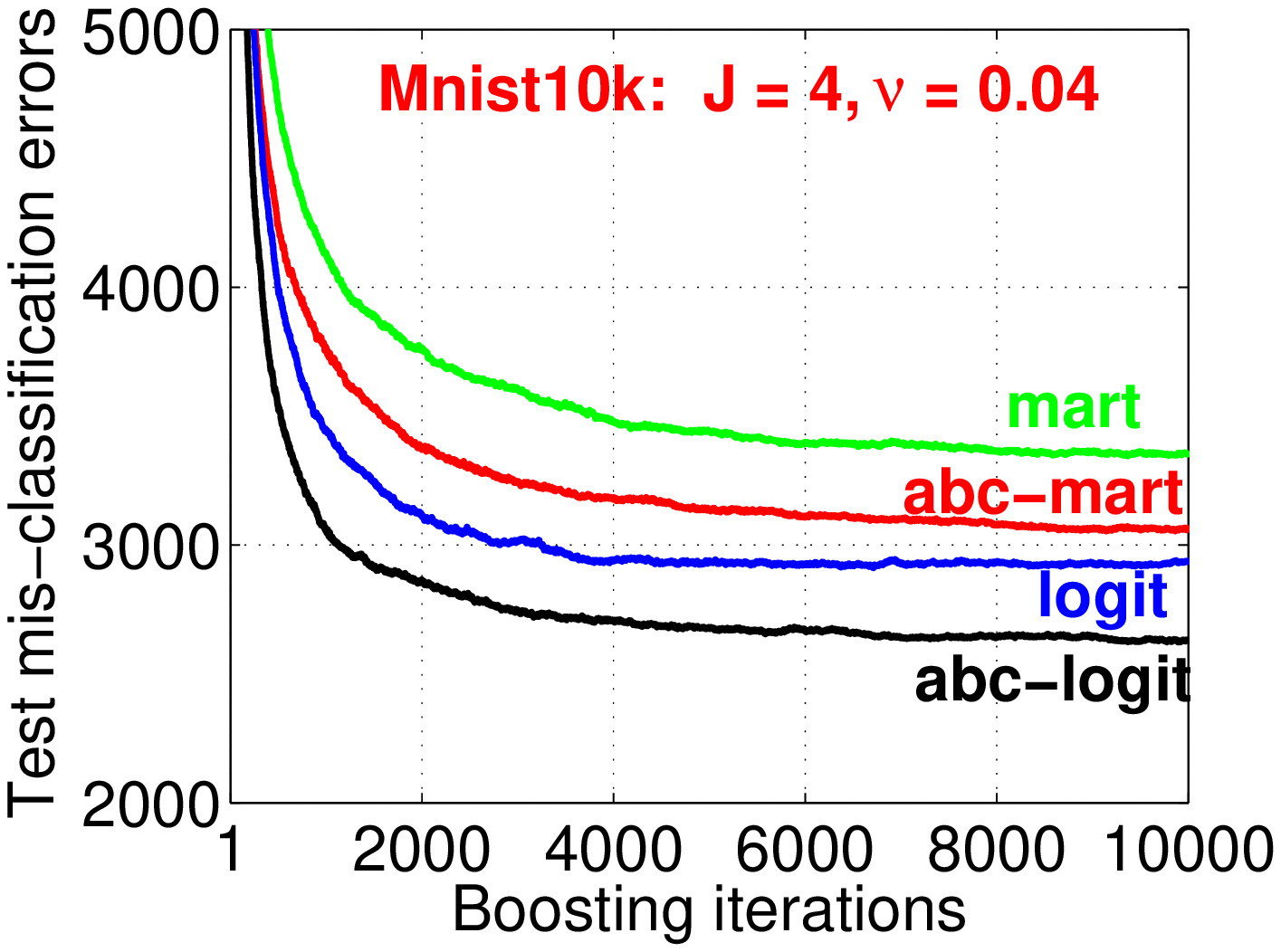}\hspace{-0.05in}
\includegraphics[width=2.2in]{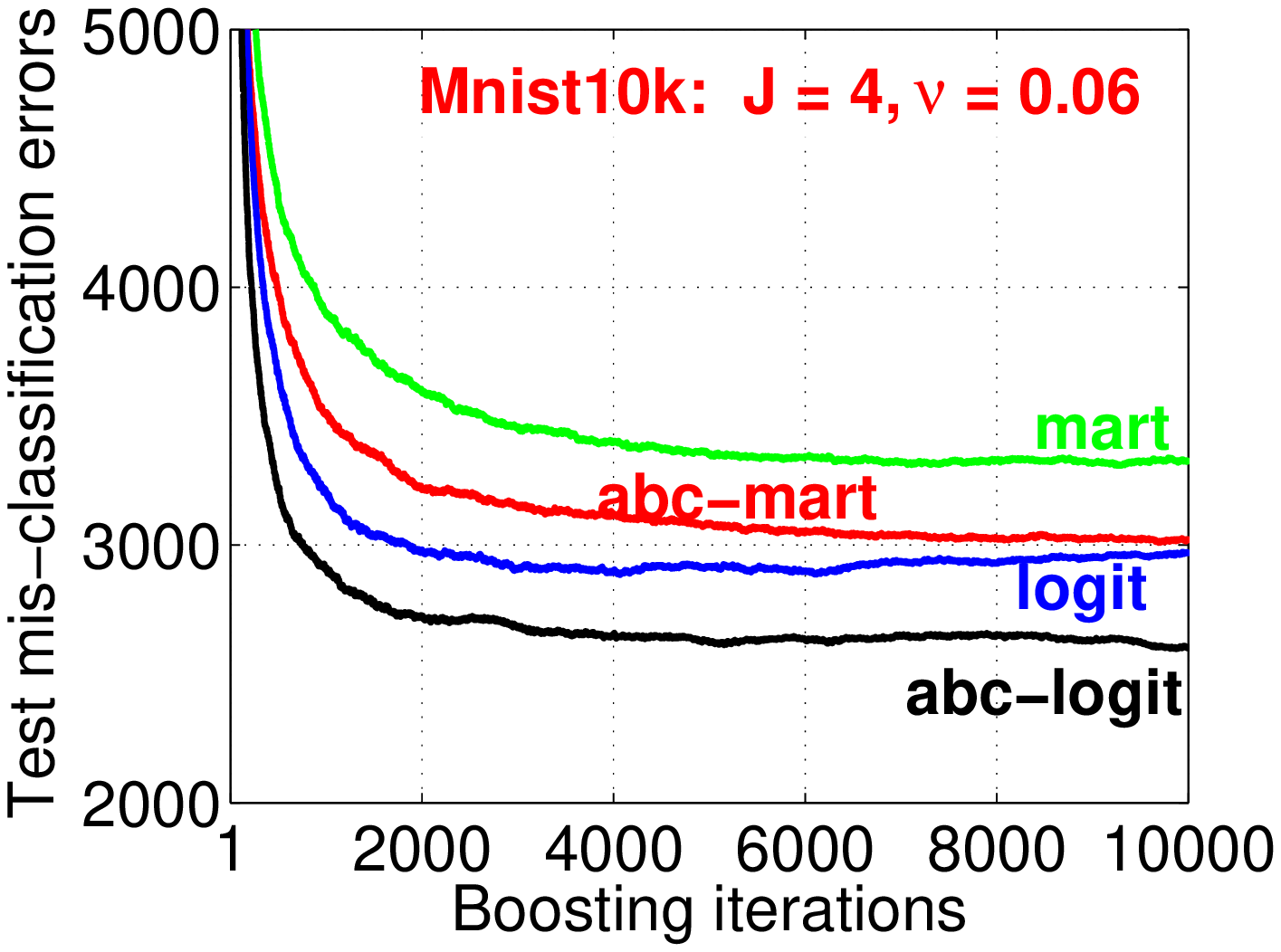}\hspace{-0.05in}
\includegraphics[width=2.2in]{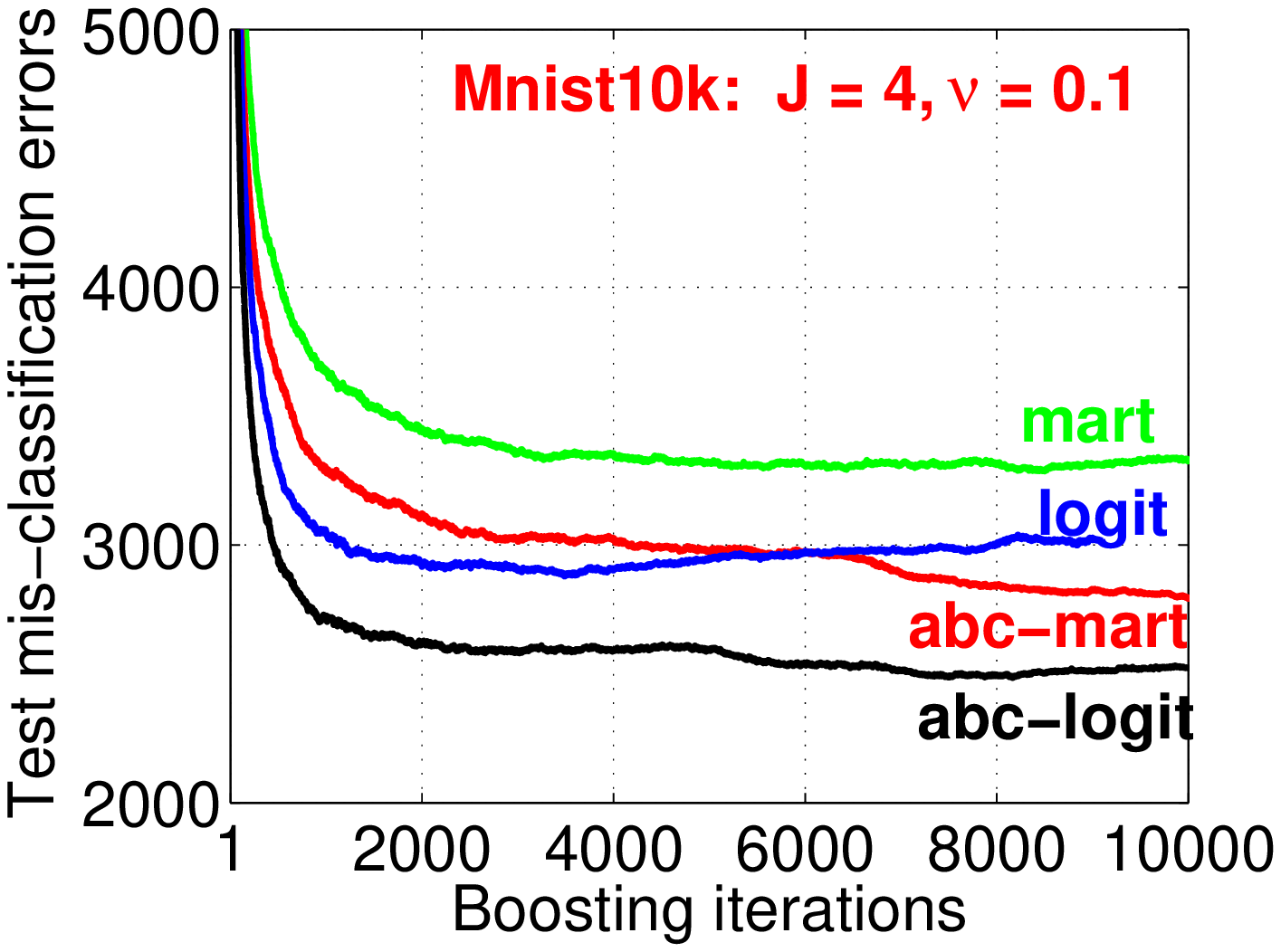}
}
\mbox{
\includegraphics[width=2.2in]{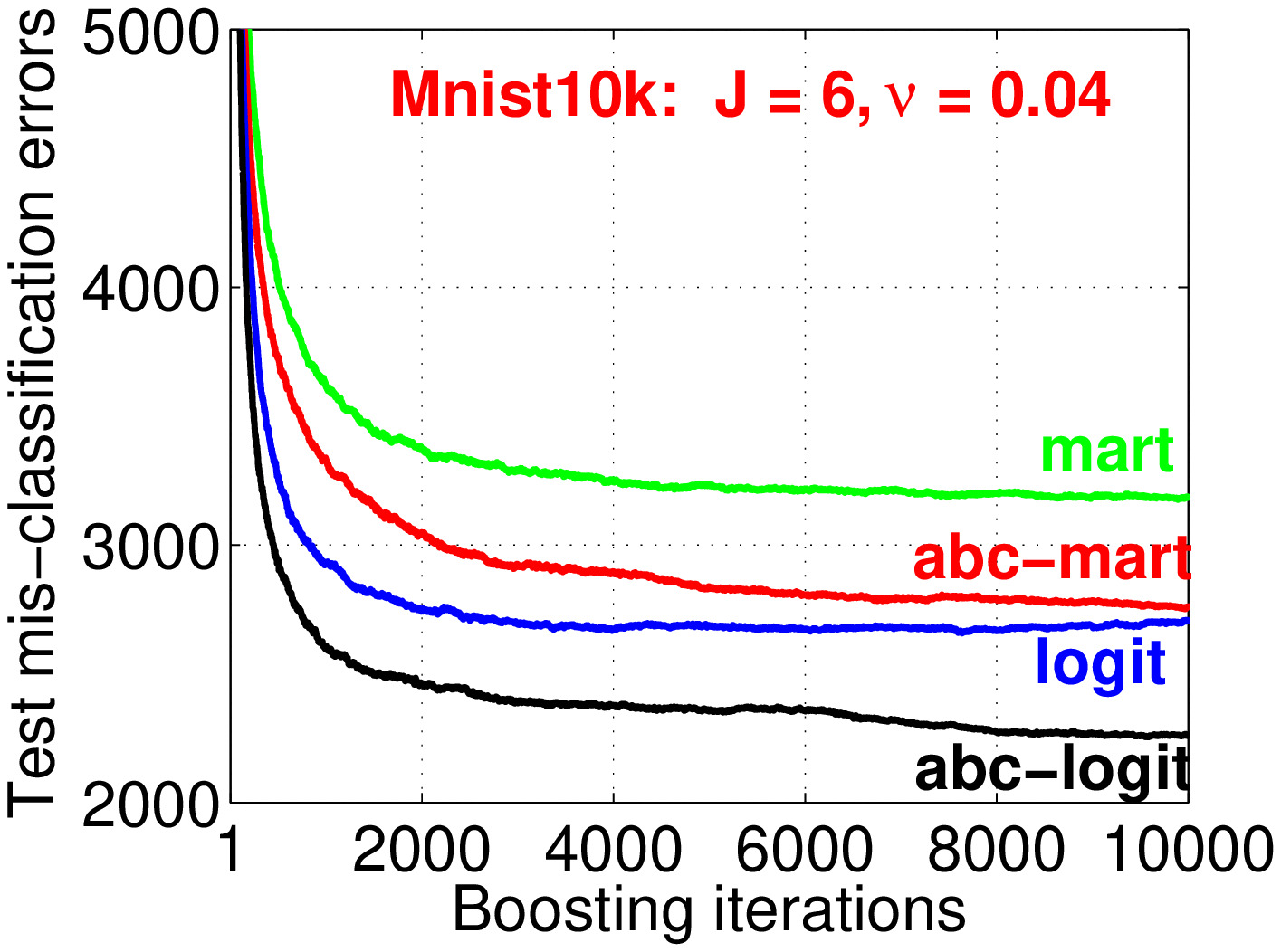}\hspace{-0.05in}
\includegraphics[width=2.2in]{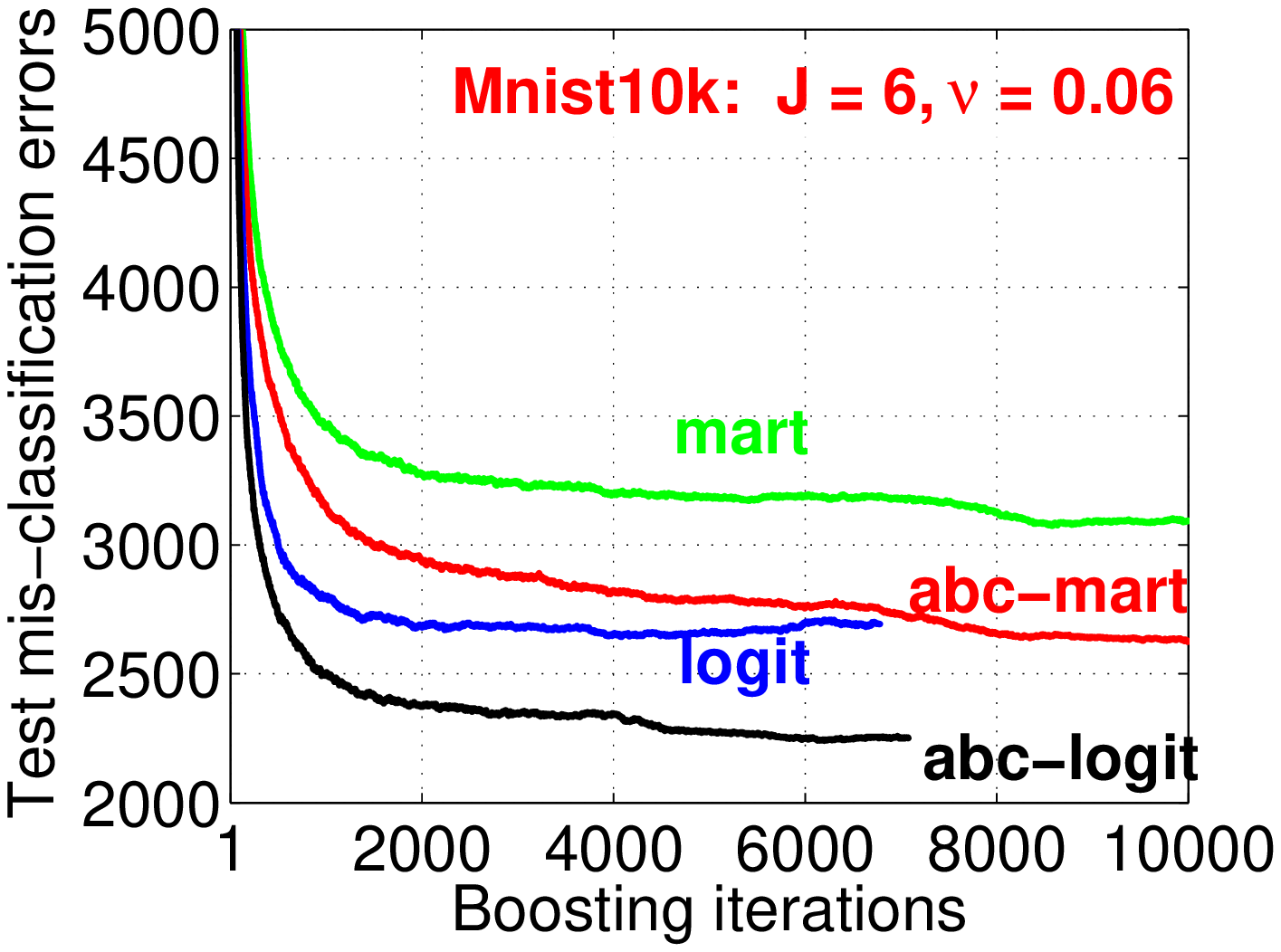}\hspace{-0.05in}
\includegraphics[width=2.2in]{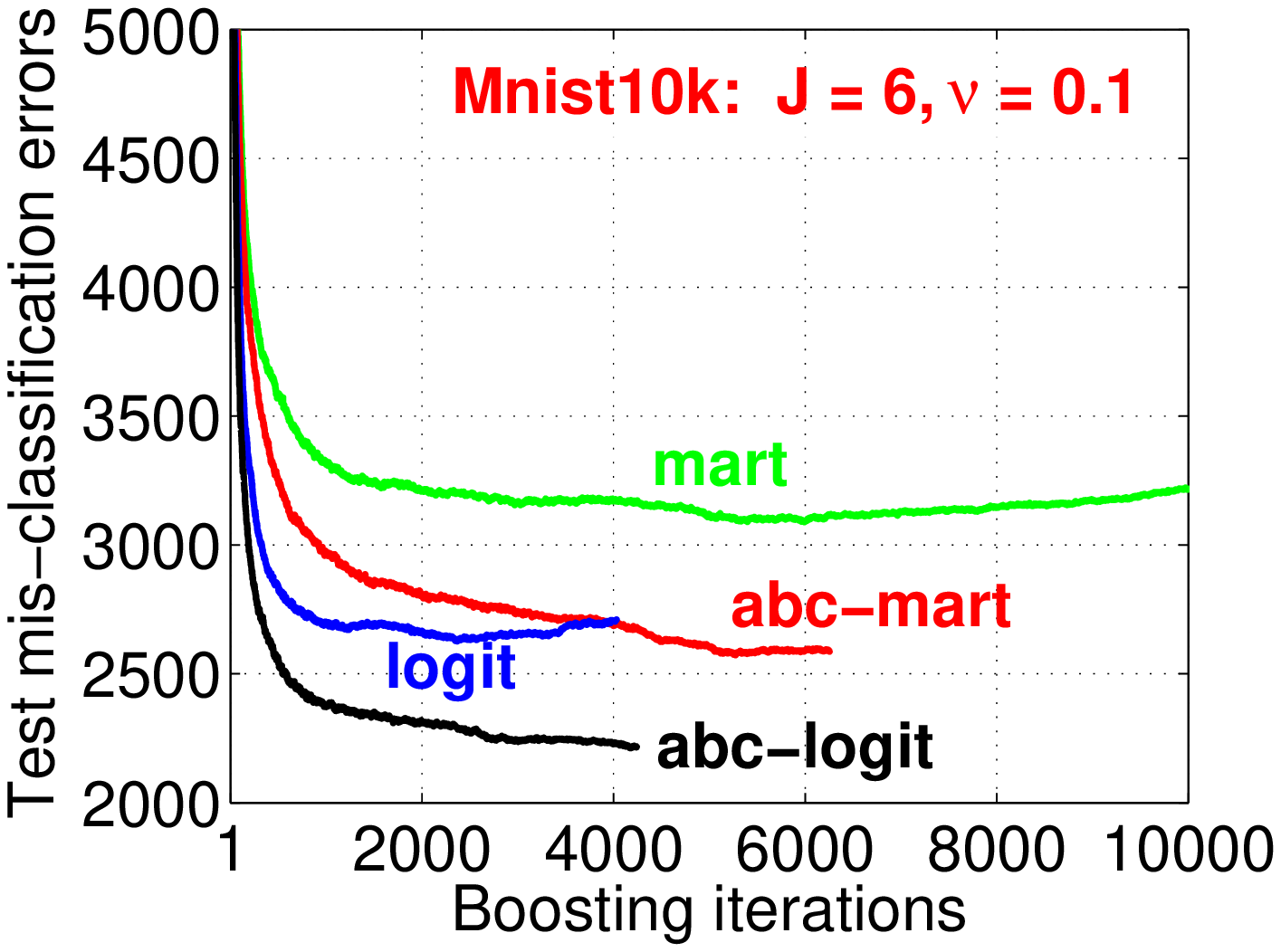}
}
\mbox{
\includegraphics[width=2.2in]{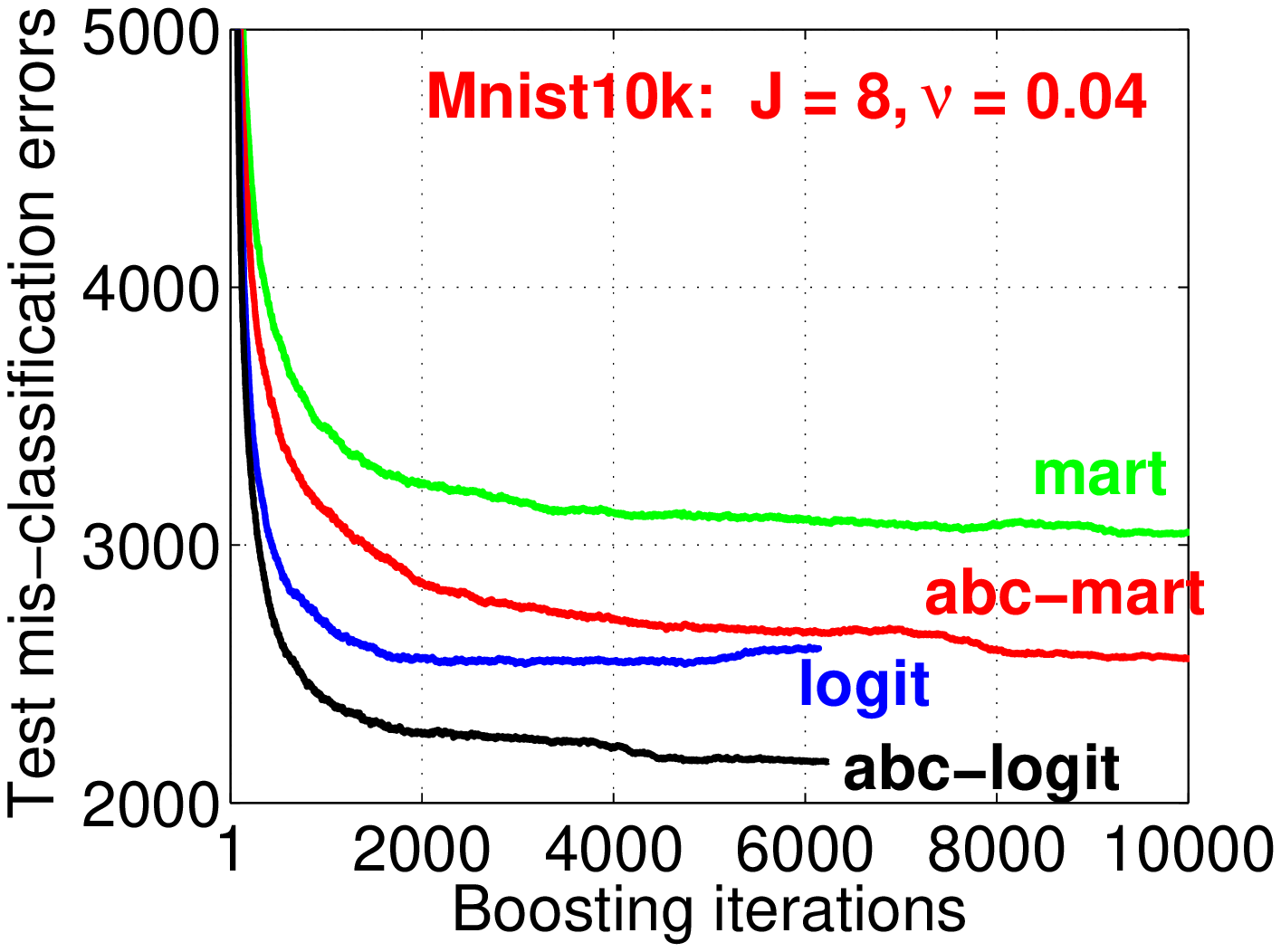}\hspace{-0.05in}
\includegraphics[width=2.2in]{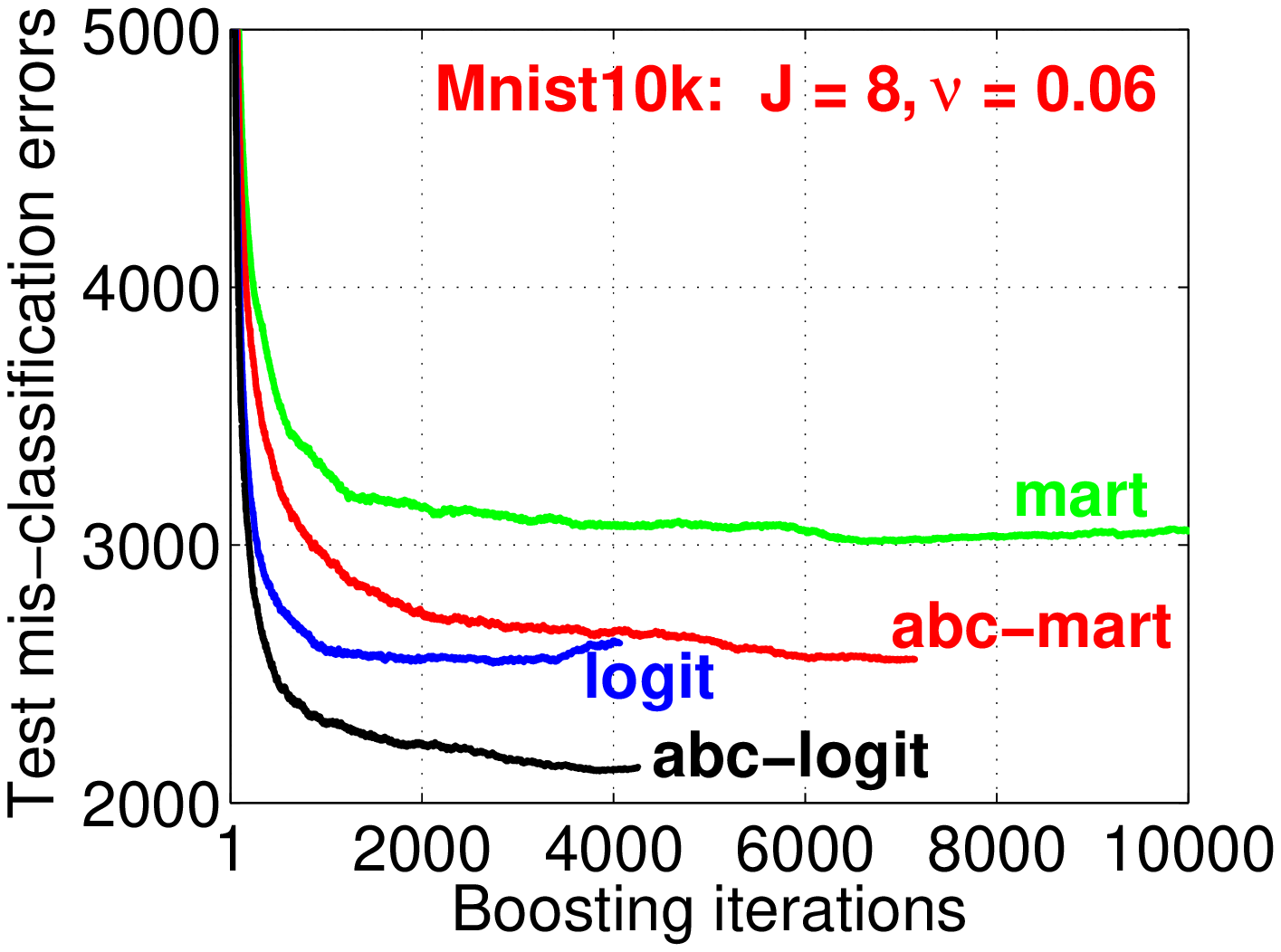}\hspace{-0.05in}
\includegraphics[width=2.2in]{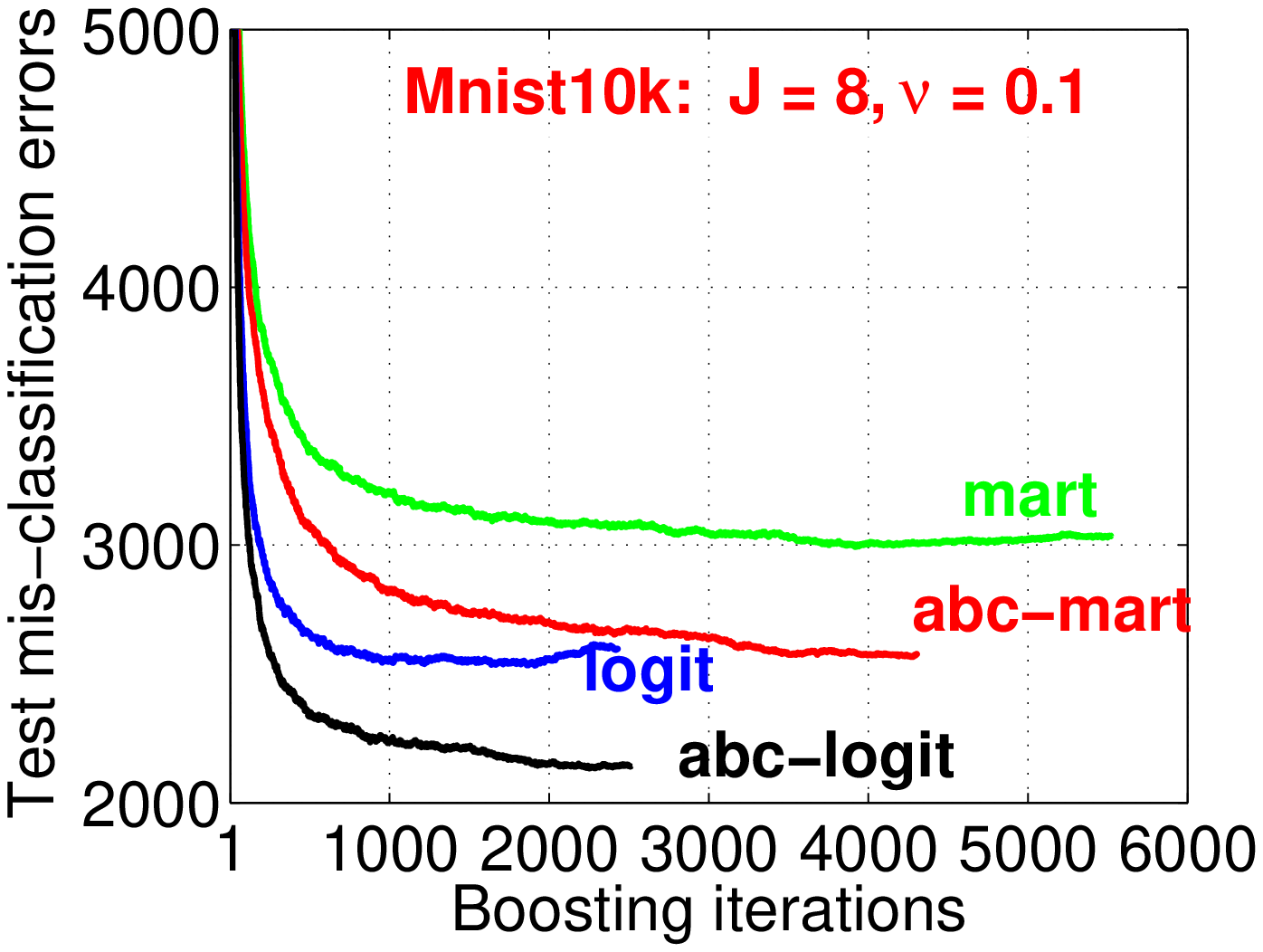}
}
\mbox{
\includegraphics[width=2.2in]{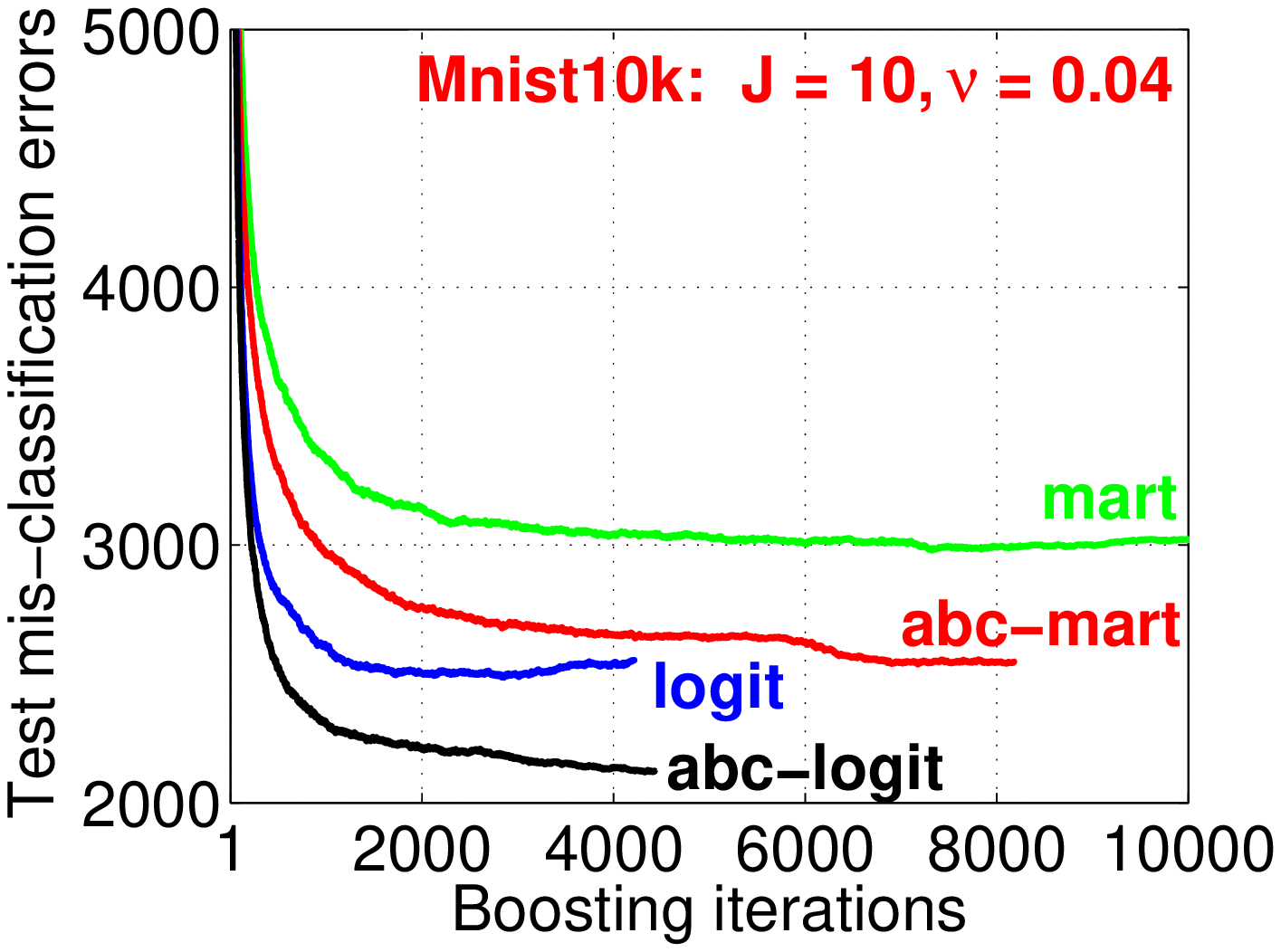}\hspace{-0.05in}
\includegraphics[width=2.2in]{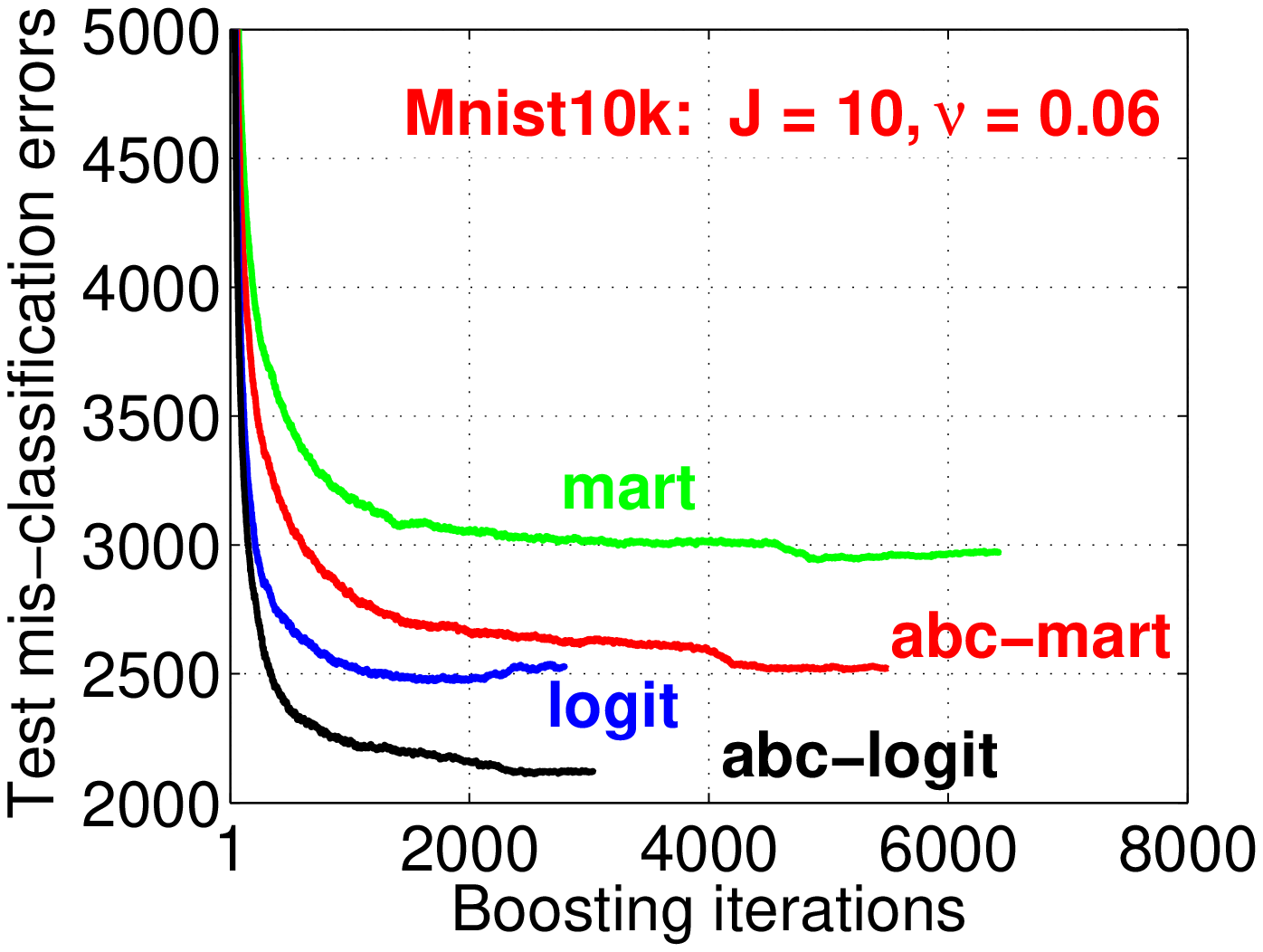}\hspace{-0.05in}
\includegraphics[width=2.2in]{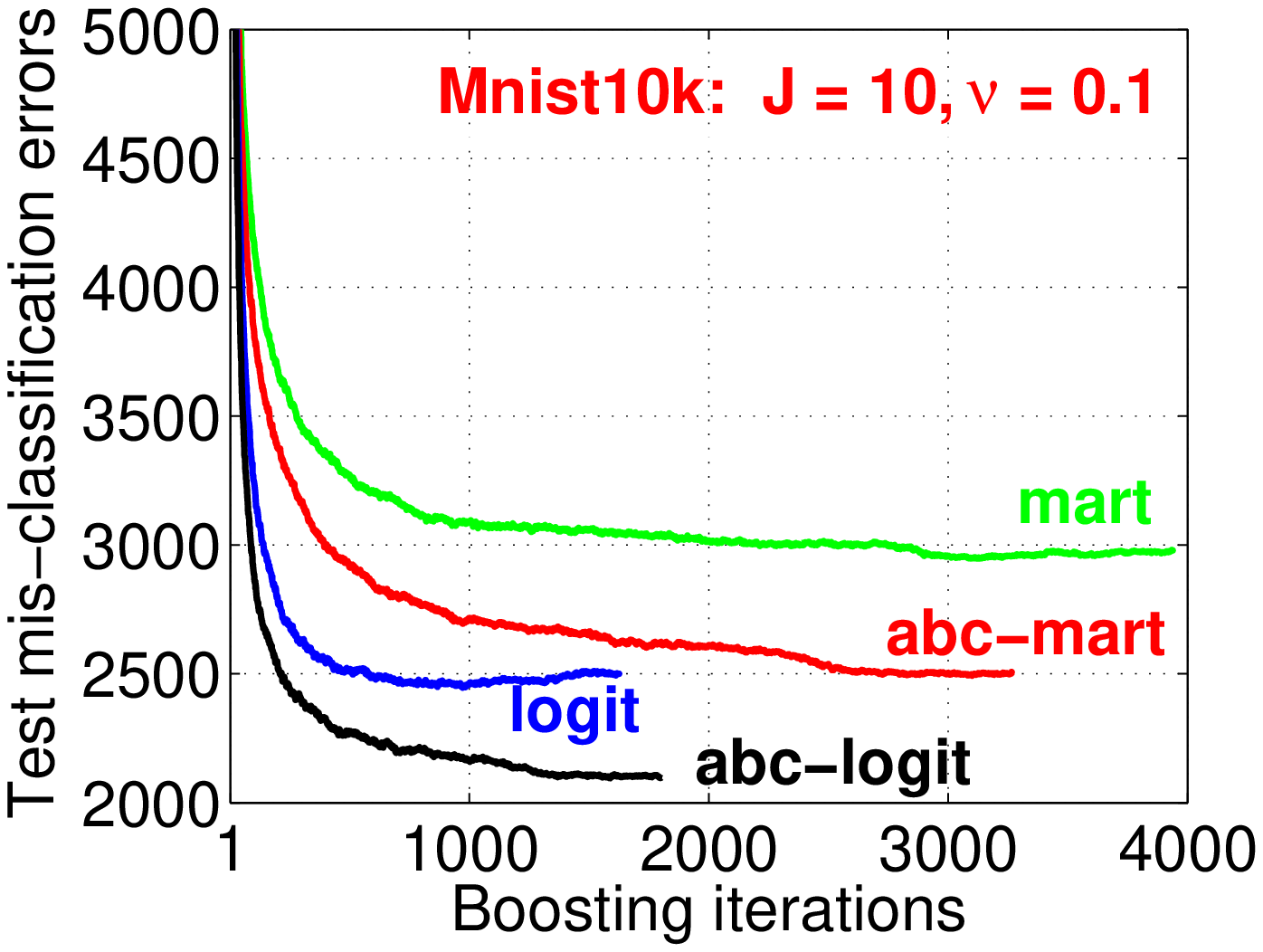}
}
\end{center}
\vspace{-0.1in}
\caption{\textbf{\em Mnist10k}.Test mis-classification errors of four algorithms.  $J=4$, 6, 8, 10. }\label{fig_Mnist10k_4-10}
\end{figure}

\begin{figure}[h]
\begin{center}
\mbox{
\includegraphics[width=2.2in]{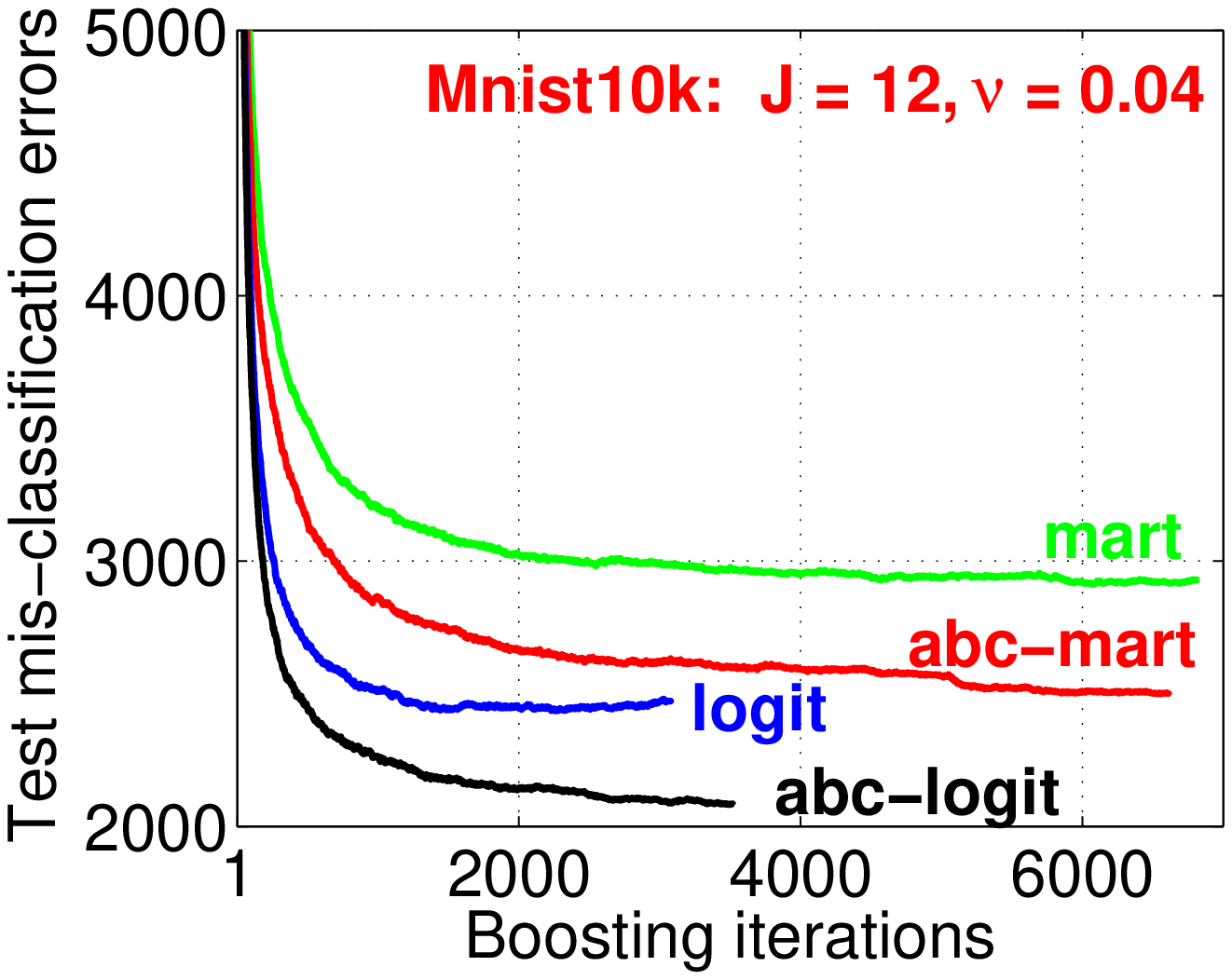}\hspace{-0.05in}
\includegraphics[width=2.2in]{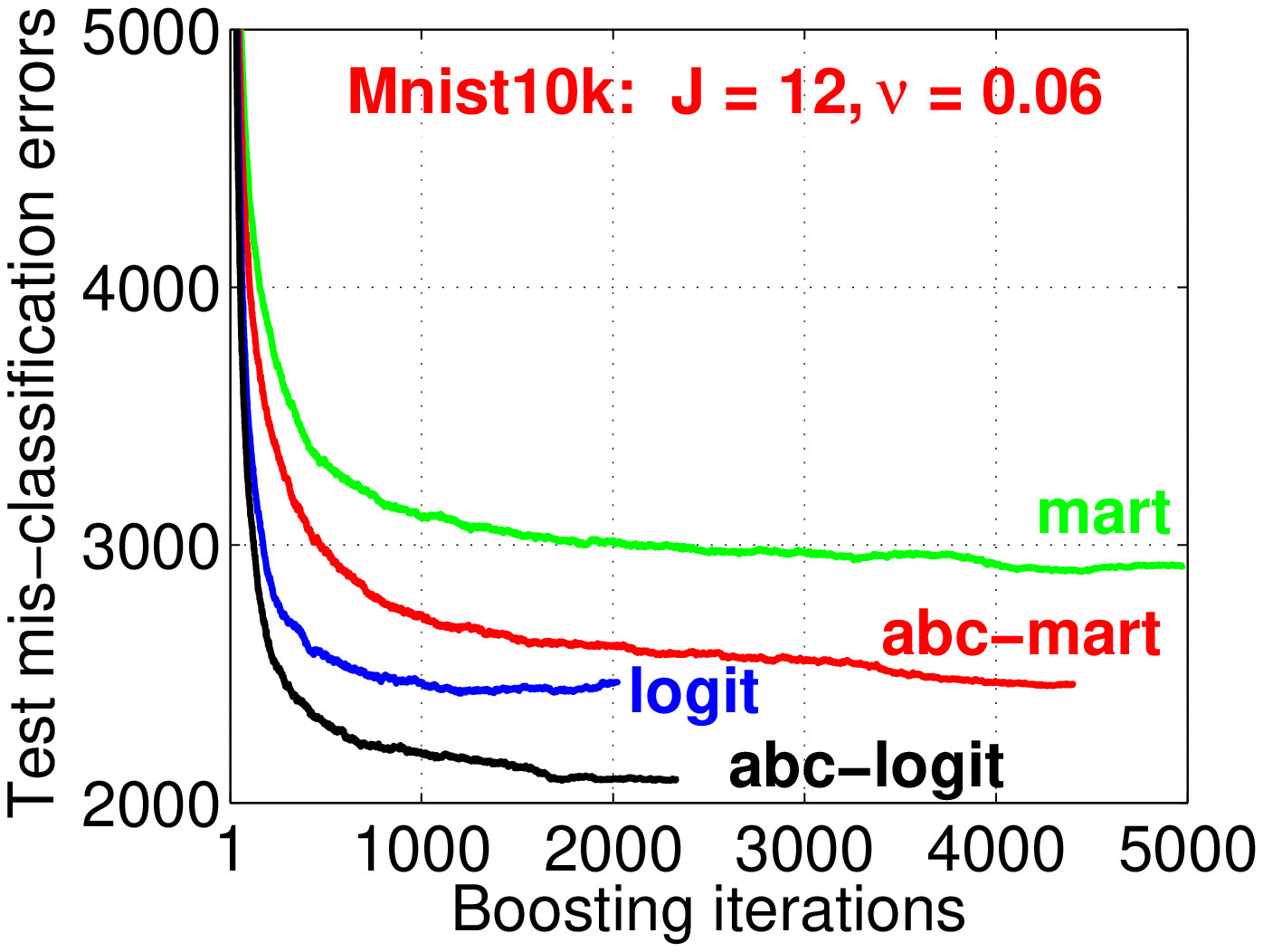}\hspace{-0.05in}
\includegraphics[width=2.2in]{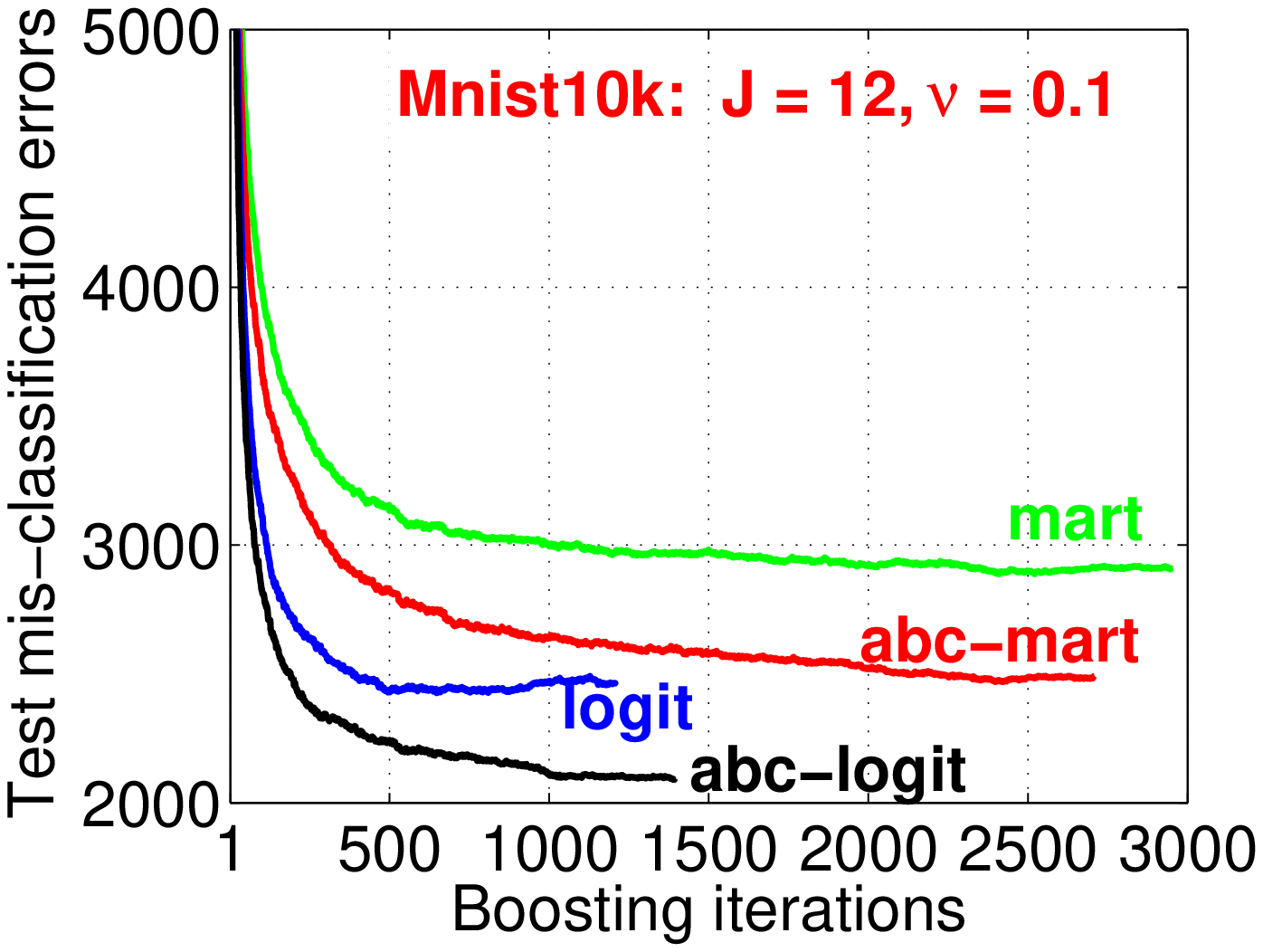}
}
\mbox{
\includegraphics[width=2.2in]{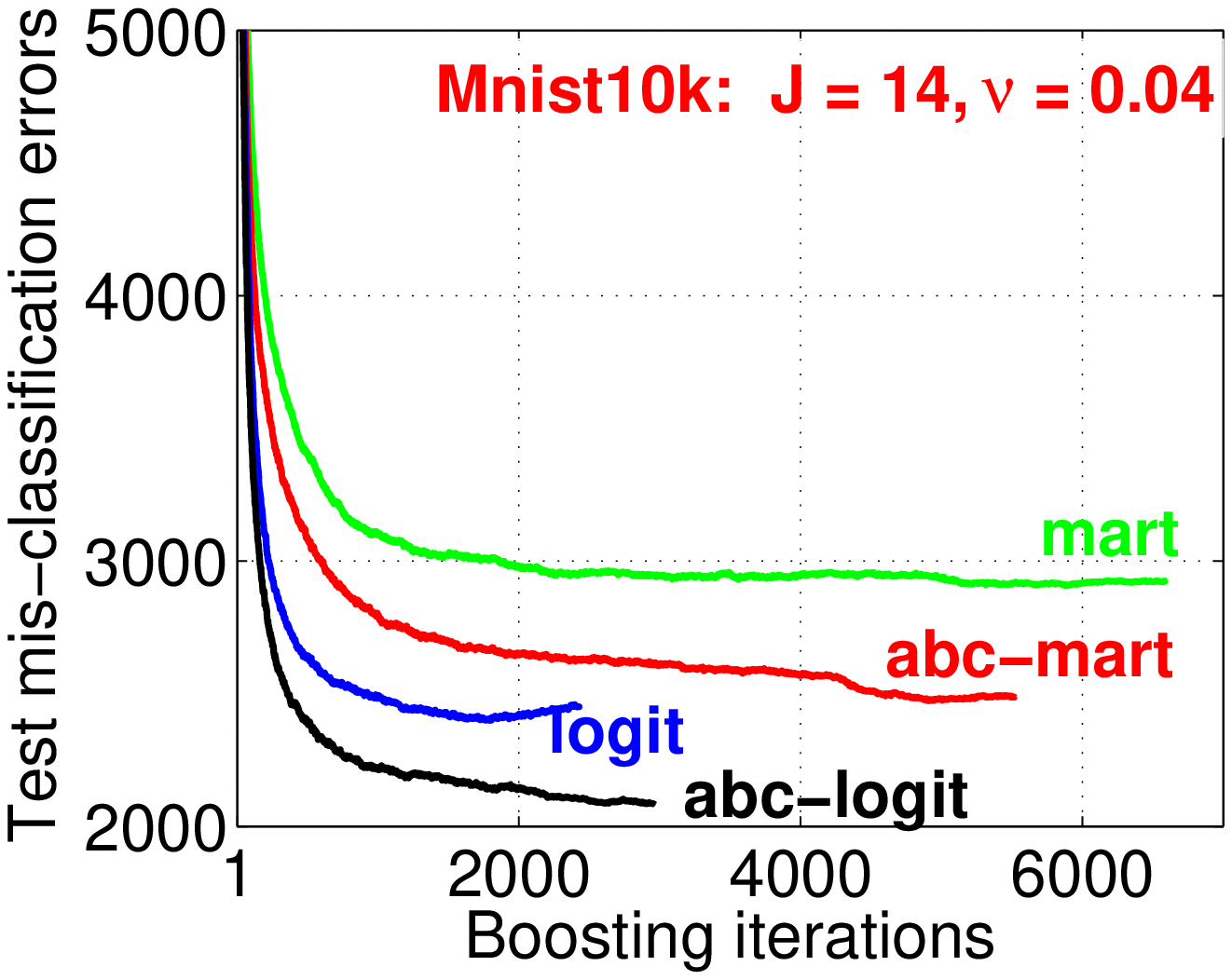}\hspace{-0.05in}
\includegraphics[width=2.2in]{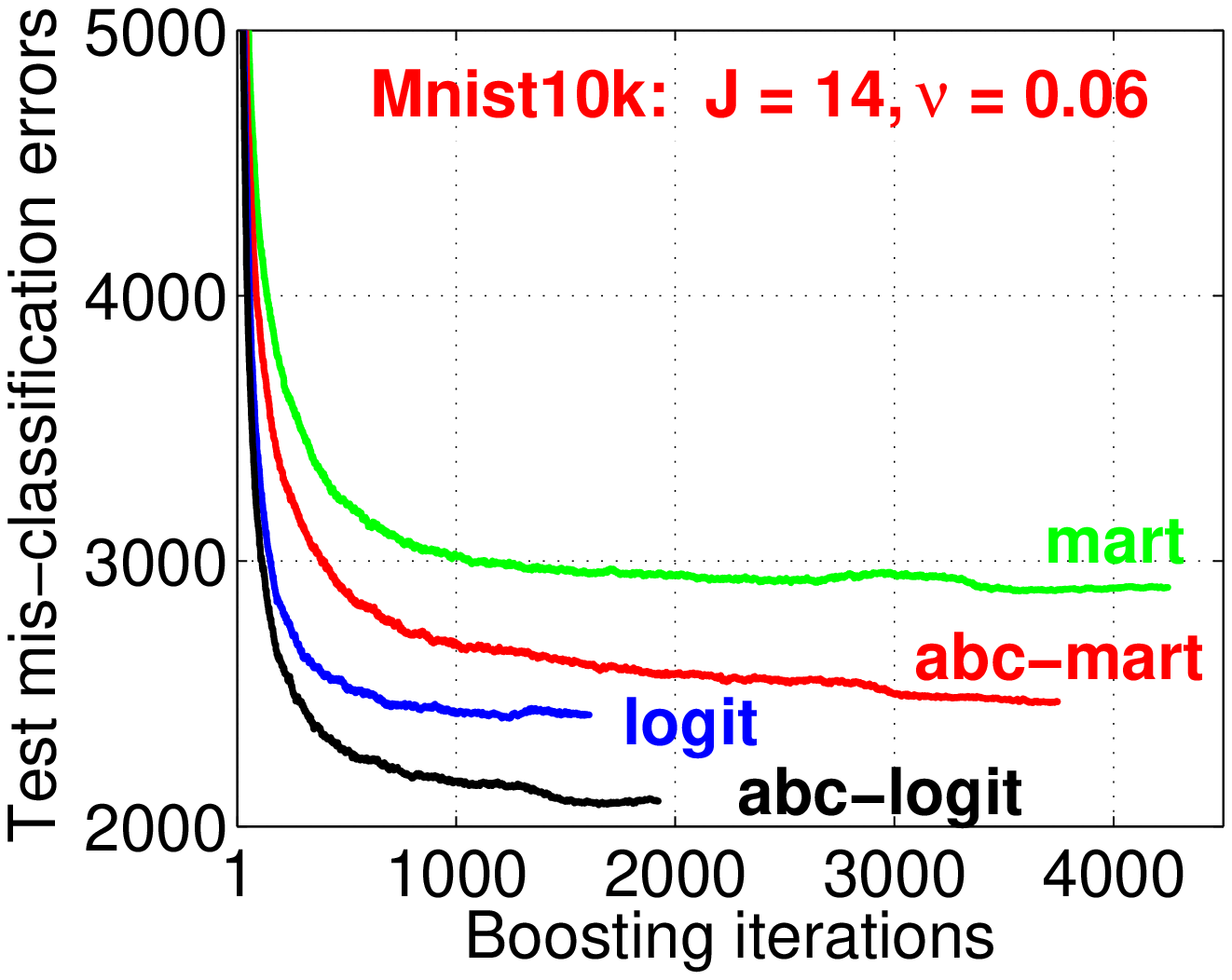}\hspace{-0.05in}
\includegraphics[width=2.2in]{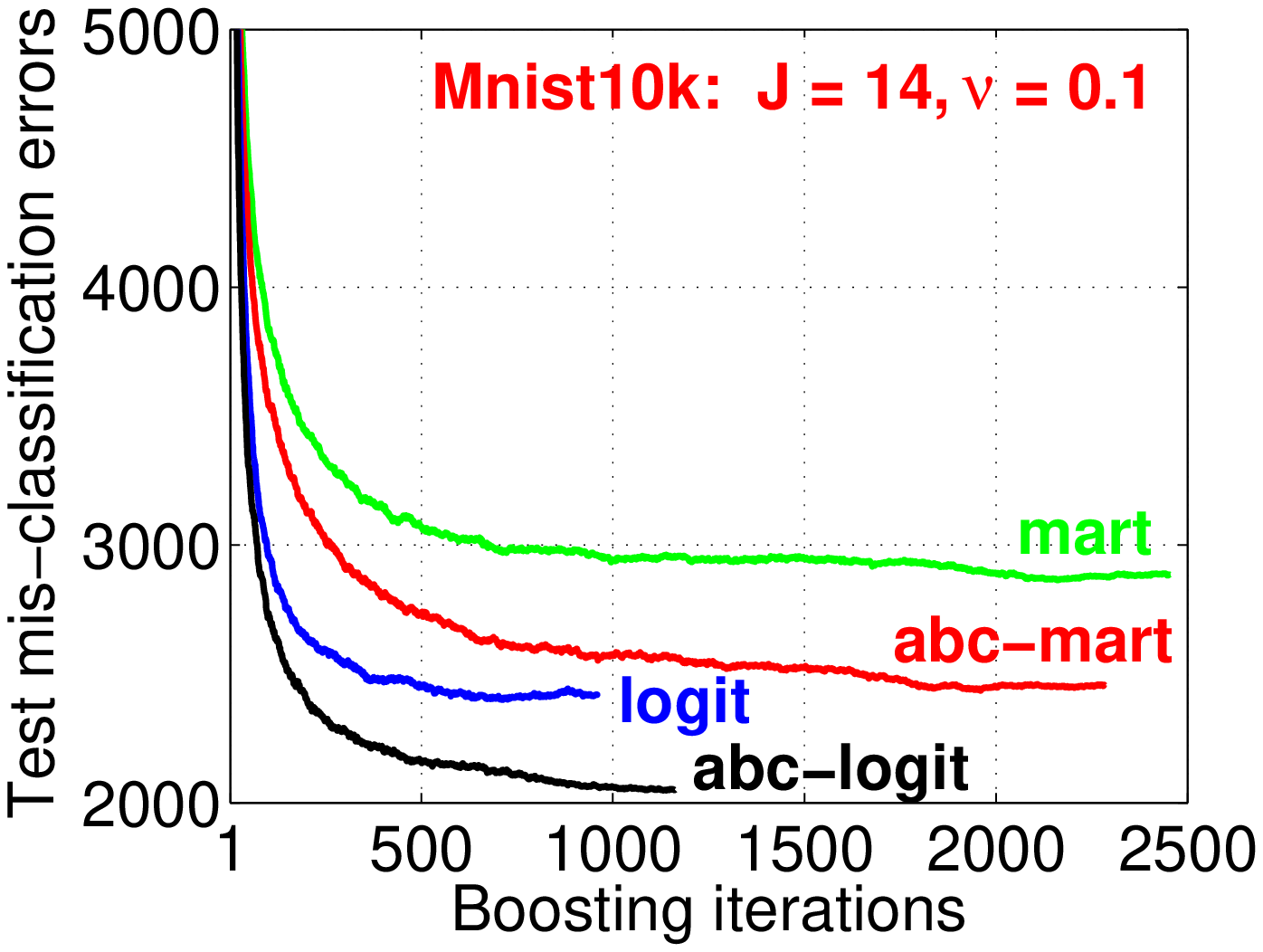}
}
\mbox{
\includegraphics[width=2.2in]{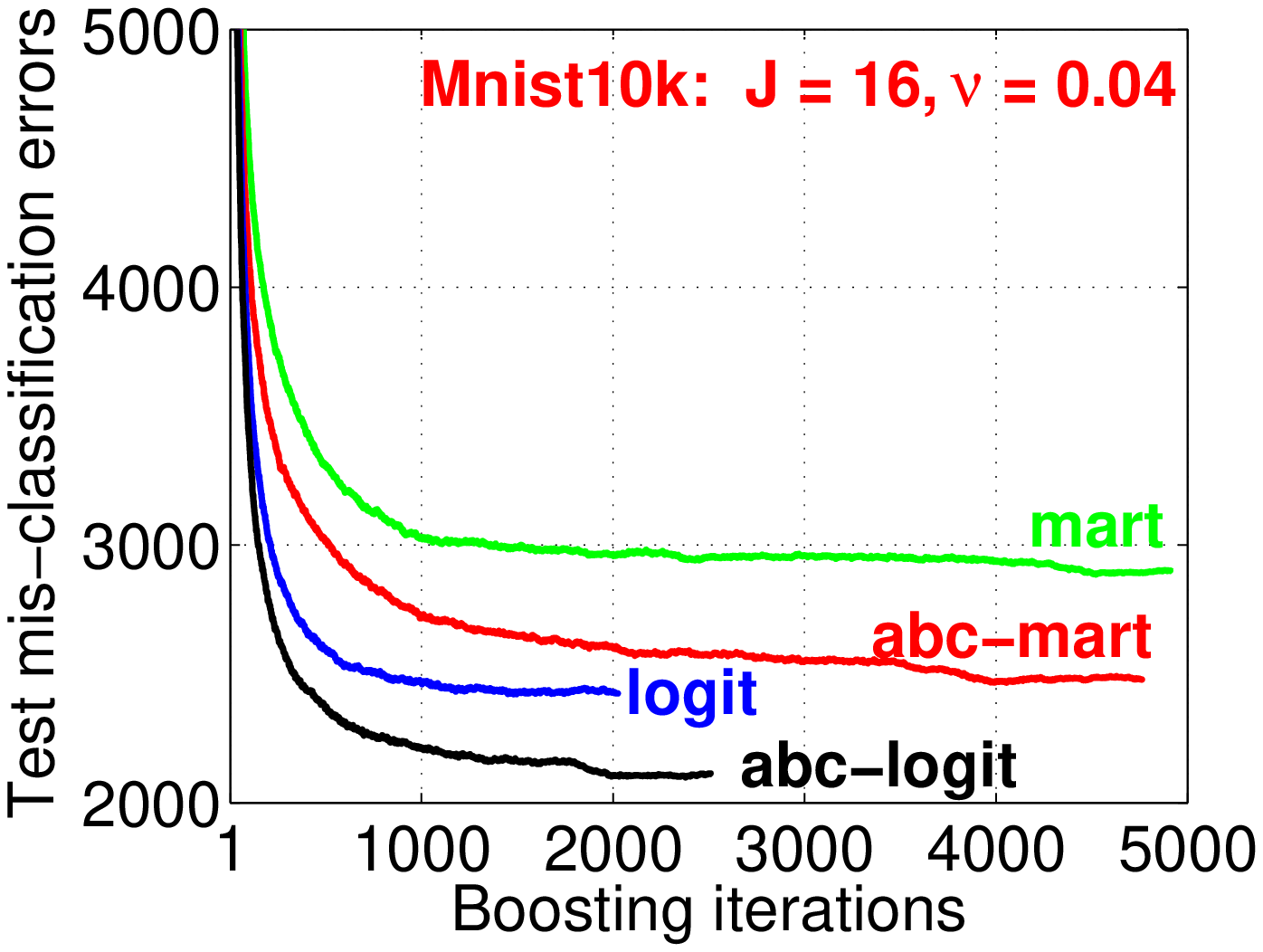}\hspace{-0.05in}
\includegraphics[width=2.2in]{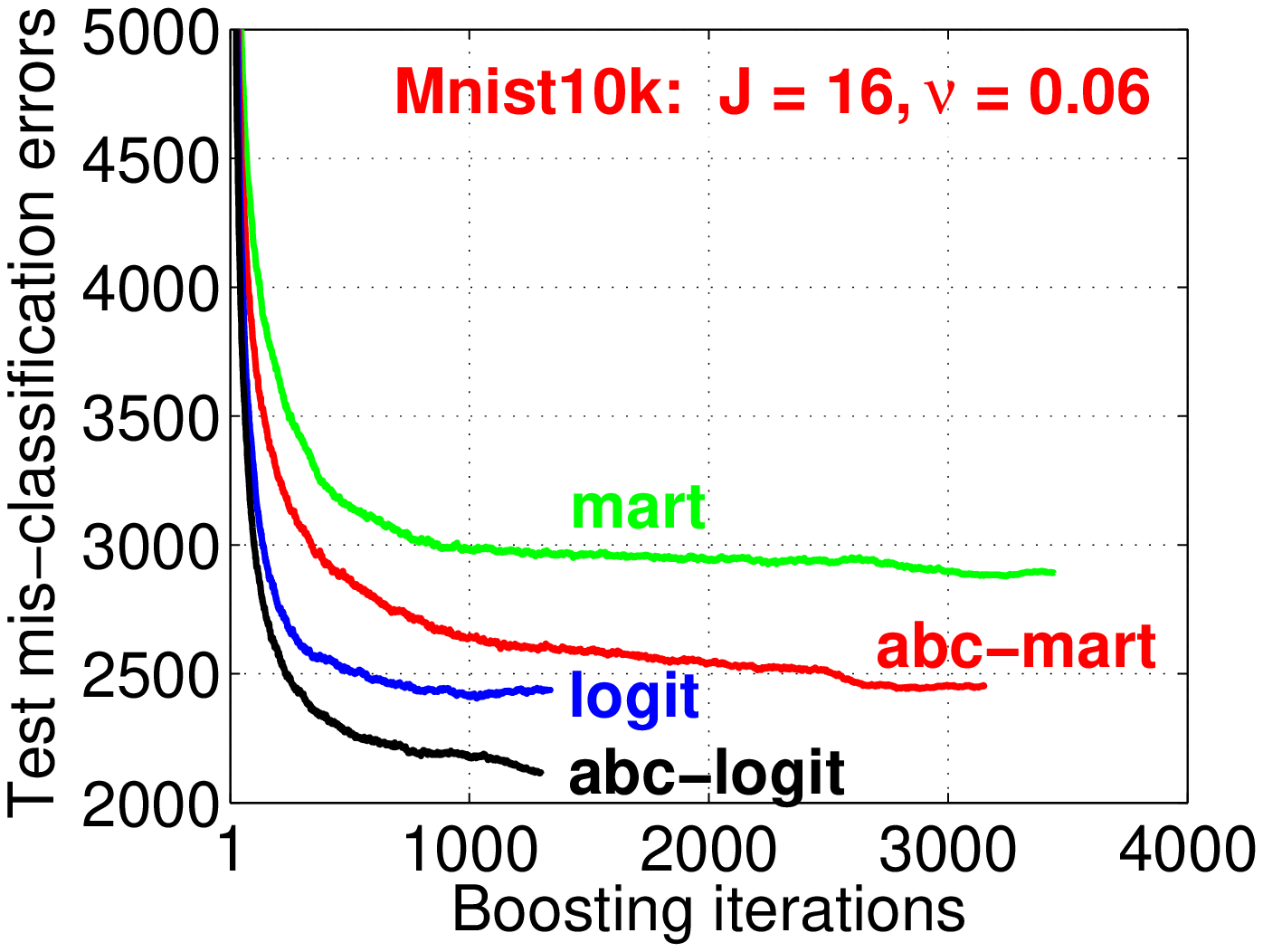}\hspace{-0.05in}
\includegraphics[width=2.2in]{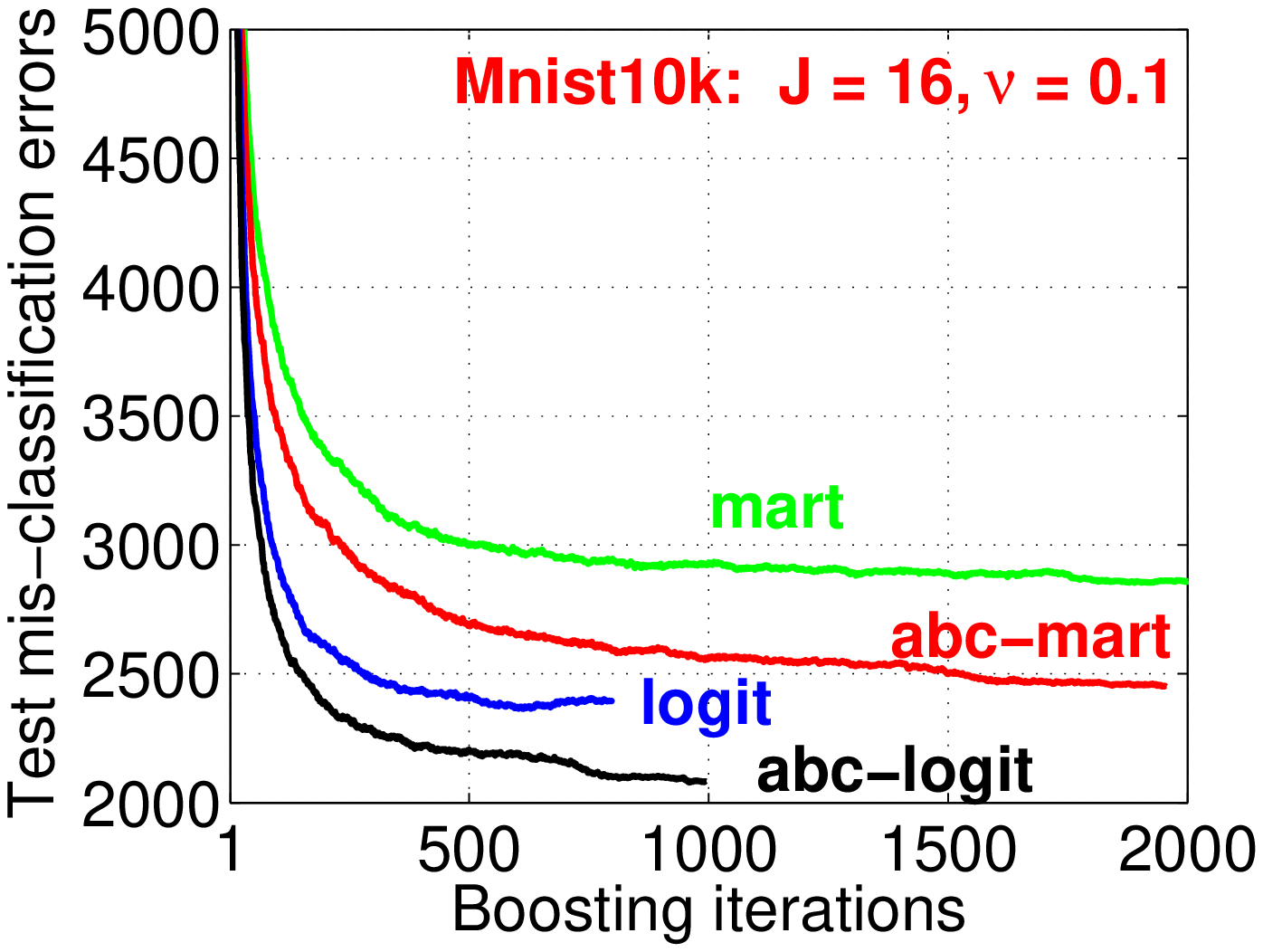}
}
\mbox{
\includegraphics[width=2.2in]{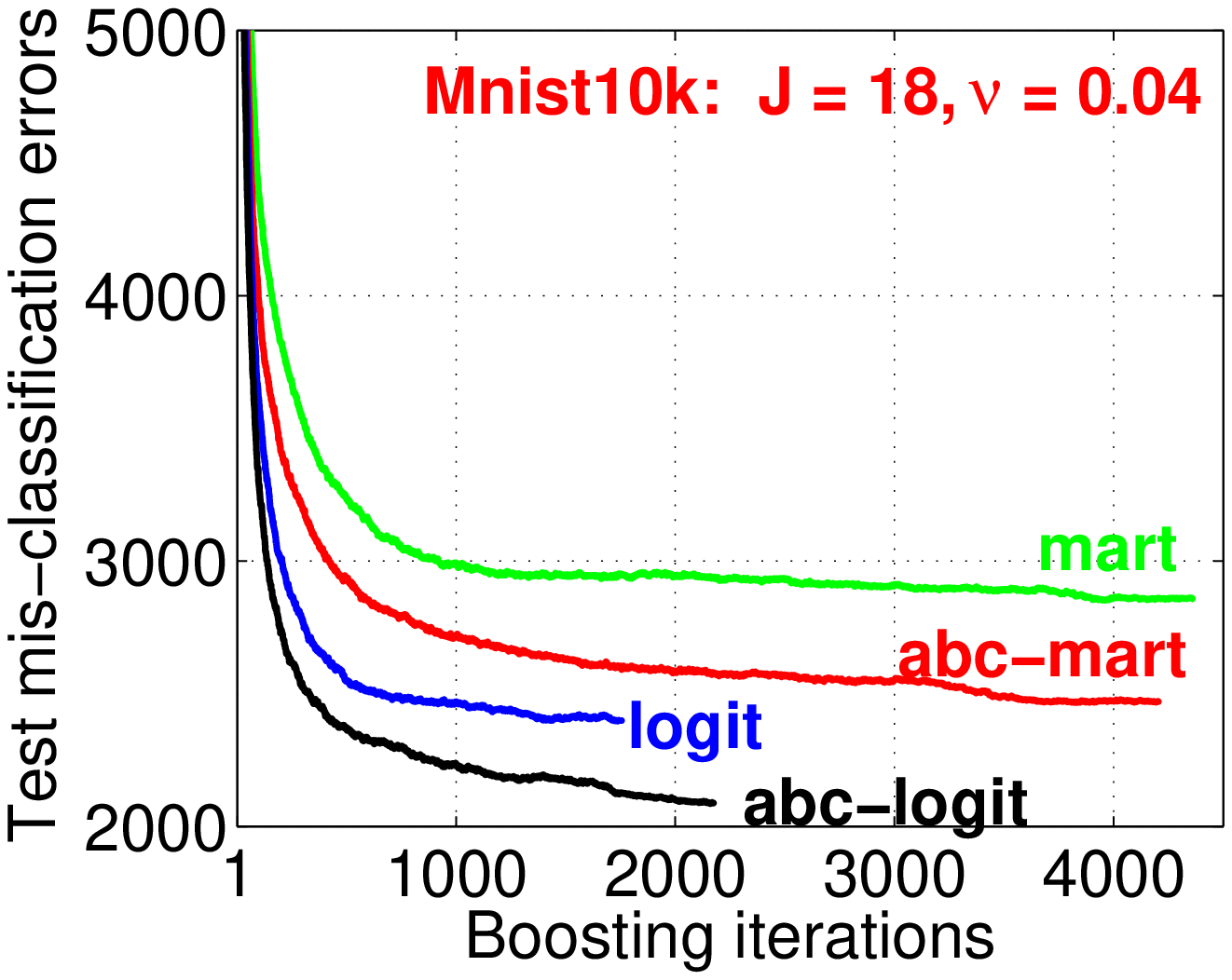}\hspace{-0.05in}
\includegraphics[width=2.2in]{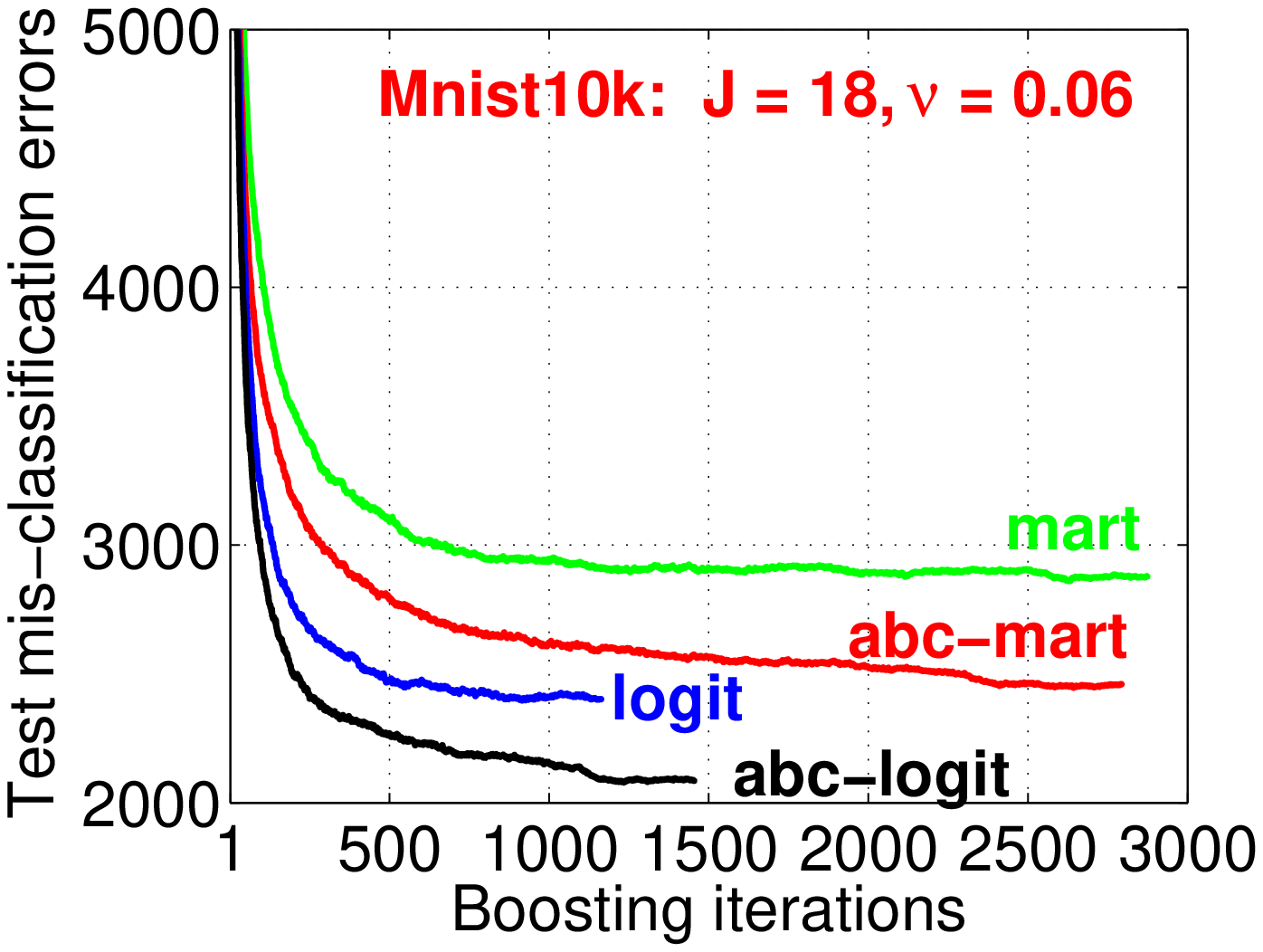}\hspace{-0.05in}
\includegraphics[width=2.2in]{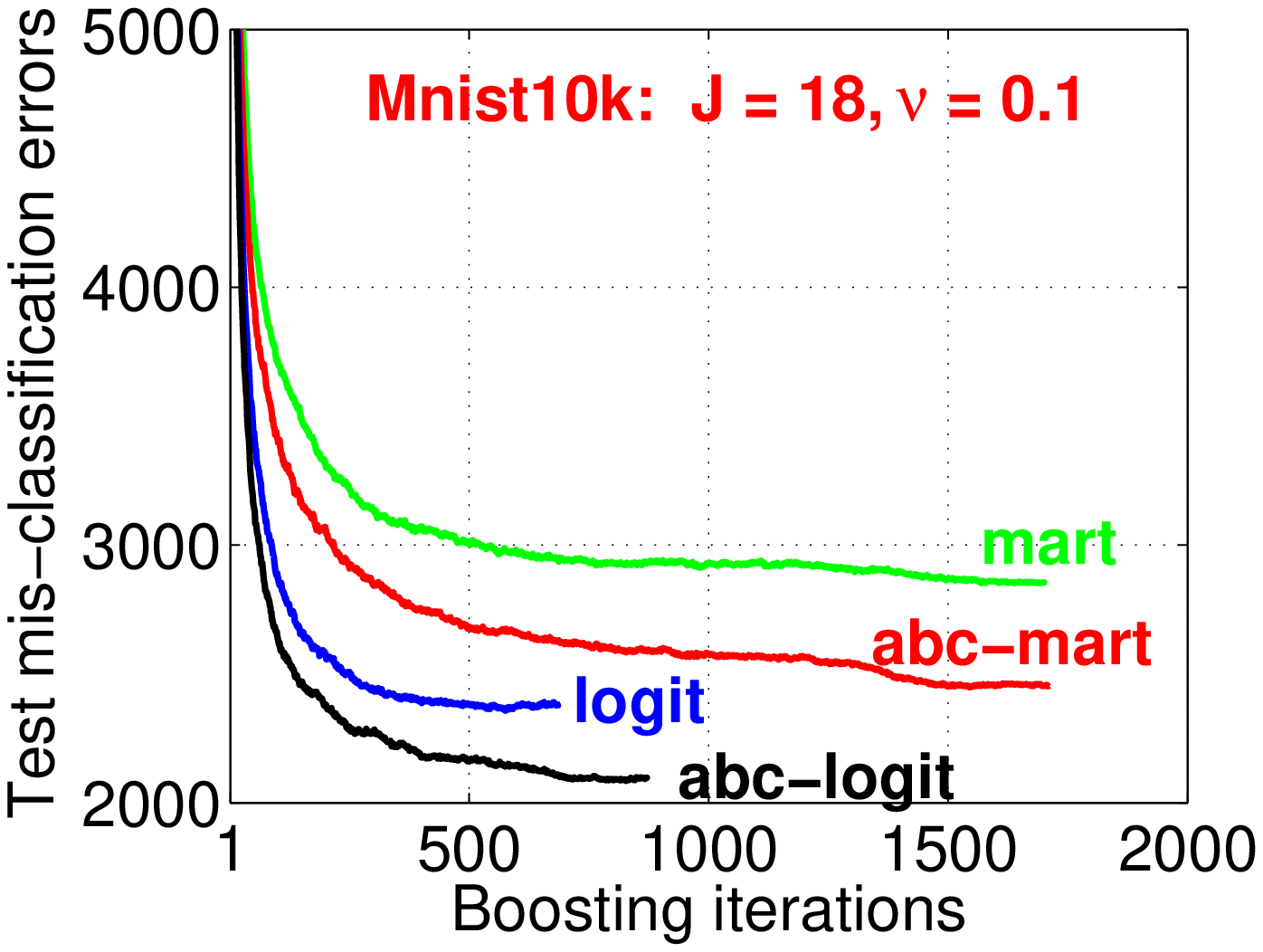}
}
\end{center}
\vspace{-0.1in}
\caption{\textbf{\em Mnist10k}. Test mis-classification errors of four algorithms.  $J=12$, 14, 16, 18. }\label{fig_Mnist10k_12-18}
\end{figure}

\begin{figure}[h]
\begin{center}
\mbox{
\includegraphics[width=2.2in]{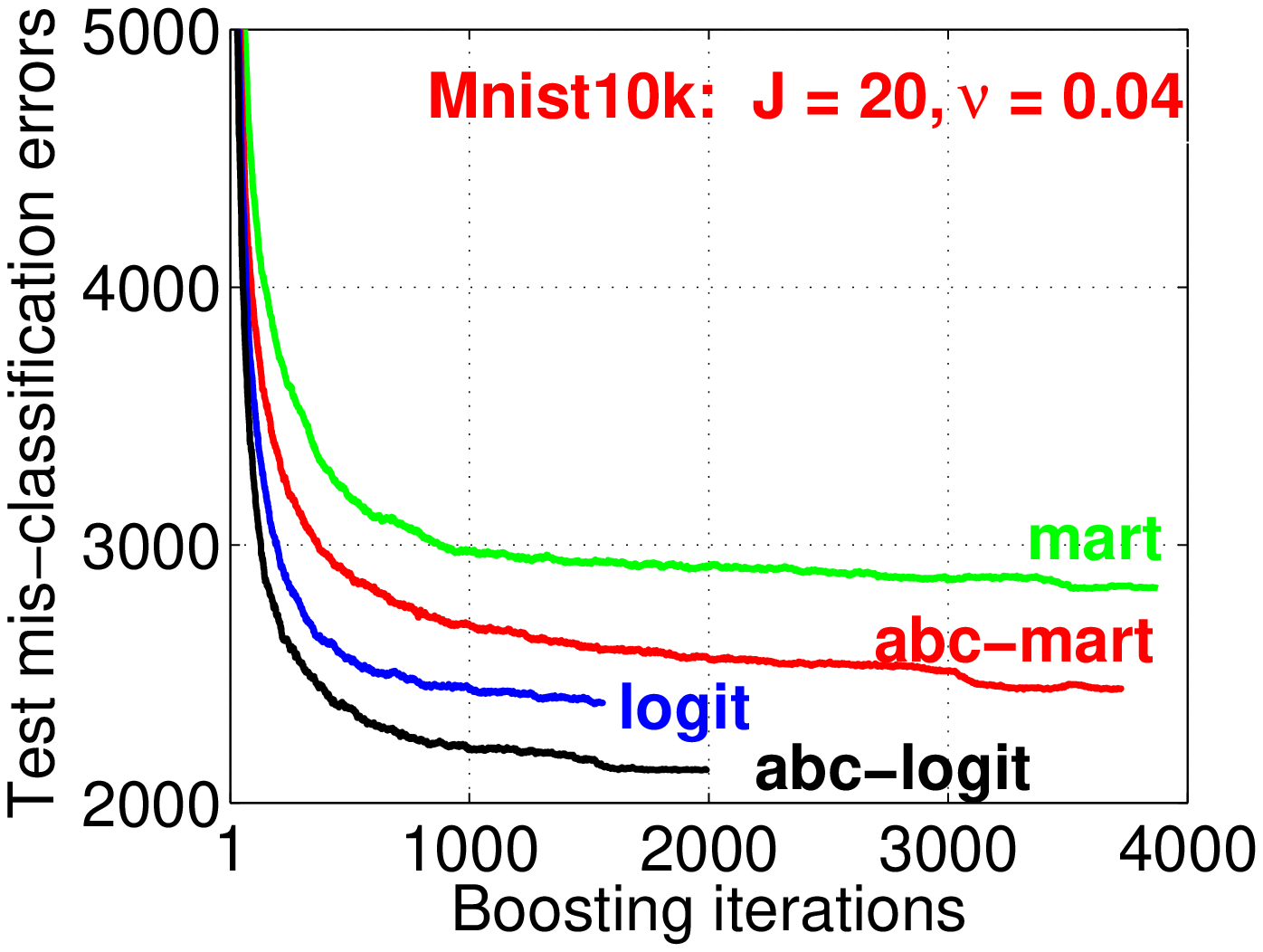}\hspace{-0.05in}
\includegraphics[width=2.2in]{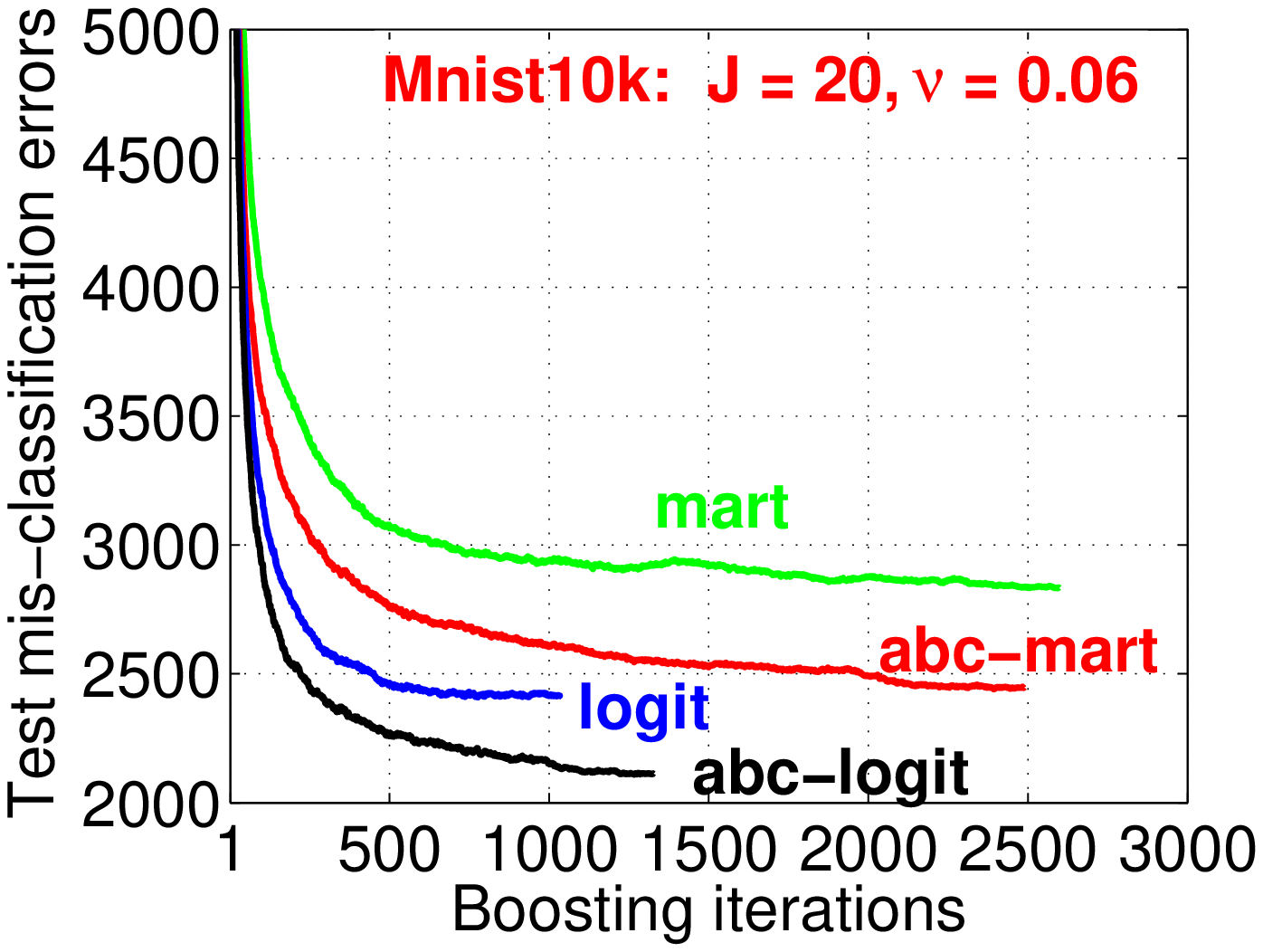}\hspace{-0.05in}
\includegraphics[width=2.2in]{Mnist10kTest4J20v01.eps}
}
\mbox{
\includegraphics[width=2.2in]{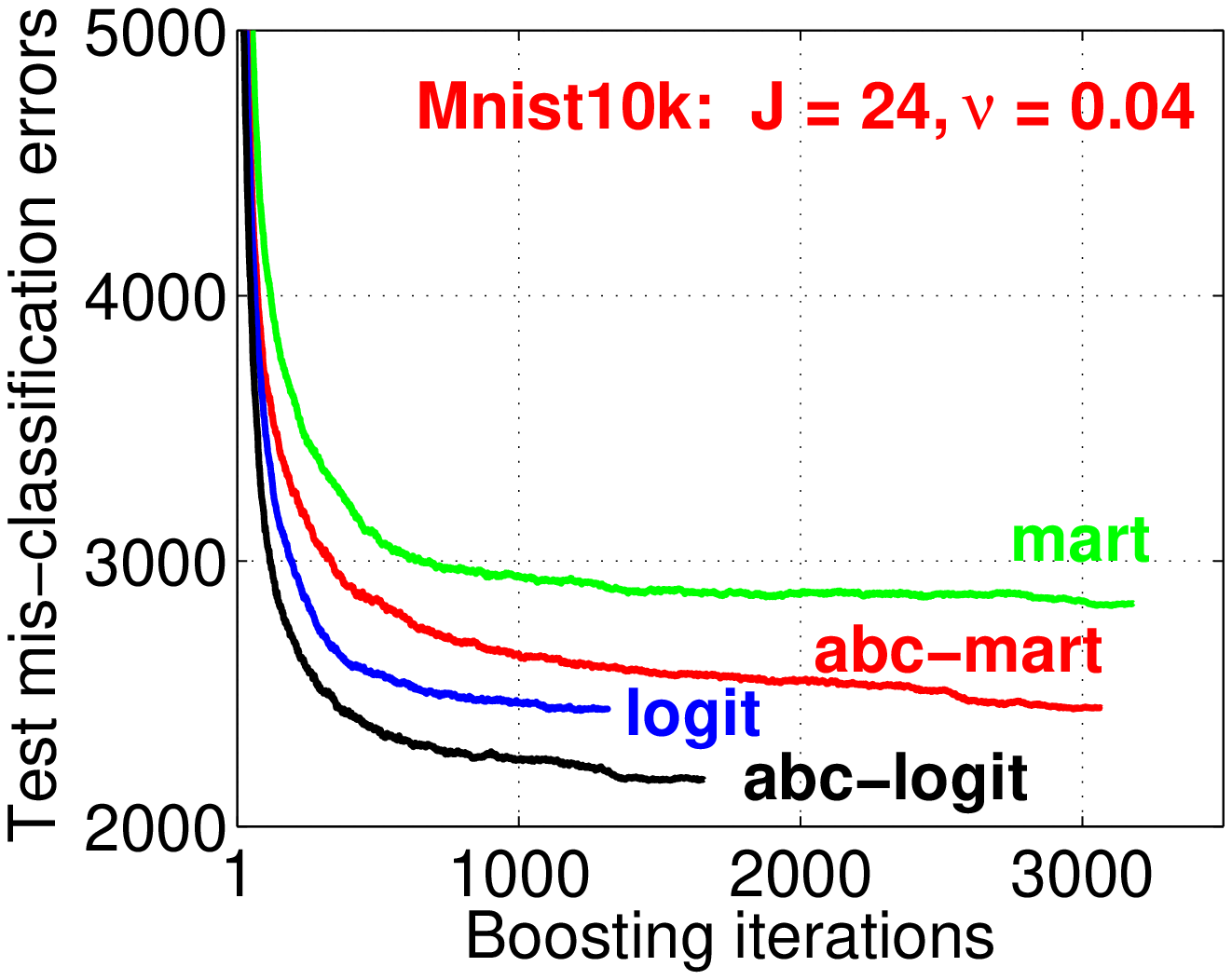}\hspace{-0.05in}
\includegraphics[width=2.2in]{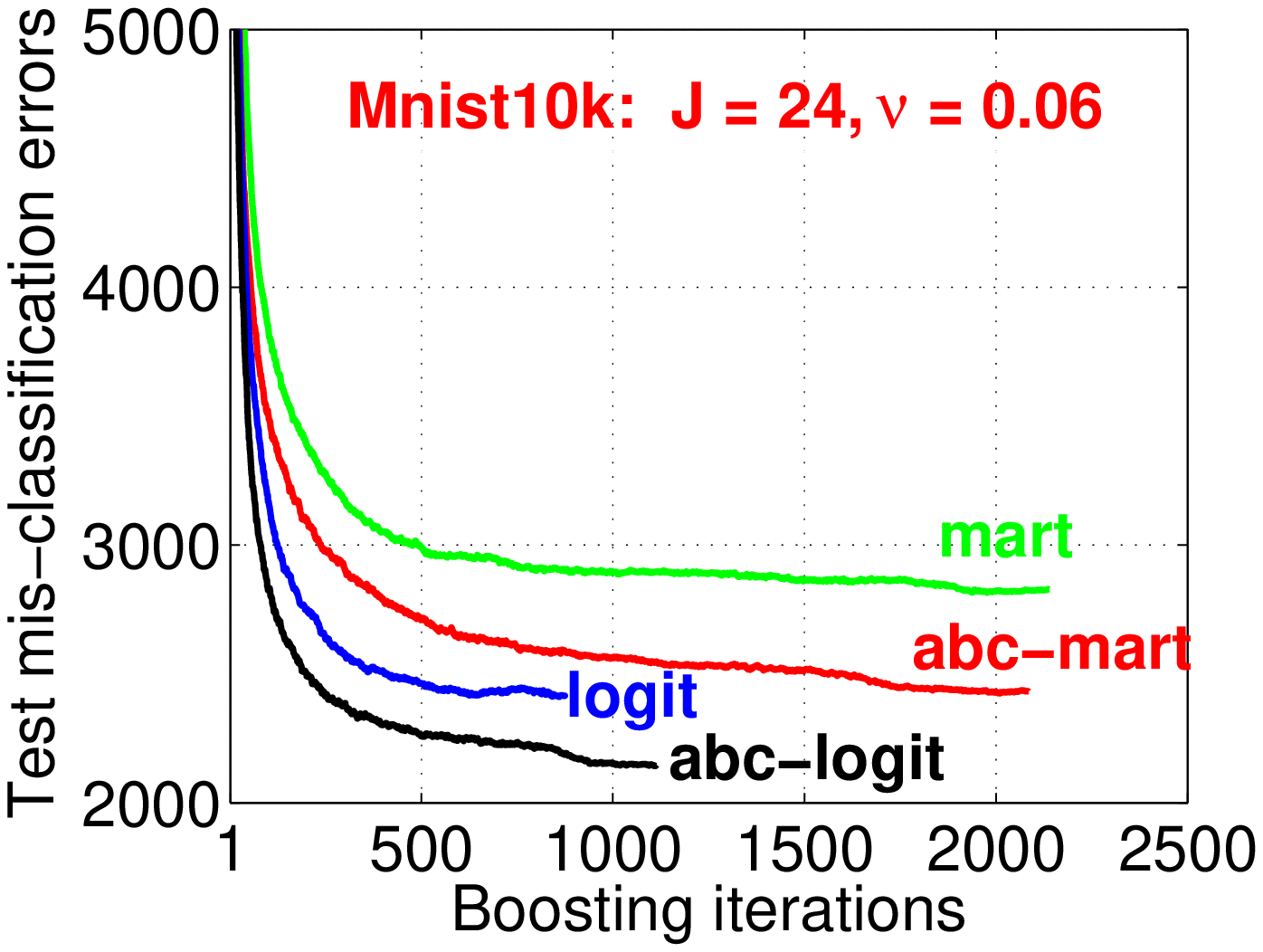}\hspace{-0.05in}
\includegraphics[width=2.2in]{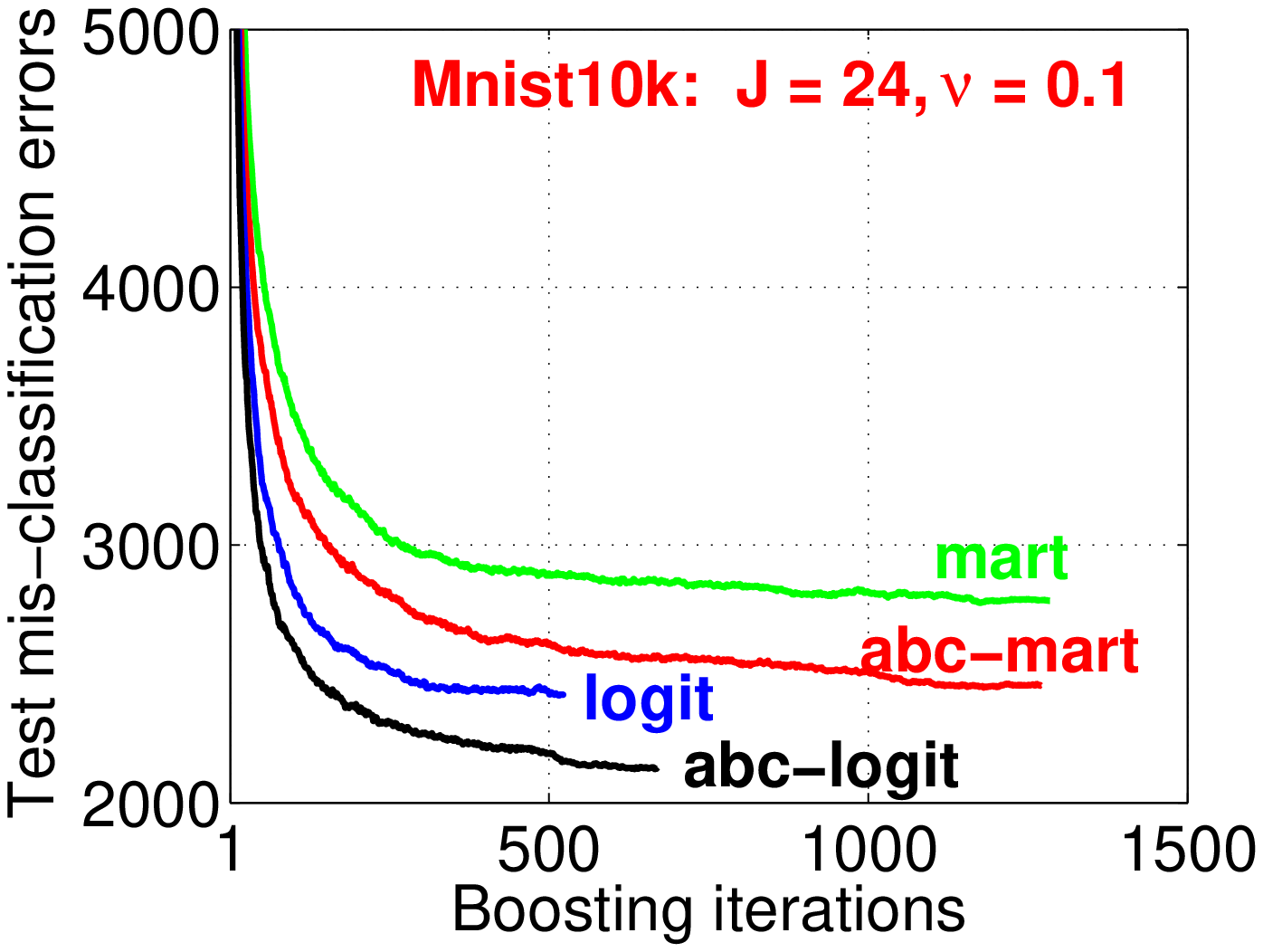}
}
\mbox{
\includegraphics[width=2.2in]{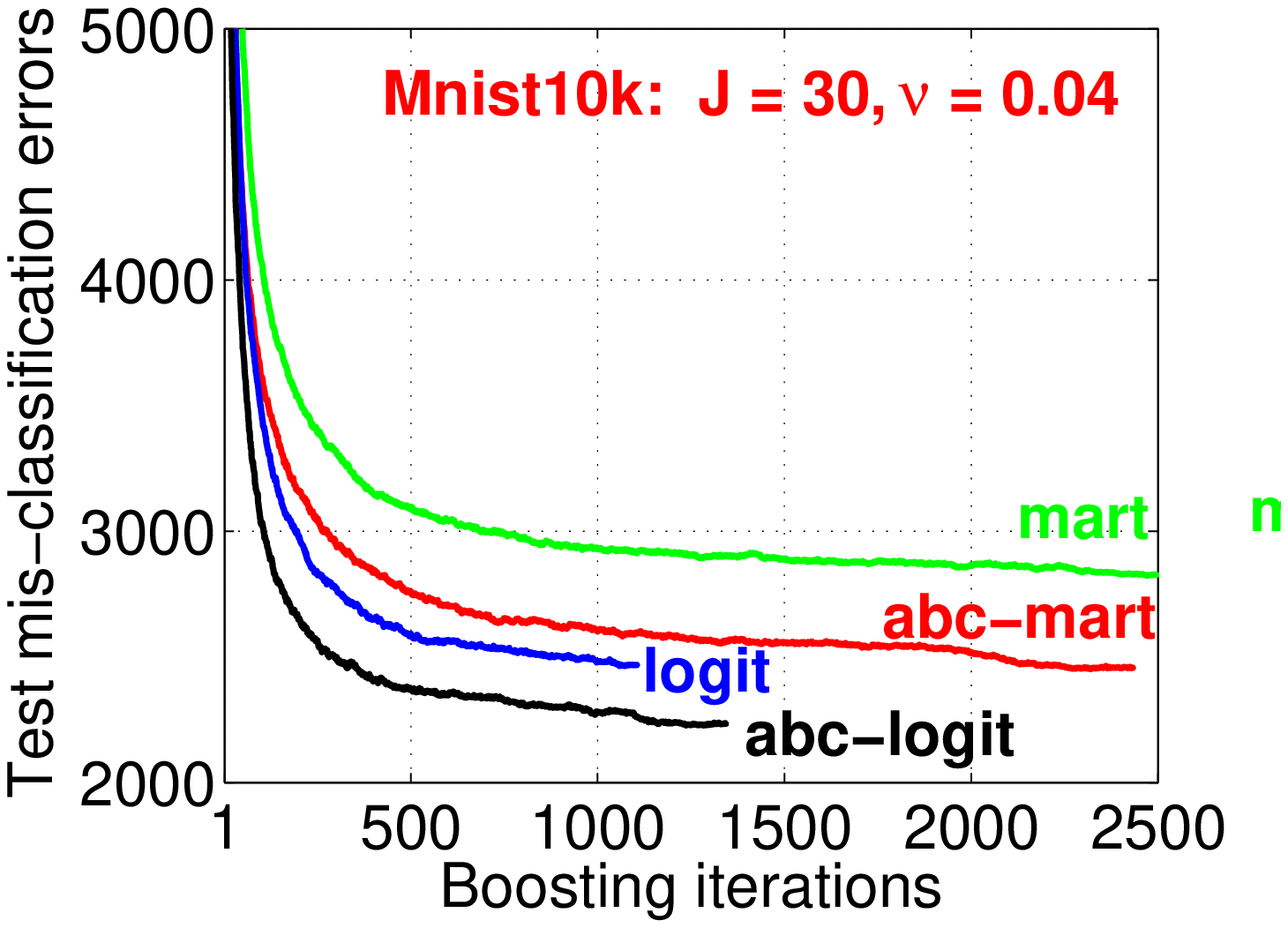}\hspace{-0.05in}
\includegraphics[width=2.2in]{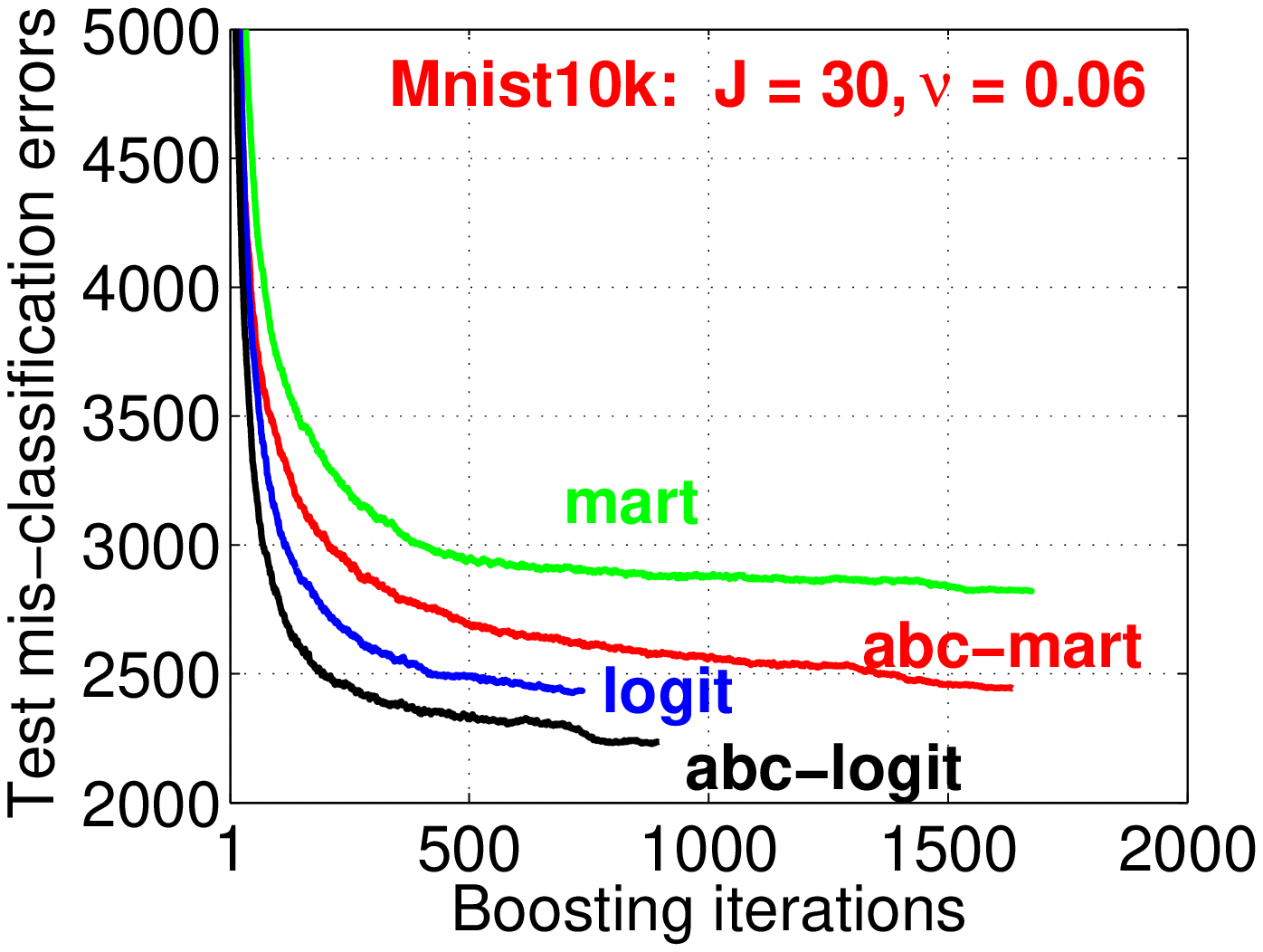}\hspace{-0.05in}
\includegraphics[width=2.2in]{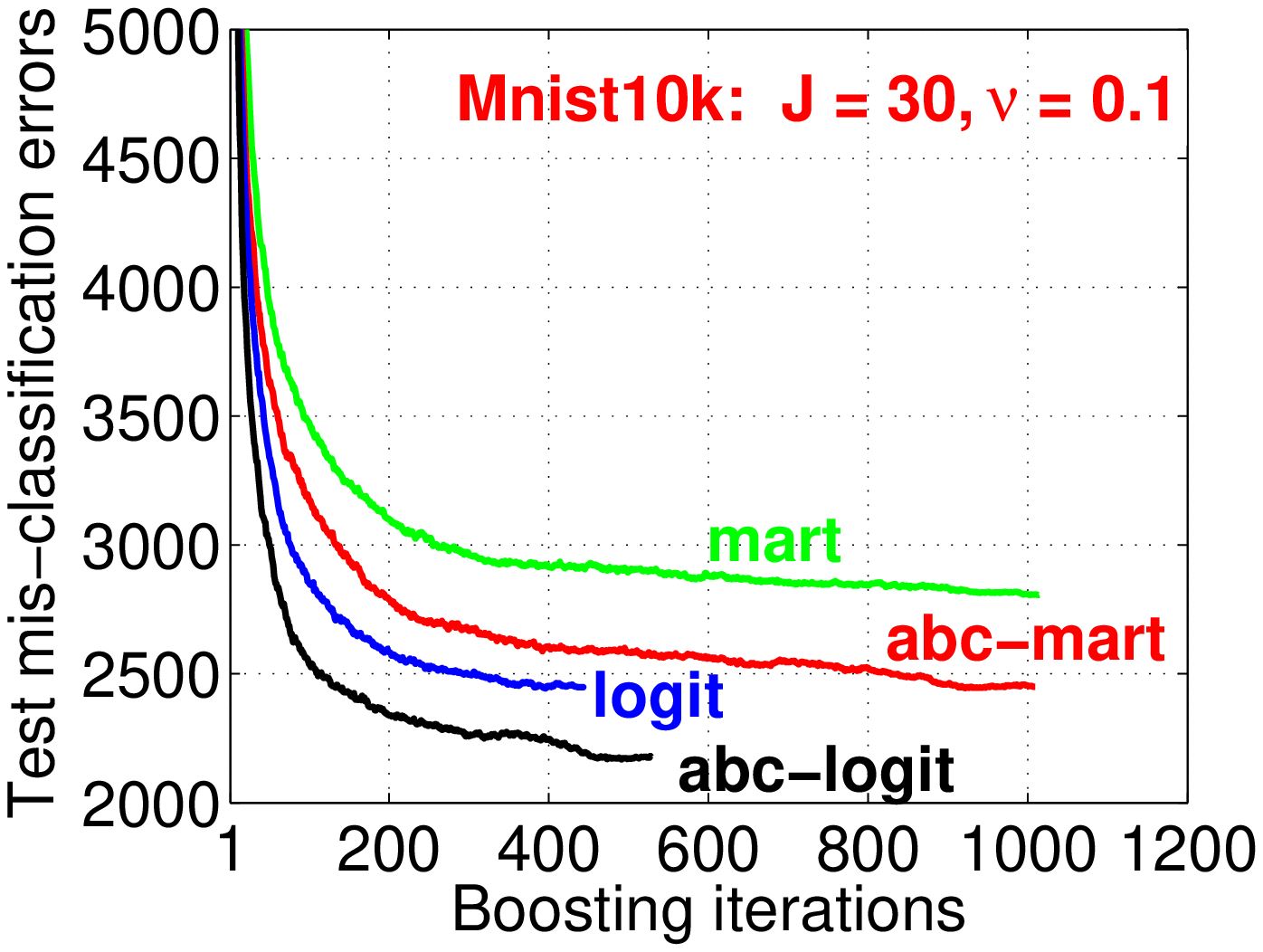}
}
\mbox{
\includegraphics[width=2.2in]{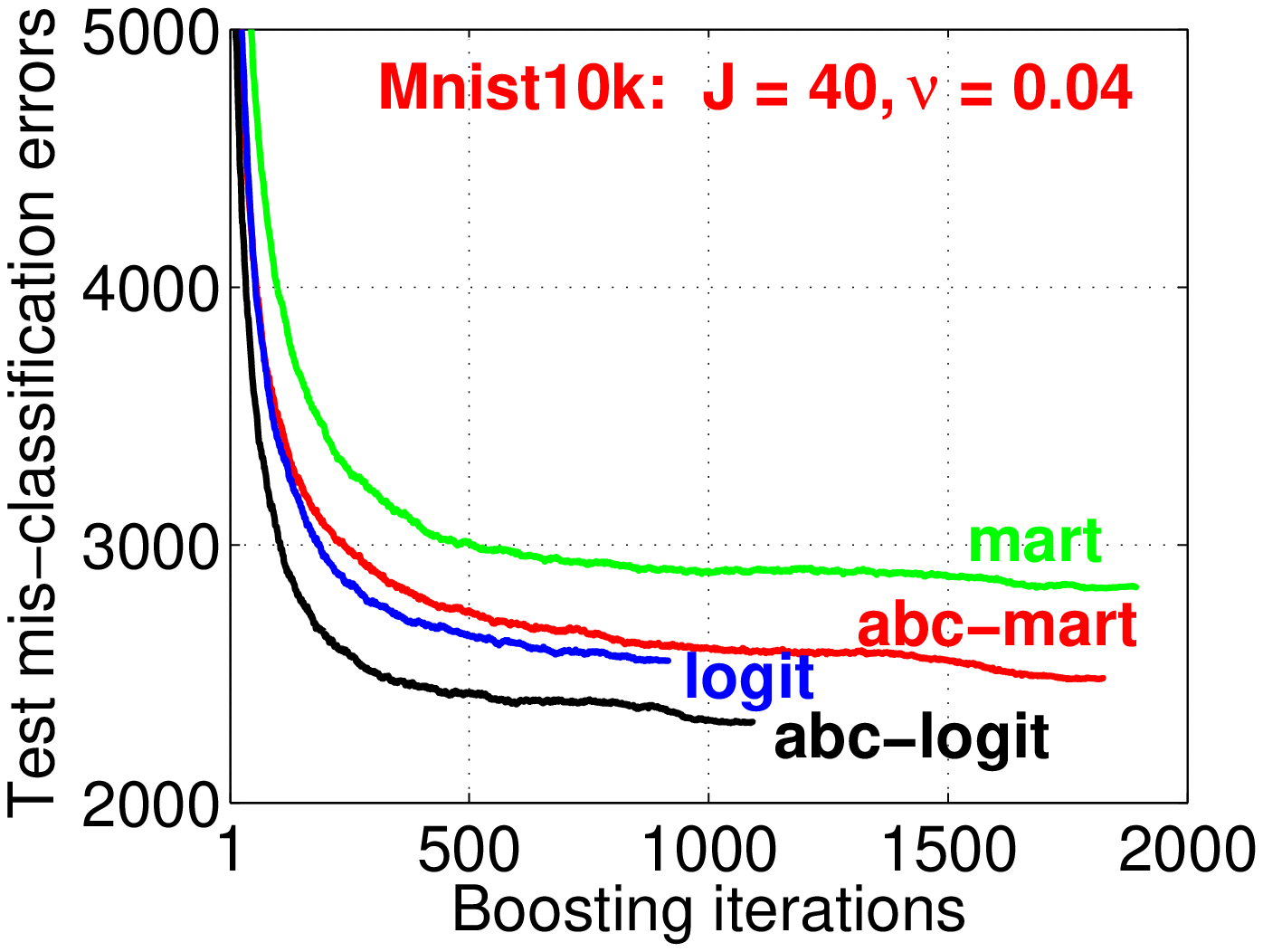}\hspace{-0.05in}
\includegraphics[width=2.2in]{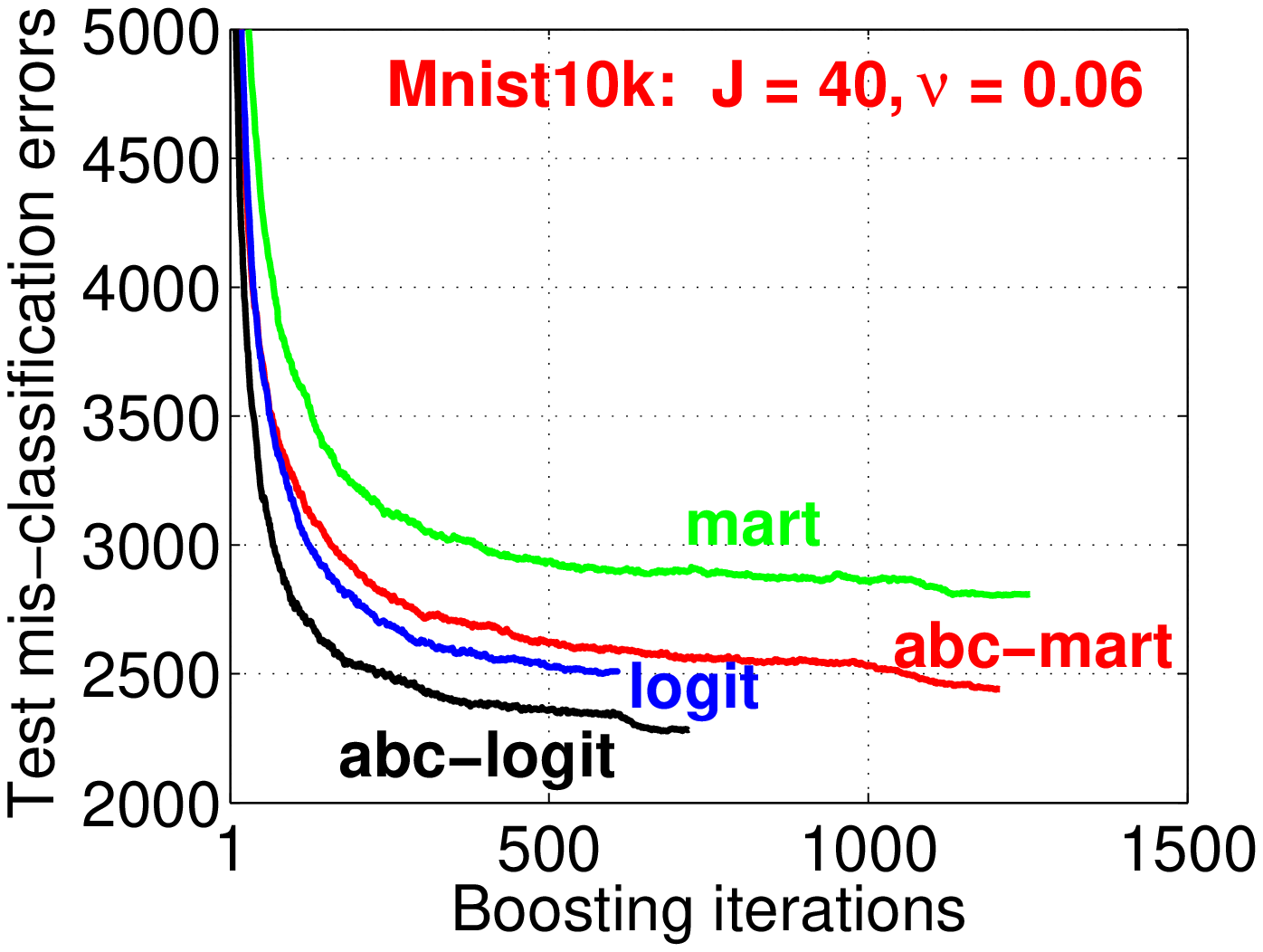}\hspace{-0.05in}
\includegraphics[width=2.2in]{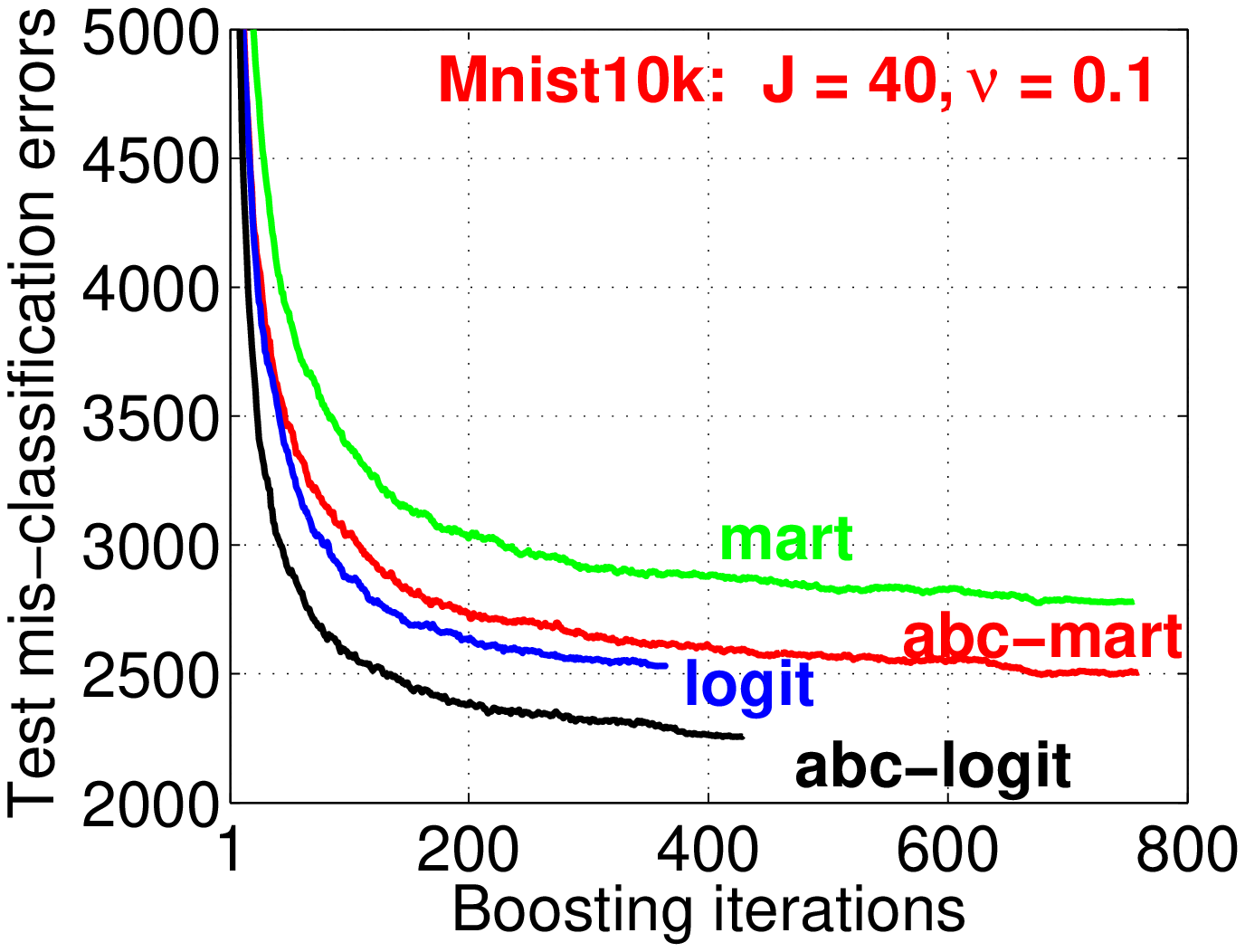}
}
\mbox{
\includegraphics[width=2.2in]{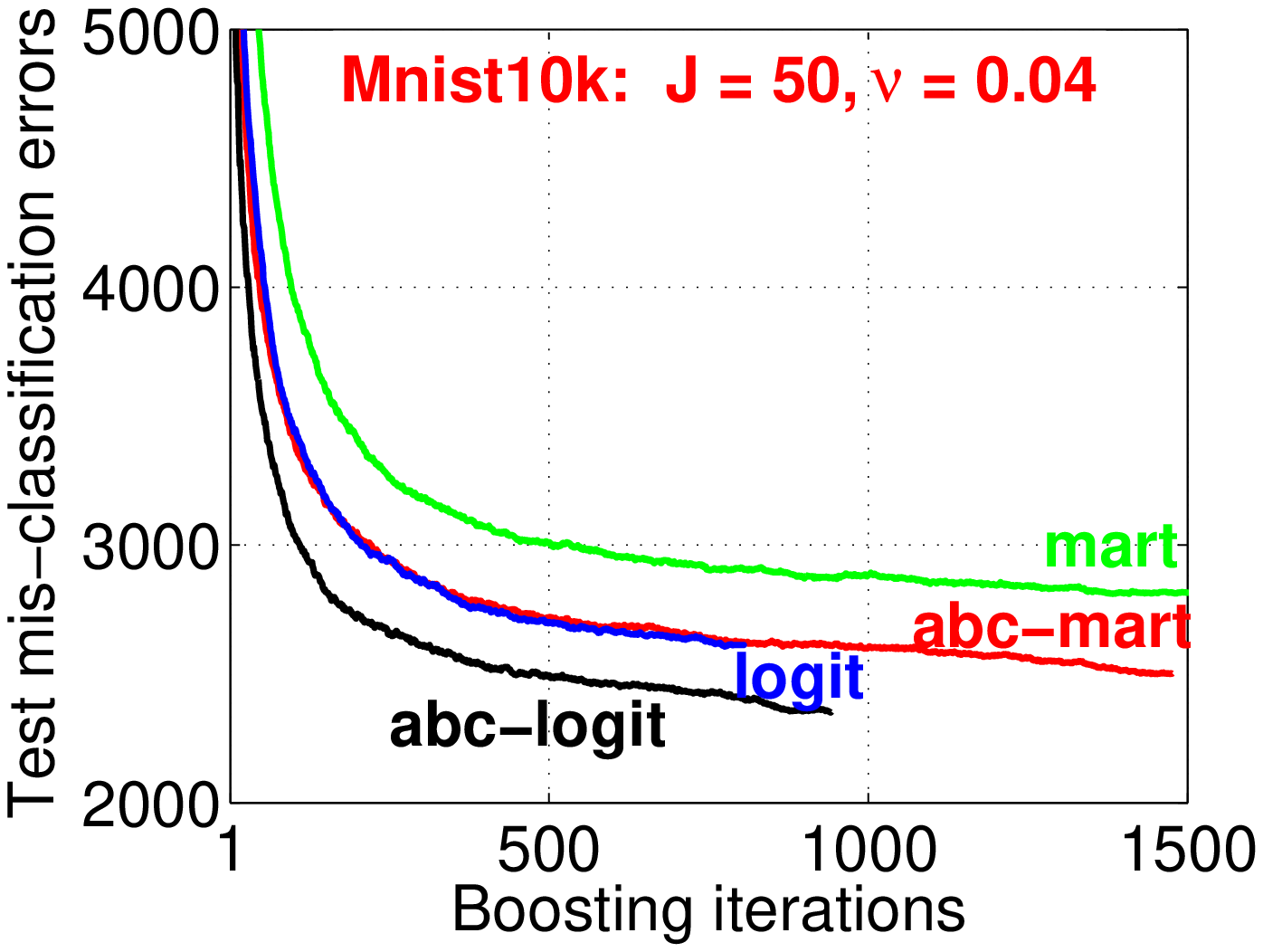}\hspace{-0.05in}
\includegraphics[width=2.2in]{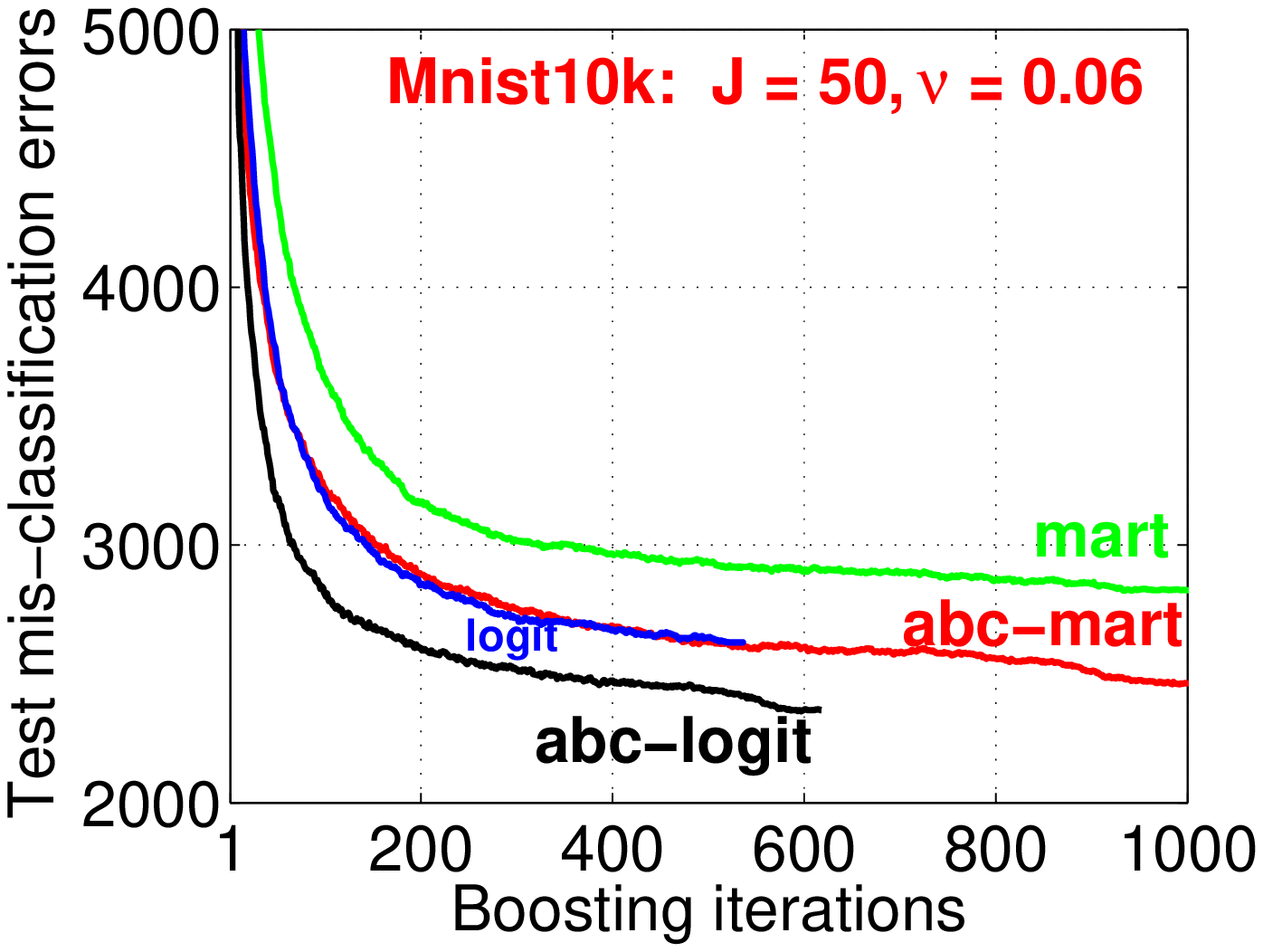}\hspace{-0.05in}
\includegraphics[width=2.2in]{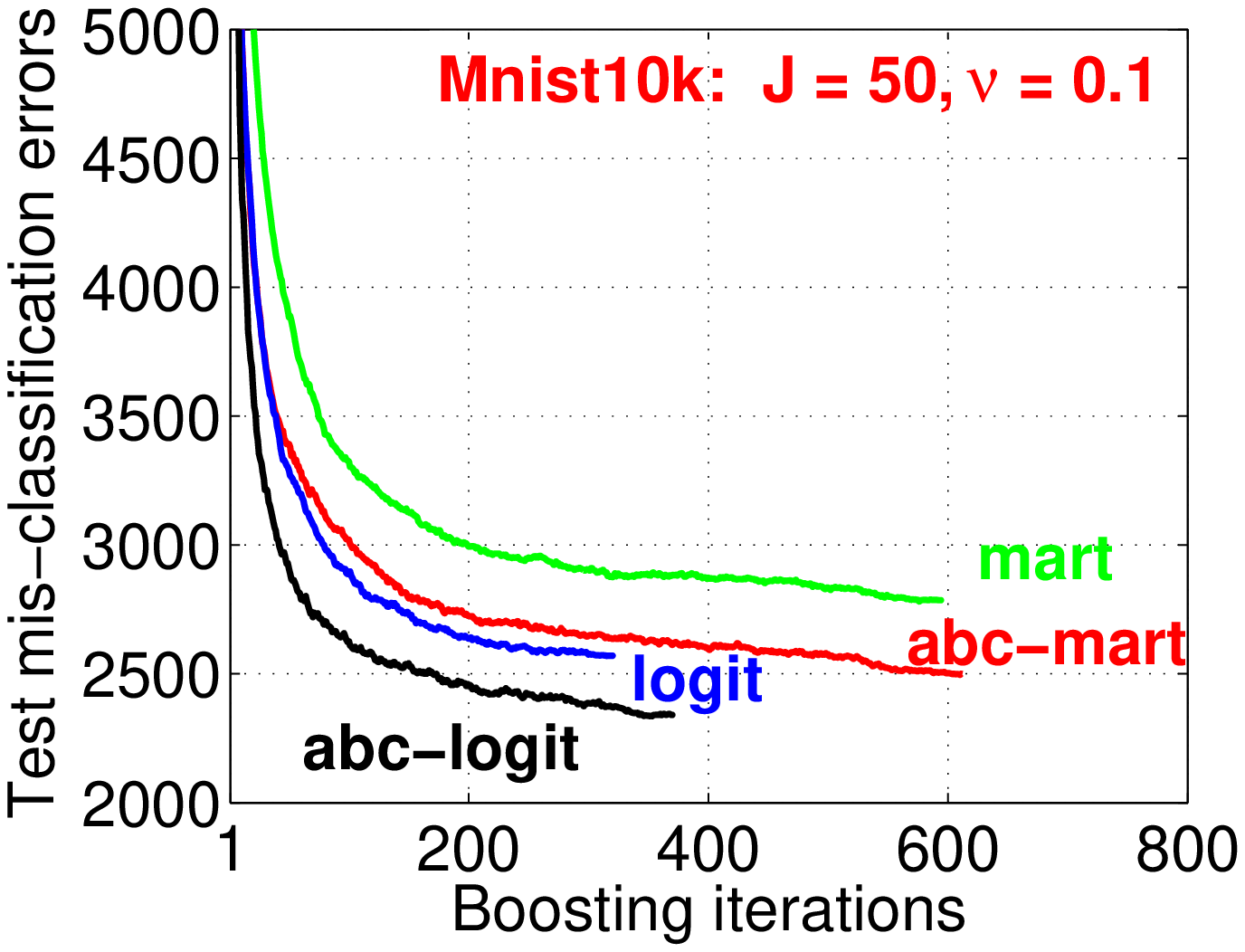}
}
\end{center}
\vspace{-0.1in}
\caption{\textbf{\em Mnist10k}. Test mis-classification errors of four algorithms.  $J=20$, 24, 30, 40, 50.}\label{fig_Mnist10k_20-50}
\end{figure}

\clearpage

The experiment results illustrate that the performances of all four algorithms are  stable on a wide-range of base class tree sizes $J$, e.g., $J\in[6,30]$. The shrinkage parameter $\nu$ does not affect much the test performance, although smaller $\nu$ values  result in more boosting iterations (before the training losses reach the machine accuracy).

We further randomly divide the test set of {\em Mnist10k} (60000 test samples) equally into two parts (I and II). We then test algorithms on Part I (using the same training results). We name this ``new'' dataset {\em Mnist10kT1}. The purpose of this experiment is to further demonstrate the stability of the algorithms.

Table \ref{tab_Mnist10kT1} presents the test mis-classification errors of {\em Mnist10kT1}. Compared to Table \ref{tab_Mnist10k}, the mis-classification errors of {\em Mnist10kT1} are roughly $50\%$ of the mis-classification errors of {\em Mnist10k} for all $J$ and $\nu$. This helps establish that our experiment results on {\em Mnist10k} provide a very reliable comparison.

\begin{table}[h]
\caption{\textbf{\em Mnist10kT1}. Upper table: The test mis-classification errors of  {\em mart} and \textbf{\em abc-mart} (bold numbers). Bottom table: The test mis-classification errors of  {\em logitboost} and \textbf{\em abc-logitboost} (bold numbers). {\em Mnist10kT1} only uses a half of the test data of {\em Mnist10k}.}
\begin{center}
{\small
{\begin{tabular}{l l l l l }
\hline \hline

 &{\em mart} & \textbf{\em abc-mart}\\\hline
  &$\nu = 0.04$ &$\nu=0.06$ &$\nu=0.08$ &$\nu=0.1$ \\
\hline

$J=4$        &1682        \textbf{1514}        &1668        \textbf{1505}        &1666        \textbf{1416}        &1663        \textbf{1380}\\
$J=6$         &1573        \textbf{1382}        &1523        \textbf{1320}        &1533        \textbf{1329}        &1582        \textbf{1288}\\
$J=8$         &1501        \textbf{1263}        &1515        \textbf{1257}        &1523        \textbf{1250}        &1491        \textbf{1279}\\
$J=10$         &1492        \textbf{1270}        &1457        \textbf{1248}        &1470        \textbf{1239}        &1459        \textbf{1236}\\
$J=12$         &1432        \textbf{1244}        &1427        \textbf{1234}        &1444        \textbf{1228}       &1436       \textbf{1227}\\
$J=14$         &1424        \textbf{1237}        &1420        \textbf{1231}        &1407        \textbf{1223}       &1419        \textbf{1212}\\
$J=16$         &1430        \textbf{1226}        &1426        \textbf{1224}        &1411        \textbf{1223}       &1418        \textbf{1204}\\
$J=18$         &1400        \textbf{1222}        &1413        \textbf{1218}        &1390        \textbf{1210}        &1404 \textbf{1211}\\
$J=20$         &1398        \textbf{1213}        &1381        \textbf{1205}        &1388        \textbf{1213} &1382 \textbf{1198}\\
$J=24$         &1402        \textbf{1221}        &1366        \textbf{1201}        &1372        \textbf{1199}        &1346        \textbf{1205}\\
$J=30$         &1384        \textbf{1211}        &1374        \textbf{1208}        &1368        \textbf{1224}        &1366        \textbf{1205}\\
$J=40$         &1397        \textbf{1244}        &1375        \textbf{1220}        &1397        \textbf{1222}        &1365        \textbf{1246}\\
$J=50$         &1371        \textbf{1239}        &1380        \textbf{1221}        &1382        \textbf{1223}        &1362       \textbf{1242}\\
\hline \hline
 &{\em logitboost} & \textbf{\em abc-logit}\\\hline
  &$\nu = 0.04$ &$\nu=0.06$ &$\nu=0.08$ &$\nu=0.1$ \\
\hline

$J=4$        &1419        \textbf{1299}        &1449        \textbf{1281}        &1446        \textbf{1251}        &1460        \textbf{1244}\\
$J=6$        &1313        \textbf{1111}        &1313        \textbf{1114}        &1326        \textbf{1101}       &1317       \textbf{1097}\\
$J=8$        &1278        \textbf{1058}        &1287        \textbf{1050}        &1270        \textbf{1036}       &1262        \textbf{1058}\\
$J=10$        &1252        \textbf{1061}        &1244        \textbf{1057}       &1237        \textbf{1040}       &1229        \textbf{1041}\\
$J=12$        &1224        \textbf{1020}        &1219        \textbf{1049}       &1217        \textbf{1053}       &1224        \textbf{1047}\\
$J=14$        &1213        \textbf{1038}        &1207        \textbf{1050}       &1201        \textbf{1039}       &1198        \textbf{1026}\\
$J=16$        &1185        \textbf{1050}        &1205        \textbf{1058}       &1189        \textbf{1044}       &1178        \textbf{1041}\\
$J=18$        &1186        \textbf{1048}        &1184        \textbf{1038}       &1184        \textbf{1046}       &1167        \textbf{1056}\\
$J=20$        &1185        \textbf{1077}        &1199        \textbf{1063}       &1183        \textbf{1042}       &1184        \textbf{1045}\\
$J=24$        &1208        \textbf{1095}        &1196        \textbf{1083}       &1191       \textbf{1064}      &1194      \textbf{1068}\\
$J=30$       & 1225        \textbf{1113}        &1201        \textbf{1117}       &1190        \textbf{1113}        &1211 \textbf{1087}\\
$J=40$       & 1254        \textbf{1159}        &1247        \textbf{1145}       &1248        \textbf{1127}        &1249        \textbf{1127}\\
$J=50$       & 1292        \textbf{1177}        &1284        \textbf{1174}       &1275        \textbf{1161}        &1276        \textbf{1176}\\
\hline\hline
\end{tabular}}}
\end{center}
\label{tab_Mnist10kT1}
\end{table}

\clearpage

\subsection{Detailed Experiment Results on {\em Poker25kT1} and {\em Poker25kT2} }

Recall the original UCI {\em Poker} dataset used 25010 samples for training and 1000000 samples for testing. To provide a reliable comparison (and validation), we form two datasets {\em Poker25kT1} and {\em Poker25kT2} by equally dividing the original test set into two parts (I and II). Both use the same training set. {\em Poker25kT1} uses Part I of the original test set for testing and {\em Poker25kT2} uses Part II for testing.

Table \ref{tab_Poker25kT1} and Table \ref{tab_Poker25kT2} present the test mis-classification errors, for $J \in \{4, 6, 8, 10, 12, 14, 16, 18, 20\}$ and $\nu \in \{0.04, 0.06, 0.08, 0.1\}$. Comparing these two tables, we can see the corresponding entries are very close to each other, which again verifies that the four boosting algorithms provide reliable results on this dataset.

For most $J$ and $\nu$, all four algorithms achieve error rates $<10\%$.  For both {\em Poker25kT1} and {\em Poker25kT2}, the lowest test errors are attained at $\nu = 0.1$ and $J=6$. Unlike {\em Mnist10k}, the test errors, especially using {\em mart} and {\em logitboost}, are slightly  sensitive to the parameters.

Note that when $J=4$ (and $\nu$ is small), only training $M=10000$ steps would not be sufficient in this case.

\begin{table}[h]
\caption{\textbf{\em Poker25kT1}. Upper table: The test mis-classification errors of  {\em mart} and \textbf{\em abc-mart} (bold numbers). Bottom table: The test mis-classification errors of  {\em logitboost} and \textbf{\em abc-logitboost} (bold numbers)}
\begin{center}
{\small
{\begin{tabular}{l r r r r }
\hline \hline
 &{\em mart} & \textbf{\em abc-mart}\\\hline
  &$\nu = 0.04$ &$\nu=0.06$ &$\nu=0.08$ &$\nu=0.1$ \\
\hline

$J=4$    &145880 \textbf{90323} &132526 \textbf{67417} &124283 \textbf{49403} &113985 \textbf{42126}\\
$J=6$    &71628  \textbf{38017} &59046 \textbf{36839} &48064 \textbf{35467} &43573 \textbf{34879}\\
$J=8$    &64090  \textbf{39220} &53400 \textbf{37112} &47360 \textbf{36407} &44131 \textbf{35777}\\
$J=10$   &60456  \textbf{39661} &52464 \textbf{38547} &47203 \textbf{36990} &46351 \textbf{36647}\\
$J=12$   &61452  \textbf{41362} &52697 \textbf{39221} &46822 \textbf{37723} &46965 \textbf{37345}\\
$J=14$   &58348  \textbf{42764} &56047 \textbf{40993} &50476 \textbf{40155} &47935 \textbf{37780}\\
$J=16$   &63518  \textbf{44386} &55418 \textbf{43360} &50612 \textbf{41952} &49179 \textbf{40050}\\
$J=18$   &64426  \textbf{46463} &55708 \textbf{45607} &54033 \textbf{45838} &52113 \textbf{43040}\\
$J=20$   &65528  \textbf{49577} &59236 \textbf{47901} &56384 \textbf{45725} &53506 \textbf{44295}\\

\hline \hline
 &{\em logitboost} & \textbf{\em abc-logit}\\\hline
  &$\nu = 0.04$ &$\nu=0.06$ &$\nu=0.08$ &$\nu=0.1$ \\
\hline

$J=4$    &147064 \textbf{102905} &140068 \textbf{71450} &128161 \textbf{51226} &117085 \textbf{42140}\\
$J=6$    &81566  \textbf{43156} &59324 \textbf{39164} &51526 \textbf{37954} &48516 \textbf{37546}\\
$J=8$    &68278  \textbf{46076} &56922 \textbf{40162} &52532 \textbf{38422} &46789 \textbf{37345}\\
$J=10$   &63796  \textbf{44830} &55834 \textbf{40754} &53262 \textbf{40486} &47118 \textbf{38141}\\
$J=12$   &66732  \textbf{48412} &56867 \textbf{44886} &51248 \textbf{42100} &47485 \textbf{39798}\\
$J=14$   &64263  \textbf{52479} &55614 \textbf{48093} &51735 \textbf{44688} &47806 \textbf{43048}\\
$J=16$   &67092  \textbf{53363} &58019 \textbf{51308} &53746 \textbf{47831} &51267 \textbf{46968}\\
$J=18$   &69104  \textbf{57147} &56514 \textbf{55468} &55290 \textbf{50292} &51871 \textbf{47986}\\
$J=20$   &68899  \textbf{62345} &61314 \textbf{57677} &56648 \textbf{53696} &51608 \textbf{49864}

\\\hline\hline
\end{tabular}}}
\end{center}
\label{tab_Poker25kT1}
\end{table}

\begin{table}[h]
\caption{\textbf{\em Poker25kT2}. Upper table: The test mis-classification errors of  {\em mart} and \textbf{\em abc-mart} (bold numbers). Bottom table: The test mis-classification errors of  {\em logitboost} and \textbf{\em abc-logitboost} (bold numbers)}
\begin{center}
{\small
{\begin{tabular}{l r r r r }
\hline \hline
 &{\em mart} & \textbf{\em abc-mart}\\\hline
  &$\nu = 0.04$ &$\nu=0.06$ &$\nu=0.08$ &$\nu=0.1$ \\
\hline

$J=4$    &144020 \textbf{89608} &131243 \textbf{67071} &123031 \textbf{48855} &113232 \textbf{41688}\\
$J=6$    &71004  \textbf{37567} &58487 \textbf{36345} &47564 \textbf{34920} &42935 \textbf{34326}\\
$J=8$    &63452  \textbf{38703} &52990 \textbf{36586} &46914 \textbf{35836} &43647 \textbf{35129}\\
$J=10$   &60061  \textbf{39078} &52125 \textbf{38025} &46912 \textbf{36455} &45863 \textbf{36076}\\
$J=12$   &61098  \textbf{40834} &52296 \textbf{38657} &46458 \textbf{37203} &46698 \textbf{36781}\\
$J=14$   &57924  \textbf{42348} &55622 \textbf{40363} &50243 \textbf{39613} &47619 \textbf{37243}\\
$J=16$   &63213  \textbf{44067} &55206 \textbf{42973} &50322 \textbf{41485} &48966 \textbf{39446}\\
$J=18$   &64056  \textbf{46050} &55461 \textbf{45133} &53652 \textbf{45308} &51870 \textbf{42485}\\
$J=20$   &65215  \textbf{49046} &58911 \textbf{47430} &56009 \textbf{45390} &53213 \textbf{43888}\\

\hline \hline
 &{\em logitboost} & \textbf{\em abc-logit}\\\hline
  &$\nu = 0.04$ &$\nu=0.06$ &$\nu=0.08$ &$\nu=0.1$ \\
\hline

$J=4$    &145368 \textbf{102014} &138734 \textbf{70886} &126980 \textbf{50783} &116346 \textbf{41551}\\
$J=6$    &80782  \textbf{42699} &58769 \textbf{38592} &51202 \textbf{37397} &48199 \textbf{36914}\\
$J=8$    &68065  \textbf{45737} &56678 \textbf{39648} &52504 \textbf{37935} &46600 \textbf{36731}\\
$J=10$   &63153  \textbf{44517} &55419 \textbf{40286} &52835 \textbf{40044} &46913 \textbf{37504}\\
$J=12$   &66240  \textbf{47948} &56619 \textbf{44602} &50918 \textbf{41582} &47128 \textbf{39378}\\
$J=14$   &63763  \textbf{52063} &55238 \textbf{47642} &51526 \textbf{44296} &47545 \textbf{42720}\\
$J=16$   &66543  \textbf{52937} &57473 \textbf{50842} &53287 \textbf{47578} &51106 \textbf{46635}\\
$J=18$   &68477  \textbf{56803} &57070 \textbf{55166} &54954 \textbf{49956} &51603 \textbf{47707}\\
$J=20$   &68311  \textbf{61980} &61047 \textbf{57383} &56474 \textbf{53364} &51242 \textbf{49506}
\\\hline\hline
\end{tabular}}}
\end{center}
\label{tab_Poker25kT2}
\end{table}

\clearpage

\subsection{Detailed Experiment Results on {\em Letter4k} and {\em Letter2k}}

\begin{table}[h]
\caption{\textbf{\em Letter4k}. Upper table: The test mis-classification errors of  {\em mart} and \textbf{\em abc-mart} (bold numbers). Bottom table: The test mis-classification errors of  {\em logitboost} and \textbf{\em abc-logitboost} (bold numbers)}
\begin{center}
{\small
{\begin{tabular}{l r r r r }
\hline \hline
 &{\em mart} & \textbf{\em abc-mart}\\\hline
  &$\nu = 0.04$ &$\nu=0.06$ &$\nu=0.08$ &$\nu=0.1$ \\
\hline

$J=4$    &1681 \textbf{1415} &1660 \textbf{1380} &1671 \textbf{1368} &1655 \textbf{1323}\\
$J=6$    &1618  \textbf{1320} &1584 \textbf{1288} &1588 \textbf{1266} &1577 \textbf{1240}\\
$J=8$    &1531  \textbf{1266} &1522 \textbf{1246} &1516 \textbf{1192} &1521 \textbf{1184}\\
$J=10$   &1499  \textbf{1228} &1463 \textbf{1208} &1479 \textbf{1186} &1470 \textbf{1185}\\
$J=12$   &1420  \textbf{1213} &1434 \textbf{1186} &1409 \textbf{1170} &1437 \textbf{1162}\\
$J=14$   &1410  \textbf{1190} &1388 \textbf{1156} &1377 \textbf{1151} &1396 \textbf{1160}\\
$J=16$   &1395  \textbf{1167} &1402 \textbf{1156} &1396 \textbf{1157} &1387 \textbf{1146}\\
$J=18$   &1376  \textbf{1164} &1375 \textbf{1139} &1357 \textbf{1127} &1352 \textbf{1152}\\
$J=20$   &1386  \textbf{1154} &1397 \textbf{1130} &1371 \textbf{1131} &1370 \textbf{1149}\\
$J=24$   &1371  \textbf{1148} &1348 \textbf{1155} &1374 \textbf{1164} &1391 \textbf{1150}\\
$J=30$   &1383  \textbf{1174} &1406 \textbf{1174} &1401 \textbf{1177} &1404 \textbf{1209}\\
$J=40$   &1458  \textbf{1211} &1455 \textbf{1224} &1441 \textbf{1233} &1454 \textbf{1215}\\

\hline \hline
 &{\em logitboost} & \textbf{\em abc-logit}\\\hline
  &$\nu = 0.04$ &$\nu=0.06$ &$\nu=0.08$ &$\nu=0.1$ \\
\hline

$J=4$    &1460 \textbf{1296} &1471 \textbf{1241} &1452 \textbf{1202} &1446 \textbf{1208}\\
$J=6$    &1390  \textbf{1143} &1394 \textbf{1117} &1382 \textbf{1090} &1374 \textbf{1074}\\
$J=8$    &1336  \textbf{1089} &1332 \textbf{1080} &1311 \textbf{1066} &1297 \textbf{1046}\\
$J=10$   &1289  \textbf{1062} &1285 \textbf{1067} &1380 \textbf{1034} &1273 \textbf{1049}\\
$J=12$   &1251  \textbf{1058} &1247 \textbf{1069} &1261 \textbf{1044} &1243 \textbf{1051}\\
$J=14$   &1247  \textbf{1063} &1233 \textbf{1051} &1251 \textbf{1040} &1244 \textbf{1066}\\
$J=16$   &1244  \textbf{1074} &1227 \textbf{1068} &1231 \textbf{1047} &1228 \textbf{1046}\\
$J=18$   &1243  \textbf{1059} &1250 \textbf{1040} &1234 \textbf{1052} &1220 \textbf{1057}\\
$J=20$   &1226  \textbf{1084} &1242 \textbf{1070} &1242 \textbf{1058} &1235 \textbf{1055}\\
$J=24$   &1245  \textbf{1079} &1234 \textbf{1059} &1235 \textbf{1058} &1215 \textbf{1073}\\
$J=30$   &1232  \textbf{1057} &1247 \textbf{1085} &1229 \textbf{1069} &1230 \textbf{1065}\\
$J=40$   &1246  \textbf{1095} &1255 \textbf{1093} &1230 \textbf{1094} &1231 \textbf{1087}
\\\hline\hline
\end{tabular}}}
\end{center}
\label{tab_letter4k}
\end{table}

\begin{table}[h]
\caption{\textbf{\em Letter2k}. Upper table: The test mis-classification errors of  {\em mart} and \textbf{\em abc-mart} (bold numbers). Bottom table: The test mis-classification errors of  {\em logitboost} and \textbf{\em abc-logitboost} (bold numbers)}
\begin{center}
{\small
{\begin{tabular}{l r r r r }
\hline \hline
 &{\em mart} & \textbf{\em abc-mart}\\\hline
  &$\nu = 0.04$ &$\nu=0.06$ &$\nu=0.08$ &$\nu=0.1$ \\
\hline

$J=4$    &2694 \textbf{2512} &2698 \textbf{2470} &2684 \textbf{2419} &2689 \textbf{2435}\\
$J=6$    &2683  \textbf{2360} &2664 \textbf{2321} &2640 \textbf{2313} &2629 \textbf{2321}\\
$J=8$    &2569  \textbf{2279} &2603 \textbf{2289} &2563 \textbf{2259} &2571 \textbf{2251}\\
$J=10$   &2534  \textbf{2242} &2516 \textbf{2215} &2504 \textbf{2210} &2491 \textbf{2185}\\
$J=12$   &2503  \textbf{2202} &2516 \textbf{2215} &2473 \textbf{2198} &2492 \textbf{2201}\\
$J=14$   &2488  \textbf{2203} &2467 \textbf{2231} &2460 \textbf{2204} &2460 \textbf{2183}\\
$J=16$   &2503  \textbf{2219} &2501 \textbf{2219} &2496 \textbf{2235} &2500 \textbf{2205}\\
$J=18$   &2494  \textbf{2225} &2497 \textbf{2212} &2472 \textbf{2205} &2439 \textbf{2213}\\
$J=20$   &2499  \textbf{2199} &2512 \textbf{2198} &2504 \textbf{2188} &2482 \textbf{2220}\\
$J=24$   &2549  \textbf{2200} &2549 \textbf{2191} &2526 \textbf{2218} &2538 \textbf{2248}\\
$J=30$   &2579  \textbf{2237} &2566 \textbf{2232} &2574 \textbf{2244} &2574 \textbf{2285}\\
$J=40$   &2641  \textbf{2303} &2632 \textbf{2304} &2606 \textbf{2271} &2667 \textbf{2351}\\
\hline \hline
 &{\em logitboost} & \textbf{\em abc-logit}\\\hline
  &$\nu = 0.04$ &$\nu=0.06$ &$\nu=0.08$ &$\nu=0.1$ \\
\hline

$J=4$    &2629 \textbf{2347} &2582 \textbf{2299} &2580 \textbf{2256} &2572 \textbf{2231}\\
$J=6$    &2427  \textbf{2136} &2450 \textbf{2120} &2428 \textbf{2072} &2429 \textbf{2077}\\
$J=8$    &2336  \textbf{2080} &2321 \textbf{2049} &2326 \textbf{2035} &2313 \textbf{2037}\\
$J=10$   &2316  \textbf{2044} &2306 \textbf{2003} &2314 \textbf{2021} &2307 \textbf{2002}\\
$J=12$   &2315  \textbf{2024} &2315 \textbf{1992} &2333 \textbf{2018} &2290 \textbf{2018}\\
$J=14$   &2317  \textbf{2022} &2305 \textbf{2004} &2315 \textbf{2006} &2292 \textbf{2030}\\
$J=16$   &2302  \textbf{2024} &2299 \textbf{2004} &2286 \textbf{2005} &2262 \textbf{1999}\\
$J=18$   &2298  \textbf{2044} &2277 \textbf{2021} &2301 \textbf{1991} &2282 \textbf{2034}\\
$J=20$   &2280  \textbf{2049} &2268 \textbf{2021} &2294 \textbf{2024} &2309 \textbf{2034}\\
$J=24$   &2299  \textbf{2060} &2326 \textbf{2037} &2285 \textbf{2021} &2267 \textbf{2047}\\
$J=30$   &2318  \textbf{2078} &2326 \textbf{2057} &2304 \textbf{2041} &2274 \textbf{2045}\\
$J=40$   &2281  \textbf{2121} &2267 \textbf{2079} &2294 \textbf{2090} &2291 \textbf{2110}
\\\hline\hline
\end{tabular}}}
\end{center}
\label{tab_letter2k}
\end{table}

\clearpage

\section{Conclusion}

Classification is a fundamental task in machine learning. This paper presents extensive experiment results of \textbf{four} tree-based boosting algorithms: {\em mart}, {\em abc-mart}, {\em (robust) logitboost}), and {\em abc-logitboost}, for multi-class classification, on a variety of publicly available datasets.  From the experiment results, we can conclude the following:
\begin{enumerate}
\item {\em Abc-mart} considerably improves {\em mart}.
\item {\em Abc-logitboost} considerably improves {\em (robust) logitboost}.
\item {\em (Robust) logitboost} considerably improves {\em mart} on most datasets.
\item {\em Abc-logitboost}  considerably improves {\em abc-mart} on most datasets.
\item These four boosting algorithms (especially {\em abc-logitboost})  outperform SVM on many datasets.
\item Compared to the best deep learning methods, these four boosting algorithms (especially {\em abc-logitboost}) are competitive.
\end{enumerate}


\end{document}